\documentclass{article}

\usepackage{iclr2027_conference,times}
\iclrfinalcopy
\usepackage{amsmath,amssymb,amsthm,mathtools}
\usepackage{array}
\usepackage{booktabs}
\usepackage{caption}
\usepackage{enumitem}
\usepackage{graphicx}
\usepackage{wrapfig}
\usepackage{tikz}
\usepackage{microtype}
\usepackage{placeins}
\usepackage{url}
\usepackage[hypertexnames=false]{hyperref}
\hypersetup{
  colorlinks=true,
  allcolors=blue!50!black,
  pdftitle={Explanation Selection in Abductive Learning},
  pdfauthor={En-Hao Gao and Yuan Jiang},
}

\newcommand{\one}{\mathbf{1}}
\newcommand{\sigmoid}{\sigma}

\setlist[itemize]{topsep=2pt,itemsep=2.5pt,parsep=1pt,partopsep=0pt,leftmargin=1.5em}
\setlist[enumerate]{topsep=2pt,itemsep=1.5pt,parsep=0pt,partopsep=0pt,leftmargin=1.8em}
\makeatletter
\def\fps@figure{!t}
\def\fps@table{!t}
\def\thm@space@setup{%
  \thm@preskip=5pt plus 1pt minus 1pt
  \thm@postskip=\thm@preskip}
\renewcommand\paragraph{\@startsection{paragraph}{4}{\z@}%
  {0.5ex \@plus 0.2ex \@minus .1ex}%
  {-1em}%
  {\normalfont\normalsize\bfseries}}
\def\section{\@startsection{section}{1}{\z@}%
  {-1.8ex plus -0.4ex minus -.2ex}%
  {1.2ex plus 0.2ex minus 0.2ex}%
  {\large\sc\raggedright}}
\def\subsection{\@startsection{subsection}{2}{\z@}%
  {-1.6ex plus -0.4ex minus -.2ex}%
  {0.7ex plus .2ex}%
  {\normalsize\sc\raggedright}}
\makeatother
\AtBeginDocument{%
  \raggedbottom
  \setlength{\abovedisplayskip}{6pt plus 1.5pt minus 1.5pt}%
  \setlength{\belowdisplayskip}{6pt plus 1.5pt minus 1.5pt}%
  \setlength{\abovedisplayshortskip}{3pt plus 1pt minus 1pt}%
  \setlength{\belowdisplayshortskip}{5.5pt plus 1.5pt minus 1.5pt}%
}

\theoremstyle{plain}
\newtheorem{theorem}{Theorem}
\newtheorem{proposition}[theorem]{Proposition}
\newtheorem{lemma}{Lemma}
\newtheorem{corollary}{Corollary}
\theoremstyle{definition}

\theoremstyle{remark}
\newtheorem{remark}{Remark}
\newtheorem{example}{Example}

\title{Explanation Selection in Abductive Learning}
\author{En-Hao Gao\\
\texttt{gaoeh@lamda.nju.edu.cn}
\And
Yuan Jiang\\
\texttt{jiangyuan@nju.edu.cn}}

\begin{document}

\maketitle
\fancyhead{}
\renewcommand{\headrulewidth}{0pt}
\begin{abstract}
\setlength{\parfillskip}{0pt}
Abductive learning (ABL) offers a promising framework that connects machine learning and logical reasoning in a mutually beneficial loop.
However, this bidirectional feedback can also reinforce incorrect concept semantics, since different interpretations of intermediate concepts may explain the same observed reasoning outcomes.
Static analyses characterize this ambiguity but leave open which explanation training selects and whether its labels can still be corrected.
We develop a dynamical analysis of a fixed-feature binary ABL model that links the local stability of competing explanations to differences in their concept labels.
We prove that Gram matrices with identical spectra can select different explanations even under the same supervision, objective, and initialization.
When any two compatible explanations differ in at least two labels, legal evidence above one half guarantees during training that admissible changes in optimization geometry preserve the selected explanation.
Even when only the true explanation satisfies all constraints under a shared predictor, training can still converge to a finite attracting state whose hard predictions violate some constraints and mislabel concepts.
Experiments on visual ABL tasks verify these predictions, including wrong attraction from random initializations under identifiable supervision, and show that in ordinary end-to-end training SGD and Adam select different explanations of the same supervision, while unseen concept accuracy drops relative to true-concept references as training concept errors grow.
\end{abstract}

\section{Introduction}
\label{sec:introduction}

Unifying machine learning and logical reasoning is a central ambition of artificial intelligence \citep{marra2024statistical}.
Abductive learning (ABL) offers a promising framework for this goal through a reciprocal interaction between machine learning and logical reasoning \citep{zhou2019abductive,zhou2022abductive}.
In ABL, the learner first predicts concept labels from data, and logical rules connect these labels to observed reasoning outcomes.
Abductive reasoning then uses those outcomes to revise the predictions, producing labels that supervise the next round of concept learning.
Repeating this cycle allows domain knowledge to shape concept learning without requiring a differentiable reasoner.

Theoretical studies have exposed a fundamental ambiguity in learning concepts from reasoning outcomes.
The same rules and observed outcomes may admit several complete sets of concept labels for the training data, each constituting a \emph{compatible explanation}.
Reproducing the observed outcomes with incorrect concept labels gives rise to \emph{reasoning shortcuts} \citep{marconato2023notall}.
Existing theoretical analyses characterize the generalization risk associated with reasoning shortcuts \citep{yang2024analysis}, establish conditions for concept identifiability \citep{bortolotti2025identifiability}, and describe the geometry of model predictions that satisfy logical constraints \citep{vankrieken2024independence}.
Further related work on neuro-symbolic learning and optimization geometry is discussed in Appendix~\ref{app:related-work}.

Knowing which explanations are compatible with supervision leaves open which one, if any, a training trajectory will reach.
Because each reasoning step depends on the current predictions, the feedback can progressively favor one explanation over equally compatible alternatives.
A dynamical account must explain which explanations attract nearby states, how optimization selects among them from a common initialization, and when subsequent updates can or cannot revise the learned labels.

We analyze these questions in a fixed-feature binary ABL model based on the formal product-posterior update introduced in A\textsuperscript{3}BL \citep{he2024ambiguity}.
At each iteration, we compute an exact posterior over all compatible explanations and use its marginal label probabilities as fixed soft targets for one full-batch cross-entropy update.
The resulting dynamics follow preconditioned gradient ascent on a common potential derived from the \emph{legal evidence}, the probability that the model's predicted labels jointly satisfy all constraints.
Local analysis then yields stability criteria based on the number and overlap of label differences between compatible explanations.
Several explanations can be locally attracting, so stability alone does not determine which one training selects.
That choice depends on the Gram matrix, which maps the potential gradient to an update direction for the model's concept predictions.
By holding the Gram eigenvalues fixed and varying the eigenvectors, we show that isospectral Grams can select arbitrarily many legal explanations from a common open set of initializations.

Sensitivity to optimization geometry need not persist throughout training.
When compatible explanations differ pairwise in at least two labels, legal evidence above \(1/2\) certifies capture of the selected explanation: changing any predicted label would require crossing a tie region of lower legal evidence, and our spectral and step-size bounds prevent such a crossing even when the Gram matrix varies adaptively.
Training can therefore entrench a compatible explanation with incorrect concept labels, placing a sharp limit on what later changes in optimization geometry can correct.
If a shared predictor rules out these incorrect compatible alternatives so that only the true explanation is realizable, must training recover it?
Even then, exact abduction performed separately on each query can reinforce consistent errors across inputs coupled by the shared predictor.
Rather than approaching a compatible endpoint, nearby training trajectories can converge to a finite, strictly suboptimal fixed point whose hard predictions violate some constraints while mislabeling concepts.

Our contributions are summarized as follows:

\begin{itemize}
\item \textbf{Local stability.}
We characterize local asymptotic stability in terms of label differences and defect overlap among compatible explanations and establish \(O(1/t)\) convergence rates in the resulting stable regimes (Sections~\ref{sec:dynamics}--\ref{sec:candidate-geometry}).

\item \textbf{Explanation selection.}
We prove that Gram matrices with identical full and active spectra can robustly select arbitrarily many distinct locally attracting explanations from a common open set of initializations (Section~\ref{sec:robust-isospectral}).

\item \textbf{Certified capture.}
When compatible explanations differ in at least two labels, we establish an optimal \(1/2\) legal-evidence certificate for label preservation and endpoint convergence under adaptive metrics and arbitrary-rank shared heads (Section~\ref{sec:capture-subsec}).

\item \textbf{Limits of recovery.}
We prove that even when only the true explanation satisfies all constraints under a shared predictor, training trajectories can converge to a finite wrong attractor whose hard predictions violate some constraints (Section~\ref{sec:recovery-subsec}).
\end{itemize}

In Section~\ref{sec:experiments}, experiments on visual ABL tasks verify these dynamical predictions and show that standard optimizers exhibit similar explanation selection in ordinary end-to-end training.

\section{Training Dynamics and a Common Potential}
\label{sec:dynamics}

We study a soft-label formulation of ABL over propositional variables with a fixed feature representation.
The binary concept labels are unknown, and background logical rules relate them to the observed reasoning outcomes.
For \(N\) concept labels, let \(\mathcal C\subseteq\{\pm1\}^N\) be the finite set of \emph{complete codes}, the full label assignments compatible with these rules and observations.
We encode the intended semantic label on each coordinate as \(+1\) and orient the margins \(y\in\mathbb R^N\) so that \(y_i>0\) favors that label.
We assume \(\mathbf1_N\in\mathcal C\) and \(|\mathcal C|\ge2\), so the correct explanation is legal but has at least one alternative.
The semantic labels are used only for analysis and are unavailable during training.

\subsection{Legal evidence and posterior refresh}

To form soft targets, we condition the model's independent label predictions on joint satisfaction of the logical constraints \citep{he2024ambiguity}.
With \(\sigma(s)=(1+e^{-s})^{-1}\), the code likelihood, legal evidence and candidate posterior are
\begin{equation}
L_z(y)=\prod_{i=1}^N\sigma(z_i y_i),\qquad
\mathcal E(y)=\sum_{z\in\mathcal C}L_z(y),\qquad
q_z(y)=\frac{L_z(y)}{\mathcal E(y)}.
\label{eq:posterior}
\end{equation}
The \emph{legal evidence} \(\mathcal E\) is the probability that the predicted labels jointly satisfy all constraints.
The posterior \(q\) normalizes likelihoods directly, corresponding to uniform priors and unit temperature.
Its marginals \(p_i(y)=\Pr_q(z_i=+1)\) sum the posterior weights of codes assigning \(+1\) to coordinate \(i\).

At each iteration, we refresh \(q\) at the current margins and use its marginals as soft targets for one full-batch cross-entropy update, holding them fixed during differentiation.
The sample Gram matrix \(K\) maps the residual \(R\) between targets and predictions to the margin update:
\begin{equation}
y_{t+1}=y_t+\eta K R(y_t),\qquad
R_i(y)=2[p_i(y)-\sigma(y_i)].
\label{eq:global-map}
\end{equation}
Here \(\eta>0\) absorbs the loss normalization, and the factor two fixes our residual convention.
This posterior and update define the \emph{native} model used unless stated otherwise.
For a fixed feature matrix \(X\in\mathbb R^{N\times d}\) and a shared linear head \(w\in\mathbb R^d\), parameter updates with a positive-definite preconditioner \(M\) induce the following margin dynamics:
\begin{equation}
y=Xw,\qquad w^+=w+\eta MX^\top R(y),\qquad K_M=XMX^\top.
\label{eq:parameter-sample-bridge}
\end{equation}
The parameter metric \(M\) controls the induced sample Gram matrix \(K_M\) for a fixed representation.
Ordinary parameter gradient descent uses \(M=I\) and hence \(K_M=XX^\top\), which generally differs from \(K=I\) in sample space.
We first analyze positive-definite Gram matrices \(K\succ0\), each realizable by \(X=K^{1/2}\), and then treat shared heads whose induced sample Grams \(K_M\) may be singular, including when repeated occurrences of an input across separate queries occupy distinct coordinates of \(y\) while sharing predictions through \(X\).

\subsection{What remains fixed when the Gram changes}

With exact posterior refresh, the residual of the soft-label update is the gradient of a common potential:
\begin{equation}
\Phi(y)=2\log\mathcal E(y),\qquad \nabla\Phi(y)=R(y).
\label{eq:potential}
\end{equation}
This potential is a rescaled negative semantic loss \citep{xu2018semantic}.
The Gram determines the ascent direction but, for \(K\succ0\), leaves finite fixed points unchanged because \(KR=0\) if and only if \(R=0\).
A singular Gram may also map a nonzero residual to zero, creating additional fixed points.

For \(K\succ0\), ascent on the common potential is guaranteed by the sufficient step-size condition
\begin{equation}
\eta\lambda_{\max}(K)(N+1)<4.
\label{eq:ascent-condition}
\end{equation}
Under this condition, the smoothness bound in Appendix~\ref{app:roadmap} gives \(\Phi(y_{t+1})>\Phi(y_t)\) whenever \(R(y_t)\ne0\).
We assume \eqref{eq:ascent-condition} for all fixed-Gram updates unless stated otherwise.

\section{Which Explanations Are Locally Stable?}
\label{sec:candidate-geometry}

A trajectory approaches the \emph{code endpoint} \(z\) when \(z_i y_{i,t}\to+\infty\) for every coordinate, so its independent label predictions concentrate on the complete assignment \(z\).
Local stability depends on which concept labels differ between a code and its competitors.
Fix \(z\in\mathcal C\) and call every other code a \emph{rival}.
Write \(d_{\min}(z)=\min_{w\ne z}d_{\mathrm H}(w,z)\), where Hamming distance counts the labels on which two codes differ.
To analyze stability at infinite signed margins, we express the dynamics in compactified coordinates that place this endpoint at a finite boundary point:
\begin{equation}
a_i=e^{-z_i y_i},\qquad r=\sum_i a_i.
\label{eq:compactified-coords}
\end{equation}
\(a_i\) is the ratio of the model probabilities for labels \(-z_i\) and \(z_i\), so small \(a_i\) means confident agreement with \(z_i\).
The endpoint lies at \(a=0\), with \(r\) its \(\ell_1\) distance.
A coordinate \(a_i=0\) represents a prediction already saturated at label \(z_i\).
The closed orthant includes these saturated predictions, so stability must cover perturbations on its boundary faces as well as states with finite logits.
Formally, the endpoint is \emph{locally asymptotically stable} if the continuous extension of the update to \([0,\infty)^N\) is Lyapunov stable at \(0\) and attracts a relative neighborhood.
\label{def:viability}
Proofs and extensions are in Appendix~\ref{app:local-stability}.

\subsection{Local stability in the native model}

\begin{theorem}[Native local endpoints]
\label{cor:native-a3bl}
Use uniform priors, unit temperature and no direct semantic supervision.
For every fixed \(K\succ0\) and \(\eta>0\) satisfying~\eqref{eq:ascent-condition}, a code \(z\) is locally asymptotically stable in the closed compactified orthant if and only if \(d_{\min}(z)\ge2\).
In the stable case, a sufficiently small neighborhood is trapping and \(r_t=O(1/t)\).
\end{theorem}

The distance condition identifies when logical conditioning can resolve uncertainty in one label.
If flipping that label yields another legal code, either value satisfies the constraints when all other labels are fixed.
Conditioning leaves the prediction unchanged, producing a flat zero-loss edge between the codes \citep{vankrieken2024independence}.
Without a restoring signal along this edge, the endpoint cannot attract a full neighborhood, though finite trajectories may still approach it as shown in Appendix~\ref{app:local-scope}.

Without a one-bit rival, a single-label deviation is illegal, and every competing legal assignment requires several deviations whose probabilities multiply.
Near the endpoint, most probability lost from the current code therefore goes to illegal assignments.
Conditioning removes this mass and reinforces the code's labels, whether or not they are semantically correct.
The Gram mixes these corrections, so individual signed margins may temporarily decrease.
Near the endpoint, the evidence deficit \(1-\mathcal E(y)\) equals \(r\) to first order.
Theorem~\ref{cor:native-a3bl} shows that ascent of the common potential drives \(r\) to zero under every admissible fixed positive-definite Gram.
This attraction persists under time-varying posterior-refresh errors if their effect on the residual remains sufficiently small relative to the exact residual, with the same \(O(1/t)\) rate under a tighter step-size bound.

\subsection{Why distance, temperature and overlap matter}

Priors and temperature modify how the posterior distributes mass between a code and its rivals.
Allow positive priors \(\pi_z\) and temperature \(\tau>0\), and set \(q_z(y)\propto[\pi_zL_z(y)]^{1/\tau}\).
We use the resulting marginals in the same detached update, with no direct semantic supervision.
The corresponding potential is \(\Phi(y)=2\tau\log\sum_{z\in\mathcal C}[\pi_zL_z(y)]^{1/\tau}\).
For a rival \(w\), the \emph{defect set} \(D(w,z)=\{i:w_i\ne z_i\}\) records the coordinates on which it differs from \(z\).
In the posterior ratio, the likelihood factors for agreeing labels cancel, leaving only the prior ratio and the predictions on the defect set:
\begin{equation}
\frac{q_w(y)}{q_z(y)}
=\left(\frac{\pi_w}{\pi_z}\right)^{1/\tau}
 \prod_{i\in D(w,z)}a_i^{1/\tau}.
\label{eq:monomial-odds}
\end{equation}
As confidence in \(z\) increases, each differing label contributes another vanishing factor to a rival's posterior odds.
Higher temperature reduces these exponents, slowing the loss of rival support, whereas priors change only the coefficients.
Since the code's confidence loss is first order in \(r\), \(d_{\min}(z)/\tau\) determines whether rival support vanishes faster or slower than this loss.

At \(d_{\min}(z)=\tau\), rival support and the loss of confidence in \(z\) have the same order.
Stability now depends on how uncertainty is distributed across labels: uncertainty on a shared coordinate can support several rivals at once.
Write \(a=r u\), where \(u\in\Delta^{N-1}\) is nonnegative and sums to one.
Its entries allocate the total \(r\) across coordinates, including boundary directions.
Define
\begin{equation}
H_z(u)=\tau\!\!\sum_{\substack{w\ne z\\d_{\mathrm H}(w,z)=\tau}}
\left(\frac{\pi_w}{\pi_z}\right)^{1/\tau}
\prod_{i\in D(w,z)}u_i^{1/\tau},
\qquad
\chi_z=\max_{u\in\Delta^{N-1}}H_z(u).
\label{eq:H-functional}
\end{equation}
Here \(H_z(u)\) sums the nearest rivals' contributions, and \(\chi_z\) maximizes this sum over allocations as a concave maximization over the simplex.
The unit threshold compares this rival support with the current code's first-order confidence loss.
Direct semantic supervision increases the restoring coefficient for the correct code but leaves the rival exponents unchanged, shifting the critical balance without changing the distance--temperature comparison (Corollary~\ref{cor:support-coefficient}).

\begin{proposition}[Local classification with priors and temperature]
\label{cor:local-iff}
For every fixed \(K\succ0\) and \(\eta>0\) satisfying \(\eta\lambda_{\max}(K)(N/\tau+1)<4\), with positive priors and no direct semantic supervision, local asymptotic stability holds precisely when
\[
d_{\min}(z)>\tau,\qquad\text{or}\qquad
d_{\min}(z)=\tau\ \text{ and }\ \chi_z<1,
\]
apart from the equality case \(d_{\min}(z)=\tau,\ \chi_z=1\).
In either stable regime above, \(z\) has a trapping neighborhood and \(r_t=O(1/t)\).
The equality case requires higher-order or exact analysis.
\end{proposition}

\begin{example}[Rivals can jointly defeat local attraction]
\label{ex:overlap}
Let \(N=4\), \(z=(+,+,+,+)\), \(\tau=2\), and give two distance-two rivals the same prior ratio \(\pi_w/\pi_z=0.64\).
Either rival alone leaves \(z\) locally attracting.
With disjoint defect sets \(\{1,2\}\) and \(\{3,4\}\), uncertainty that supports one rival cannot support the other, and attraction persists.
Changing the second defect set to \(\{1,3\}\) lets uncertainty in coordinate 1 support both rivals at once, and \(z\) loses local attraction.
\end{example}

\section{Same Spectrum, Different Explanations}
\label{sec:robust-isospectral}

We have proved that several explanations can be locally attracting.
In this section, we study how Gram orientation affects selection among them while holding the spectra fixed.
Proofs are in Appendix~\ref{app:routing}.

\subsection{A single slow direction inside the candidate subspace}

For any \(m\ge2\), partition \(N=2m\) coordinates into pairs and let \(\mathcal C_m=\{z_0,\ldots,z_m\}\), where \(z_0=\mathbf1_{2m}\) and \(z_j\) flips only pair \(j\).
In the native model, each code has minimum rival distance two and is locally asymptotically stable by Theorem~\ref{cor:native-a3bl}.
All codes give the two coordinates of a pair the same label, so their relative posterior weights depend on the sum of its logits.
Equal and opposite logit changes leave these weights unchanged.
Equal changes within each pair follow the directions
\[
c_j=(e_{2j-1}+e_{2j})/\sqrt2,\qquad B=[c_1,\ldots,c_m].
\]
Here \(e_i\) is the \(i\)th standard basis vector.
More generally, relative posterior weights depend only on logits projected onto the \emph{candidate-active subspace} \(S_{\mathcal C}=\operatorname{span}\{z-w:z,w\in\mathcal C\}\).
For \(\mathcal C_m\), \(B^\top K B\) is the Gram in the orthonormal basis \(B\), and its eigenvalues form the active spectrum.

Choose a slow eigenvalue \(\alpha\in(0,1)\) and a unit vector \(v\in\mathbb R^m\) orthogonal to \(\mathbf1_m\), and set
\begin{align}
H_0&=I_m-(1-\alpha)vv^\top,\qquad
H_j=I_m-(1-\alpha)e_je_j^\top\quad(1\le j\le m),\notag\\
K_j&=BH_jB^\top+I_{2m}-BB^\top,\qquad j=0,\ldots,m.
\label{eq:main-isospectral-grams}
\end{align}

\begin{wrapfigure}[14]{r}{0.32\textwidth}
\vspace{-9pt}
\centering
\makebox[\linewidth][l]{\hspace*{-0.05\linewidth}\includegraphics[width=\linewidth]{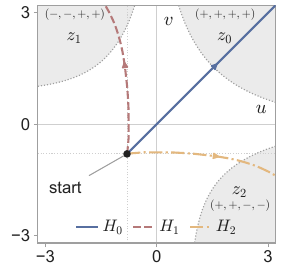}}
\captionsetup{width=\linewidth,justification=centering,singlelinecheck=false}
\caption{Isospectral selection.
Gray shading marks \(\mathcal E>1/2\).}
\label{fig:routing-capture}
\end{wrapfigure}
Each \(K_j\) acts as \(H_j\) on the candidate-active subspace and as the identity on its orthogonal complement.
The slow direction stays inside this subspace, so rotating it preserves both the full and active spectra.
Consider the common symmetric initialization \(y_0=s_0\mathbf1_{2m}\) with \(s_0<0\).
Under \(K_0\), the slow direction is orthogonal to this initialization, so all pairs move together toward positive labels.
Under \(K_j\), pair \(j\) initially moves toward positive labels too, but remains negative while the faster pairs cross zero.
Rivals requiring negative labels on those other pairs lose posterior weight, whereas \(z_j\) remains supported by pair \(j\).
Once the other pairs are confidently positive, \(z_0\) is the main remaining competitor, and the target pair's negative margins favor \(z_j\).
With the rate separation in Theorem~\ref{thm:arbitrary-many-isospectral}, the resulting feedback reverses the target pair's update, reinforcing its negative labels.
Gram orientation thus changes which explanation the inferred targets reinforce, rather than merely the speed of learning.

Figure~\ref{fig:routing-capture} visualizes the case \(N=4\) (\(m=2\)) with \(\alpha=1/6\), \(\eta=0.02\), and \(y_0=-0.8\mathbf1_4\).
The trajectories follow the reduced \(H_j\) dynamics in the plane \(y=(u,u,v,v)\), so the two axes track the common margin in each pair.
The posterior refresh and all three Grams preserve equality within each pair.
The Grams share full spectrum \(\{1/6,1,1,1\}\) and active spectrum \(\{1/6,1\}\).
Each trajectory enters a different shaded region, meeting the capture criterion that will be introduced in Section~\ref{sec:capture}.

\subsection{Robust routing from a common open set}

The preceding analysis starts from a symmetric initial state.
We now show that this symmetry is not essential: each Gram retains its selected endpoint under small perturbations of both the initial predictions and the matrix itself.
Below, \(\mathbb S_{++}^{2m}\) denotes the cone of symmetric positive-definite \(2m\times2m\) matrices, and bracketed superscripts record eigenvalue multiplicities.

\begin{theorem}[Robust full- and active-isospectral routing]
\label{thm:arbitrary-many-isospectral}
For every \(m\ge2\) and \(s_0<0\), there exist \(\alpha_m\in(0,1)\) and a common \(\eta_m>0\) such that the Grams in \eqref{eq:main-isospectral-grams} satisfy the ascent condition and
\begin{equation}
\operatorname{spec}(K_j)=\{\alpha_m,1^{[2m-1]}\},\qquad
\operatorname{spec}(B^\top K_jB)=\{\alpha_m,1^{[m-1]}\}.
\label{eq:m-spectrum-identity}
\end{equation}
There are a common open neighborhood \(\mathcal U_m\subset\mathbb R^{2m}\) of \(y_\star=(s_0,\ldots,s_0)\) and open Gram neighborhoods \(\mathcal V_{m,j}\subset\mathbb S_{++}^{2m}\) of \(K_j\) such that
\begin{equation}
y_0\in\mathcal U_m,\quad K\in\mathcal V_{m,j}
\quad\Longrightarrow\quad F_K^t(y_0)\to z_j,
\qquad j=0,\ldots,m,
\label{eq:m-robust-routing}
\end{equation}
where \(F_K(y)=y+\eta_m KR(y)\) and convergence is in the endpoint sense.
\end{theorem}

The open neighborhoods allow all initial margins to vary, so exact symmetry is unnecessary.
Each route enters a region that attracts its selected code, and small perturbations of the preceding trajectory preserve this outcome.
The proof in Appendix~\ref{app:routing} makes this argument uniform over nearby initial states and Gram matrices.
General Gram perturbations may change the spectra, but small rotations of the slow direction within the active subspace preserve both spectra and the endpoint (Appendix~\ref{app:routing-main-proof}).
Moreover, tying the two predictions of each pair through the shared features \(X=B\) retains every direction that distinguishes the candidates, so singular sample Grams induced by the parameter metrics \(H_j\) likewise produce robust isospectral selection from a common open set of heads (Appendix~\ref{app:rank-deficient-selection}).

\section{Capture and the Limits of Recovery}
\label{sec:capture}

Section~\ref{sec:robust-isospectral} shows that Gram orientation can change the selected explanation.
We now ask two questions: when does the current training state certify that later metric changes cannot revise a selected explanation, and does excluding incorrect compatible explanations from a shared predictor ensure recovery of the true concepts?
Proofs and extensions are in Appendices~\ref{app:capture} and~\ref{app:identifiable-recovery}.

\subsection{Certifying the selected explanation}
\label{sec:capture-subsec}

For native updates, legal evidence above one half at a finite training state certifies label preservation and endpoint convergence under every subsequent Gram sequence satisfying the spectral and step-size bounds below.
We assume that candidate codes differ pairwise in at least two labels.

\begin{theorem}[Half-evidence capture]
\label{thm:routing-certificate}
Let \(|\mathcal C|\ge2\) and \(d_{\mathrm H}(z,w)\ge2\) for all distinct \(z,w\in\mathcal C\).
Use the native exact residual.
After any finite prefix, allow the continuation
\begin{equation}
y_{t+1}=y_t+\eta K_tR(y_t),\qquad
0<\kappa_-I\preceq K_t\preceq\kappa_+I,\qquad
\eta\kappa_+(N+1)<4,
\label{eq:capture-continuation}
\end{equation}
with fixed \(\eta>0\) and \(0<\kappa_-\le\kappa_+<\infty\).
If the finite state \(y_T\) satisfies
\begin{equation}
\mathcal E(y_T)>\tfrac12,
\label{eq:half-evidence-capture}
\end{equation}
then \(z=\operatorname{sign}(y_T)\) is well defined and belongs to \(\mathcal C\).
Every continuation obeying \eqref{eq:capture-continuation} preserves \(\operatorname{sign}(y_t)=z\) for \(t\ge T\), and
\[
r_t=\sum_i e^{-z_i y_{i,t}}=O((t-T+1)^{-1}).
\]
The Gram matrices \(K_t\) may depend arbitrarily on the training history.
Conversely, every trajectory approaching a code endpoint eventually satisfies \eqref{eq:half-evidence-capture}.
\end{theorem}

Along a continuous path, a hard label changes only when its logit crosses zero, where both labels are equally likely.
With the other labels fixed, at most one choice can complete a legal code, since legal codes differ in at least two coordinates.
Averaging over the other labels therefore gives \(\mathcal E\le1/2\) at every such tie.
These ties form an \emph{evidence barrier} between different hard codes, which extends directly to categorical concepts.
The constant \(1/2\) is optimal for a guarantee based on legal evidence alone: for every smaller threshold, an admissible update changes the labels of some state above it while the evidence increases.
Below \(1/2\), a nearly tied pair of coordinates carries only a weak restoring residual, so a stronger residual on another unsettled coordinate can couple through off-diagonal Gram entries to push both coordinates across zero.

Applying this argument to discrete updates requires more than an increase in evidence at successive iterates, since a step could jump across a label tie.
Under the step-size bound, evidence throughout each update segment stays at or above its starting value.
An update from \(\mathcal E>1/2\) therefore preserves the selected labels, and the spectral bounds ensure convergence to the corresponding endpoint.
In particular, an admissible Gram that selects another explanation from the initial state must preserve the captured code when introduced later.
This guarantee extends to controlled errors in posterior updates under the same spectral bounds and a tighter step-size bound.

Theorem~\ref{thm:routing-certificate} requires positive-definite sample Grams.
For a shared head on fixed features, \(K_t=XM_tX^\top\) is singular whenever \(X\) lacks full row rank.
The head then cannot choose each sample's prediction freely, because some predictions are determined by the others.
The following corollary establishes capture under this restriction, with spectral bounds on the parameter metrics.

\begin{corollary}[Capture with limited control over individual predictions]
\label{cor:fixed-linear-capture}
Under the candidate and native-update assumptions of Theorem~\ref{thm:routing-certificate}, let \(X\in\mathbb R^{N\times d}\) have arbitrary rank.
From a finite head \(w_T\in\mathbb R^d\), use fixed \(\eta>0\), constants \(0<m_-\le m_+<\infty\), and
\begin{equation}
y_t=Xw_t,\qquad w_{t+1}=w_t+\eta M_tX^\top R(Xw_t),
\qquad 0<m_-I_d\preceq M_t\preceq m_+I_d.
\label{eq:fixed-linear-capture}
\end{equation}
If \(\eta m_+\|X\|_{\mathrm{op}}^2(N+1)<4\), then \(\mathcal E(Xw_T)>1/2\) implies the same code preservation and \(O((t-T+1)^{-1})\) compactified rate for every history-dependent parameter-metric continuation.
\end{corollary}

For a shared head, the residuals from all samples enter the parameter gradient \(X^\top R\), where opposing contributions can cancel and halt training.
Capture prevents this cancellation: the head \(w_T\) already separates the selected labels, so moving along it increases all their signed margins.
Since subsequent residuals favor these labels, each sample contributes positively to the gradient along \(w_T\), keeping the parameter gradient nonzero.
The metric bounds then ensure continued progress toward the selected endpoint.
This guarantee includes ordinary full-batch parameter gradient descent (\(M_t=I\)), even when the selected labels are semantically incorrect.

\subsection{Identifiability does not ensure recovery}
\label{sec:recovery-subsec}

Once a training state's legal evidence exceeds \(1/2\), Section~\ref{sec:capture-subsec} certifies that the selected compatible explanation persists, whether its concept labels are right or wrong.
If a shared predictor rules out every wrong compatible explanation so that only the true concepts satisfy all constraints, must training recover them?
Proposition~\ref{prop:identifiable-minimum} shows that it need not: instead of approaching a compatible endpoint, training trajectories can converge to a finite, strictly suboptimal state whose hard predictions violate some constraints.

\begin{proposition}[A suboptimal attracting state]
\label{prop:identifiable-minimum}
Let \(\mathcal C_X=\{z\in\mathcal C:\exists w,\ \min_i z_i(Xw)_i>0\}\) denote the legal codes realizable by the shared predictor \(Xw\).
There exist \(X\) and a candidate set \(\mathcal C\) with pairwise Hamming distance at least two and \(\mathcal C_X=\{\one_N\}\), for which ordinary native updates \(w^+=w+\eta X^\top R(Xw)\) admit a finite, strictly suboptimal, locally asymptotically stable fixed point for every sufficiently small \(\eta>0\).
Its hard predictions violate some constraints, and \(\mathcal E(Xw)<1/2\).
\end{proposition}

To see how separate per-query abduction can sustain wrong predictions under a shared model, consider binary classification where each query reveals the label sum of a pair of inputs.
For inputs \(x_0\) and \(x_1\) with true labels \(0\) and \(1\), same-class pairs \((x_0,x_0)\) and \((x_1,x_1)\) have sums \(0\) and \(2\), forcing \((0,0)\) and \((1,1)\), whereas mixed pairs \((x_0,x_1)\) have sum \(1\) and permit both \((0,1)\) and \((1,0)\).
Because the shared predictor assigns one logit per input across every pair containing that input, only the true concept labels satisfy all observed sums.

Each mixed query, however, computes its exact abductive posterior from its own sum alone before the soft-target gradients from all queries are combined.
When current predictions lean toward reversed labels, every mixed pair favors \((1,0)\) in its posterior and reinforces the wrong predictions, while same-class pairs supply corrective targets.
As reversed predictions approach certainty, mixed-pair predictions nearly match \((1,0)\) and their reinforcement vanishes while same-class correction persists, ruling out an infinite wrong endpoint.
Yet when mixed-query reinforcement is strong enough relative to same-class correction, the two forces can balance at finite confidence and hold the shared predictor at a constraint-violating wrong state.
Section~\ref{sec:identifiable-counts} tests on CIFAR-10 whether standard random initializations actually converge to such wrong attractors under uniquely identifying counts.

\section{Empirical Evaluation}
\label{sec:experiments}

Fitting logical supervision need not recover the true concept labels.
On CIFAR-10, we test our predictions across fixed-feature count supervision, uniquely identifying pairwise counts, and end-to-end joint training with SGD and Adam.
Code is available on \href{https://github.com/WaTerminator/explanation-selection}{GitHub}.

\subsection{Different explanations under the same counts}
\label{sec:selection-experiments}

\begin{figure}[!t]
\centering
\includegraphics[width=\textwidth]{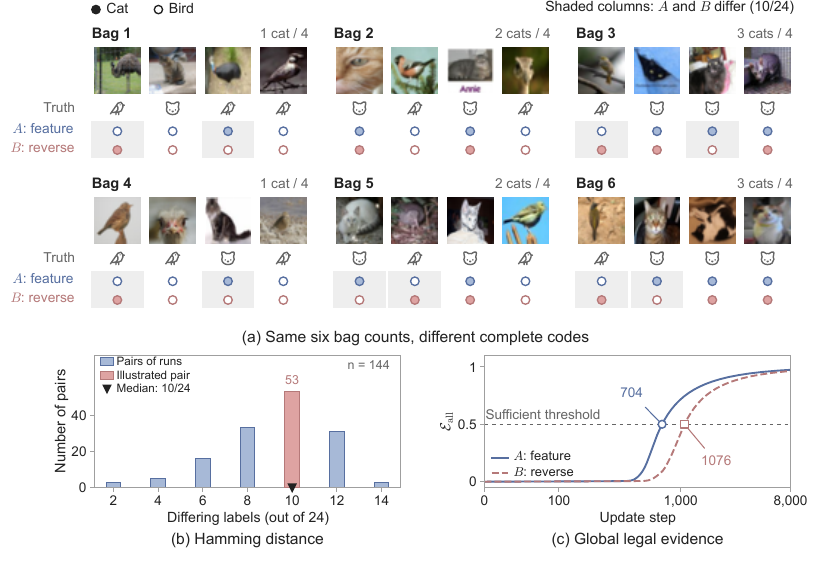}
\caption{Gram matrices with matched spectra select different captured assignments under the same CIFAR-10 bag counts.
(a) Cat/bird icons give ground truth, and filled/open circles mark cat/bird predictions in two learned codes from the same initial head.
Shaded columns mark the 10 labels on which the two codes differ.
(b) Number of differing training labels across all 144 pairs of runs sharing data and initialization but using different Gram orientations.
(c) Both runs cross the global evidence threshold \(1/2\), certifying capture, on a step axis that is linear through 100 and logarithmic thereafter.}
\label{fig:cifar-llp}
\end{figure}

We consider distinguishing birds from cats in CIFAR-10 images \citep{krizhevsky2009learning}, where training images are grouped into bags and only each bag's cat count is revealed \citep{barucic2022fast}.
Each training set contains 24 images partitioned into six disjoint four-image bags with cat counts \((1,2,3,1,2,3)\), admitting \(\bigl[\binom{4}{1}\binom{4}{2}\binom{4}{3}\bigr]^2=\text{9,216}\) compatible complete label assignments.
Every pair of compatible assignments differs in at least two labels, matching Theorem~\ref{cor:native-a3bl}, and the one-cat and three-cat bags exclude a global swap of the two classes.

To isolate update orientation from eigenvalue scale, we train a shared linear head on frozen ImageNet-supervised ResNet-18 features \citep{he2016deep}, refreshing exact posterior targets at every full-batch step as in Section~\ref{sec:dynamics}.
We compare four fixed positive-definite Grams constructed from input features, bag counts, and fixed random seeds without per-image labels: two align opposite eigenvalue orderings with the within-bag feature eigenbasis, and two use random orthogonal bases.
All four Grams share identical full and candidate-active spectra and use a common step size obeying the ascent condition.
Experimental details and additional results are provided in Appendices~\ref{app:cifar-llp-model} and~\ref{app:cifar-llp-results}.

We sample six disjoint training sets with four random initial heads per set, forming \(6\times4=24\) matched groups.
Training each initial head under all four Grams yields \(24\times4=96\) runs and \(24\times\binom{4}{2}=144\) within-group pairs that share the same images, counts, objective, and initialization while differing only in Gram eigenvector orientation.
All 96 runs fit every bag count with high confidence, yet in every group the four Grams select four distinct complete assignments, and 93 of the 96 assignments mislabel at least one training image, as illustrated for one initial head in Figure~\ref{fig:cifar-llp}(a).
Across the 144 matched pairs, Figure~\ref{fig:cifar-llp}(b) shows that the learned explanations differ on a median of 10 of the 24 training images.
On 2,000 unseen test images and 500 unseen four-image count questions, these paired models differ in test accuracy by a median of 11.2 percentage points on concept classification and 5.0 percentage points on count reasoning.

\subsection{Certifying the selected explanation in the same learner}
\label{sec:capture-experiments}

Section~\ref{sec:selection-experiments} shows that rotating Gram eigenvectors at step zero routes the same initial head to a rival explanation.
We now use the same 96 trajectories to test whether that intervention can still alter labels once capture is certified.
Across the six disjoint bags, global legal evidence factors as \(\mathcal E_{\rm all}(y)=\prod_{b=1}^{6}\mathcal E_b(y_b)\), where \(\mathcal E_b(y_b)\) is the probability of satisfying bag \(b\)'s count.
All 96 runs cross \(\mathcal E_{\rm all}>1/2\) during training, as illustrated for two matched trajectories in Figure~\ref{fig:cifar-llp}(c).
At each run's first crossing, we launch three admissible continuations: one swaps in an isospectral Gram that selected a different explanation from the same initial head at step zero, while the other two apply fixed or adaptive metrics that slow confidence growth on the weakest captured label.
Although the swapped Gram selects another explanation from initialization, after \(\mathcal E_{\rm all}>1/2\) all \(96\times3=288\) continuations preserve every captured label, locking in the concept errors in 93 of the 96 runs.

Capture also triggers well before high-confidence convergence, at a median 65\% of the updates needed to meet the fitting criterion across step sizes.
Searching backward along the 96 baseline trajectories for saved states from which an admissible continuation can still flip a label yields per-trajectory upper bounds on the delay between true irreversibility and \(\mathcal E_{\rm all}>1/2\), with a median upper bound of 26.6\% of the capture time.
Label preservation likewise holds without exception for a rank-deficient shared linear head with a singular sample Gram, matching Corollary~\ref{cor:fixed-linear-capture}.
Protocols and additional results appear in Appendices~\ref{app:cifar-llp-timing}--\ref{app:cifar-restricted}.

\subsection{Suboptimal attraction and reachability under identifiable counts}
\label{sec:identifiable-counts}

\begin{figure}[!t]
\centering
\includegraphics[width=\textwidth]{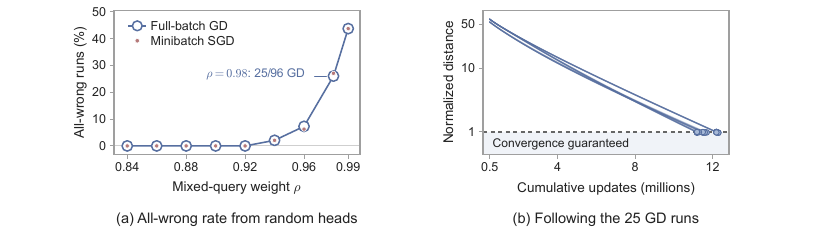}
\caption{Wrong attraction from standard random initializations despite uniquely identifying counts.
(a) Percentage of runs out of 96 (three 256-image sets \(\times\) 32 random heads) per optimizer and mixed-query weight \(\rho\) whose training concept predictions are all wrong after 500,000 updates, while every other run recovers all 256 labels.
(b) Continuing the 25 all-wrong full-batch GD runs from \(\rho=0.98\) under the unchanged objective and step size brings all 25 into the shaded region (normalized distance \(\le1\)), certifying convergence to the reversed-prediction state.
Distance is measured in effective parameter space in units of each set's certified radius, and curves end at their first recorded entry.}
\label{fig:identifiable-minima}
\end{figure}

Sections~\ref{sec:selection-experiments} and~\ref{sec:capture-experiments} study supervision with many compatible explanations.
We now test Proposition~\ref{prop:identifiable-minimum} when count constraints uniquely determine every training label under a shared predictor.
On balanced sets of 256 CIFAR-10 bird and cat images with full-row-rank frozen ResNet-18 features and a bias, each image participates in one same-class pair of count 0 or 2, which directly fixes both labels, and eight mixed-class pairs of count 1.
Only the true concept labels satisfy all counts, yet each mixed pair reinforces reversed labels whenever both of its shared inputs lean wrong.
We give the average loss of the mixed pairs weight \(\rho\) and the average loss of the same-class pairs weight \(1-\rho\).

We evaluate three disjoint 256-image sets with 32 shared label-free random linear heads per set, giving 96 runs for each optimizer and weight.
As shown in Figure~\ref{fig:identifiable-minima}(a), every run finishes with either all 256 training labels correct or all 256 reversed.
From the same 96 initial heads under full-batch GD, all 96 runs recover every label at \(\rho=0.90\), whereas 25 reverse every training concept prediction at \(\rho=0.98\), with minibatch SGD yielding nearly the same all-wrong proportions across the tested weights.
As mixed-pair feedback grows stronger relative to direct same-class supervision, standard random initializations increasingly settle into reversed predictions that violate every same-class count.

Continuing the 25 all-wrong full-batch GD runs at \(\rho=0.98\) past 500,000 updates under the unchanged objective and step size confirms that all 25 approach a finite wrong attractor.
Figure~\ref{fig:identifiable-minima}(b) shows that every trajectory enters a certified contracting neighborhood within which geometric convergence to the same reversed predictions on its training set is proved.
Appendices~\ref{app:wrong-basin-certificate} and~\ref{app:identifiable-experiments}--\ref{app:natural-initialization} derive this certificate and provide experimental details and sweep results.

\subsection{Explanation selection and test consequences in end-to-end training}
\label{sec:end-to-end}

\begin{figure}[!t]
\centering
\includegraphics[width=\textwidth]{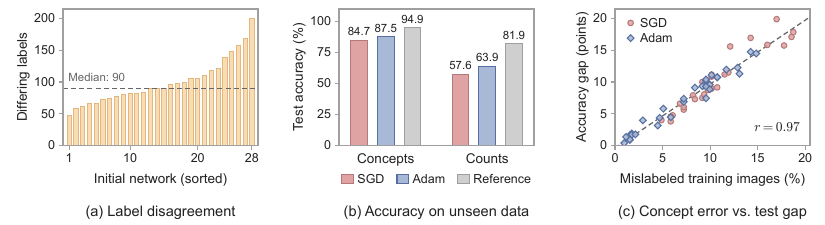}
\caption{End-to-end joint training with SGD and Adam under identical count supervision.
(a) Training labels, out of 1,024, on which SGD and Adam from the same initial network disagree across the 28 pairs where both satisfy all counts.
(b) Mean unseen test accuracy of the 28 SGD and 32 Adam count-satisfying models alongside 32 Adam references trained from the same initial networks on the true concept labels.
(c) Test concept accuracy gap relative to the paired reference increases linearly with the percentage of mislabeled training images across all 60 count-satisfying models.}
\label{fig:joint-training}
\end{figure}

Whereas Sections~\ref{sec:selection-experiments} and~\ref{sec:capture-experiments} control update orientation on frozen features through designed Grams, ordinary ABL trains the full network on minibatches with a standard optimizer.
We sample four disjoint sets of 1,024 CIFAR-10 bird and cat images partitioned into 256 four-image bags whose cat counts cycle through \(1,2,3\), and initialize eight random linear heads per set on an ImageNet-pretrained ResNet-18, giving 32 initial networks.
From each initial network, we train one model with minibatch SGD and one with Adam on identical bag minibatches using bag counts alone, and evaluate every model on 2,000 unseen test images and 500 unseen four-image count questions.
Training details and additional results, including four-class visual Sudoku, appear in Appendices~\ref{app:end-to-end-optimizers} and~\ref{app:sudoku}.

We first examine whether changing the standard optimizer from the same initial network alters the learned explanation and its unseen accuracy.
Across the 32 initial networks, Adam satisfies all 256 training counts in every run and SGD does so in 28 runs, while the four remaining SGD runs have lower test concept accuracy than every count-satisfying model.
In all 28 pairs where both optimizers satisfy the counts, SGD and Adam select different compatible explanations.
As shown in Figure~\ref{fig:joint-training}(a), the two optimizers disagree on a median of 90 of the 1,024 training labels.
On unseen data, they differ in accuracy by a median of 4.1 percentage points on concepts and 7.5 points on counts.

Because both optimizers satisfy the bag counts while mislabeling concepts, comparing SGD with Adam alone does not reveal how much unseen accuracy is lost relative to learning the true explanation.
We therefore compare these 60 count-satisfying models, comprising 28 SGD and 32 Adam runs, against 32 reference models trained with Adam from the same initial networks on the true training concept labels under the same Adam learning rates, schedule, minibatches, and budget.
Every reference model learns all training concepts and outperforms every count-satisfying model from the same initial network on both unseen concepts and unseen counts, with group means compared in Figure~\ref{fig:joint-training}(b).
Across all 60 count-satisfying models, Figure~\ref{fig:joint-training}(c) shows that the unseen concept accuracy gap relative to each model's paired reference grows nearly linearly with its percentage of mislabeled training images, with Pearson correlation \(r=0.97\) and least-squares slope \(1.01\).

\FloatBarrier
\section{Conclusion}
\label{sec:conclusion}

{\setlength{\parfillskip}{0pt}
This work shifts the study of ambiguity in abductive learning from static compatibility to training dynamics, showing how iterative feedback between perception and reasoning selects and entrenches concept semantics.
Our analysis demonstrates that optimization geometry governs which compatible explanation training selects from a shared initialization, while crossing half legal evidence can lock in the chosen explanation along with its concept errors.
Excluding incorrect compatible explanations through a shared predictor still does not guarantee recovery, because abductive feedback computed on individual queries can reinforce shared prediction errors and trap training at a finite, constraint-violating state.
In end-to-end training on identical count minibatches, SGD and Adam likewise select different compatible explanations from the same initial networks, and models with more training concept errors suffer proportionally larger unseen concept accuracy gaps relative to true-label references.
A limitation of our theoretical analysis is that it relies on fixed features and exact full-batch updates.
In future work, we plan to extend this dynamical perspective to learned representations, stochastic updates, and approximate abduction, and to develop training mechanisms that prevent premature commitment to wrong concepts.
\par}

\clearpage

\section*{AI-Use Disclosure}
We used generative AI tools as assistants for theoretical analysis, experimental work, and manuscript preparation.
Their assistance included literature search, formulation and refinement of mathematical claims, development and checking of proof arguments, experimental design and protocol refinement, construction of synthetic test cases, implementation and debugging of research and verification code, and interpretation of numerical results.
These tools also assisted with manuscript polishing and the preparation of figures and tables.
The authors evaluated the suggestions, reviewed mathematical arguments against their stated assumptions, and checked reported numerical results against experimental records.
The authors made all final decisions on the research methodology, claims, and presentation and take full responsibility for the paper and its accompanying artifacts.

\section*{Reproducibility Statement}
The model and assumptions are specified in Sections~\ref{sec:dynamics}--\ref{sec:capture}, with full proofs provided in Appendices~\ref{app:roadmap}--\ref{app:identifiable-recovery}.
Appendix~\ref{app:experiments} specifies the data construction, feature preprocessing, model architectures, step-size selection rules, and evaluation procedures for all experiments in Section~\ref{sec:experiments}.

\bibliography{references}
\bibliographystyle{iclr2027_conference}

\clearpage
\appendix
\makeatletter
\def\fps@figure{!htbp}
\def\fps@table{!htbp}
\makeatother
\setlength{\parskip}{4.5pt plus 1pt minus 1pt}
\setlength{\intextsep}{11pt plus 2pt minus 1pt}
\setlength{\textfloatsep}{12pt plus 2pt minus 2pt}
\setlength{\floatsep}{10pt plus 2pt minus 2pt}
\captionsetup[table]{position=top,aboveskip=3pt,belowskip=6pt}
\captionsetup[figure]{position=bottom,aboveskip=6pt,belowskip=3pt}
\setlength{\abovedisplayskip}{5pt plus 1.5pt minus 1.5pt}
\setlength{\belowdisplayskip}{5pt plus 1.5pt minus 1.5pt}
\setlength{\belowdisplayshortskip}{4.5pt plus 1.5pt minus 1.5pt}
\section{Related Work}
\label{app:related-work}

\paragraph{Reasoning and label revision in ABL.}
ABL uses logical knowledge to revise perceptual predictions and guide further learning through a cycle of perception, abduction, and label revision \citep{zhou2019abductive,dai2019bridging}.
Methods refine this cycle through candidate weighting \citep{he2024ambiguity}, curricula \citep{wei2025curriculum,hu2025curriculum}, and probabilistic symbol perception \citep{jia2025smooth}.
Abductive Reflection learns to locate inconsistent predictions before symbolic repair \citep{hu2025reflection}.
Consistency-distance analysis relates scoring and pseudo-label sampling to generalization risk and shows how abductive consistency can reduce it \citep{yang2024analysis}.
These methods change how training obtains or uses labels.
We fix the complete-code posterior rule and analyze how Gram orientation affects selection among compatible explanations on a common evidence potential.

\paragraph{Identifiability, learnability and shortcut risk.}
Identifiability and constraint-based analyses characterize which concept interpretations the knowledge and observations can distinguish \citep{marconato2023notall,bortolotti2025identifiability,takemura2026constraint}.
Learnability results give conditions for recovering latent labels from multi-instance supervision or logical rules \citep{wang2023multiinstance,tao2024deciphering,he2025learnability,jia2025verification}.
Risk analyses show how the symbolic component can create unequal class-specific errors \citep{tsamoura2025imbalances}.
The distinction between symbolic ambiguity, architectural realizability and optimization selection is explicit in \citet[Sections~6.3--6.4]{xu2026reasoning}, which also studies how absorbing arithmetic values suppress correction gradients.
When multiple explanations remain compatible with supervision, we ask which attract nearby training states and when their labels become captured under fixed logical constraints and exact posterior updates.

\paragraph{Learning from label proportions.}
Small-bag count supervision uses a posterior over all binary assignments consistent with each count \citep{barucic2022fast}.
Its exact EM formulation alternates product-posterior normalization with classifier training against marginal label probabilities, including experiments on CIFAR-10 bird/cat images.
We use this likelihood and posterior training method with a fixed pretrained representation, refreshing the posterior after every shared-head update.
For disjoint bags, global evidence factors into bag evidences, connecting the count likelihood to our selection and capture analysis under sample-space or fixed-linear parameter updates.

\paragraph{Semantic loss and probabilistic neuro-symbolic learning.}
In the native model, \(\Phi=2\log\mathcal E\) is twice the negative semantic loss of \citet{xu2018semantic}, where \(\mathcal E\) is the probability that independent Bernoulli predictions satisfy the logical constraints.
This probability is a weighted model count, connecting our objective to probabilistic neuro-symbolic learning \citep{manhaeve2018deepproblog}.
The posterior refresh connects this gradient update to latent-variable evidence optimization and generalized EM \citep{dempster1977maximum}.
Exact normalization has computational costs characterized by hardness results for probabilistic neuro-symbolic learning \citep{maene2024hardness}.
Deterministic approximate-EM analysis gives convergence conditions for controlled E-step surrogates \citep{lartigue2022deterministic}.
Our error bounds on posterior approximation are relative to the exact residual, yielding quantitative ascent and preserving local stability and sample-SPD capture (Appendices~\ref{app:inexact} and~\ref{app:inexact-capture}).

\citet[Theorems~4.3 and~4.10]{vankrieken2024independence} characterize the zero-loss independent distributions as a union of implicant cubes, and their Theorem~4.13 describes its connected components: in our native regime, a valid one-bit neighbor creates a flat zero-loss edge, whereas a vertex without one is isolated in the zero-loss set.
Building on this static geometry, we characterize discrete trapping neighborhoods and \(O(1/t)\) convergence under general priors, temperature, and controlled posterior-refresh errors (Appendices~\ref{app:local-scope} and~\ref{app:inexact}).

\paragraph{Ambiguous-target optimization.}
Softmax regression with partial-label likelihood exhibits winner-take-all behavior among initially maximal acceptable labels \citep[Theorem~4]{zombori2024unbiased}, while changing the loss can preserve their probability ratios.
Self-training analyses study confidence thresholds and fixed points \citep{oymak2021statistical}, generalization from unlabeled data \citep{zhang2022unlabeled}, and spurious-feature removal under domain shift \citep{chen2020selftraining}.
Test-time adaptation theory studies gradient descent with hard and conjugate pseudo-labels \citep{wang2023gd}.
With a complete-code posterior, we derive stability criteria involving defect overlap and temperature and construct isospectral Grams that select different endpoints from a common initialization.

\paragraph{Implicit bias and optimizer geometry.}
Implicit-bias theory asks which solution gradient methods select and how optimizer geometry affects that selection \citep{soudry2018implicit,gunasekar2018geometry,ravi2024multiclass,wu2023edge}.
Metric-dependent basins for a fixed potential are classical \citep{fiedler1999basins}, and selection of different minimizers with a fixed spectrum follows from optimization-geometry theory \citep[Theorem~1]{gunasekar2018geometry}.
We establish this dependence for discrete legal explanations of a common evidence potential from an open initialization neighborhood under matched full and candidate-active spectra.

\paragraph{Regions of attraction, certificates, and continuous constraint solving.}
Lyapunov sublevel sets and optimization-based searches can certify regions of attraction for nonlinear discrete systems \citep{leisenberger2021convergence}, while bounded variable metrics and controlled inexactness are standard tools for descent and stationarity \citep{bonettini2016variable}.
Local basin stability for preconditioned SGD around minimizer sets uses a fixed positive-definite parameter metric, local Polyak--\L{}ojasiewicz and quadratic-growth conditions, and stochastic-gradient bounds up to a finite horizon \citep{scott2026design}.
In our exact binary setting, the static bound \(\mathcal E\le1/2\) on every hard-sign boundary combines with segment ascent to certify permanent label preservation and \(O(1/t)\) endpoint convergence at positive loss under arbitrary history-dependent admissible metric sequences.
Relatedly, FourierCSP minimizes expected constraint violations over probability simplices under fixed projected or mirror-descent rules \citep{cen2026fouriercsp}, whereas our updates act on logits or shared linear parameters to maximize log joint legality and certify invariance of a specific complete code across changing metrics.

\clearpage
\section{Extended Model, Proof Roadmap, and Common Tools}
\label{app:roadmap}

\subsection{Proof locations and principal dependencies}

Table~\ref{tab:proof-dependencies} maps the theoretical results in the main text and appendices to their proof locations and main intermediate steps.

\begin{table}[!htbp]
\centering
\caption{Proof locations and principal dependencies.}
\label{tab:proof-dependencies}
\small
\begin{tabular}{@{}>{\raggedright\arraybackslash}p{0.27\textwidth}>{\raggedright\arraybackslash}p{0.45\textwidth}>{\raggedright\arraybackslash}p{0.20\textwidth}@{}}
\toprule
Result & Main dependency & Proof \\
\midrule
Local classifications (Theorem~\ref{cor:native-a3bl}, Proposition~\ref{cor:local-iff}) & Anchored factorization, gap bounds, discrete trapping & Appendix~\ref{app:local-classifications} \\
\addlinespace[3pt]
Isospectral routing (Theorem~\ref{thm:arbitrary-many-isospectral}) & Slow--fast construction, absorbing cone, robust continuation & Appendix~\ref{app:routing-main-proof} \\
\addlinespace[3pt]
Tied-prediction selection (Corollary~\ref{cor:rank-deficient-selection}) & Lift of parameter metrics to the routing Grams & Appendix~\ref{app:rank-deficient-selection} \\
\addlinespace[3pt]
Half-evidence capture (Theorem~\ref{thm:routing-certificate}) & Conditional slices, segment ascent, global gap bounds & Appendix~\ref{app:capture}, p.~\pageref{proof:half-evidence-capture} \\
\addlinespace[3pt]
Optimal capture threshold (Proposition~\ref{prop:half-evidence-sharpness}) & Explicit three-concept Gram-coupled update & Appendix~\ref{app:half-evidence-sharpness} \\
\addlinespace[3pt]
Shared-head and categorical barrier (Corollary~\ref{cor:fixed-linear-capture}, Lemma~\ref{lem:categorical-barrier}) & Separation by the captured head; disjoint categorical completions & Appendices~\ref{app:fixed-linear-capture} and~\ref{app:multiclass-capture} \\
\addlinespace[3pt]
Suboptimal attraction (Propositions~\ref{prop:identifiable-minimum} and~\ref{prop:regular-query-minimum}) & Stationary reversal branch, graph-Laplacian Hessian, occurrence lift & Appendix~\ref{app:identifiable-minima} \\
\addlinespace[3pt]
Regular pairwise counts and safe restoration (Proposition~\ref{prop:global-recovery}, Corollary~\ref{cor:safe-restoration}) & Least-correct coordinate gradient bound, segment loss barrier & Appendix~\ref{app:global-recovery} \\
\bottomrule
\end{tabular}
\end{table}

\subsection{Parameterized posterior and direct supervision}
\label{app:general-model}

The native update of Section~\ref{sec:dynamics} uses uniform priors, unit temperature, and no direct supervision.
The local theory also allows positive priors \(\pi_z\), temperature \(\tau>0\), and direct semantic supervision of relative weight \(\lambda\ge0\).
The extended posterior is
\begin{equation}
q_z(y)=\frac{[\pi_z\prod_i\sigmoid(z_i y_i)]^{1/\tau}}
{\sum_{w\in\mathcal C}[\pi_w\prod_i\sigmoid(w_i y_i)]^{1/\tau}},
\qquad p_i(y)=\Pr_q(z_i=+1),
\label{eq:general-posterior}
\end{equation}
and a detached, full-batch cross-entropy step gives
\begin{equation}
y^+=y+\eta KR(y),\qquad
R_i(y)=2[p_i(y)+\lambda-(1+\lambda)\sigmoid(y_i)].
\label{eq:general-map}
\end{equation}
Its potential is
\begin{equation}
\Phi_{\mathcal C,\lambda}(y)
=2\tau\log\!\sum_{z\in\mathcal C}\pi_z^{1/\tau}
\prod_i\sigmoid(z_i y_i)^{1/\tau}
+2\lambda\sum_i\log\sigmoid(y_i),
\label{eq:general-potential}
\end{equation}
with \(\nabla\Phi_{\mathcal C,\lambda}=R\) and \(\Phi_{\mathcal C,\lambda}\le2\tau\log\sum_z\pi_z^{1/\tau}\).
For normalized uniform priors, \(\tau=1\), and \(\lambda=0\), \(\Phi_{\mathcal C,0}=2\log\mathcal E-2\log|\mathcal C|\), differing from~\eqref{eq:potential} by an additive constant and inducing the same update.
In the parameterized local-stability and posterior-error analyses, \(\Phi\) abbreviates \(\Phi_{\mathcal C,\lambda}\), and potential deficits relative to a code corner are written \(V_z(a)=\Phi(0)-\Phi(a)\).
Taking logarithms in~\eqref{eq:monomial-odds} shows that \(\log(q_w(y)/q_z(y))=[\log(\pi_w/\pi_z)-\sum_{i\in D(w,z)}z_i y_i]/\tau\), so \(q\) is a softmax over affine defect-set log-odds.
For any \(K\succ0\), finite critical points of~\eqref{eq:general-map} are independent of \(K\) since \(KR=0\iff R=0\), and strict ascent whenever \(R(y)\ne0\) holds under the sufficient step-size condition
\begin{equation}
\eta\lambda_{\max}(K)
\left(\frac N\tau+1+\lambda\right)<4,
\label{eq:general-ascent-condition}
\end{equation}
established in Lemmas~\ref{lem:global-smoothness} and~\ref{lem:ascent-inequality} below.
Throughout the proofs, we write \(K_z=D_z K D_z\) with \(D_z=\operatorname{diag}(z)\), \(F_K(y)=y+\eta KR(y)\) for the one-step map, and \(\lVert v\rVert_K^2=v^\top Kv\) for \(K\succ0\).

\subsection{Common analytic tools}

Splitting off the semantic factor gives the equivalent potential
\begin{equation*}
\Phi_{\mathcal C,0}(y)
=2\sum_i\log\sigmoid(y_i)
+2\tau\log\!\!\sum_{z\in\mathcal C}\pi_z^{1/\tau}
e^{-\langle\one_{D(z,\one_N)},y\rangle/\tau}.
\end{equation*}
The first term is separable and strictly concave.
The second is a convex log-sum-exp of the defect functionals, with weights determined by the candidate priors and temperature.

\begin{lemma}[Hessian bound and global smoothness]
\label{lem:global-smoothness}
The potential \(\Phi : \mathbb R^N \to \mathbb R\) is twice continuously differentiable with gradient \(\nabla\Phi(y) = R(y)\).
Its Hessian satisfies
\[
\nabla^2 \Phi(y)
=
\frac{2}{\tau} \operatorname{Cov}_{q(y)}[b(z)]
-
2(1 + \lambda) \operatorname{diag}(\sigma'(y)),
\]
where \(b(z) = (z + \mathbf{1})/2 \in \{0,1\}^N\) and \(q(y)\) is the complete-code posterior.
Consequently, for any \(y \in \mathbb R^N\),
\[
\|\nabla^2 \Phi(y)\|_{\mathrm{op}}
\le
L_\Phi
:=
\frac12\left(\frac{N}{\tau} + 1 + \lambda\right).
\]
The gradient \(\nabla\Phi\) is globally Lipschitz with constant \(L_\Phi\).
\end{lemma}
\begin{proof}
For the unnormalized posterior weight in~\eqref{eq:general-posterior}, the \(y_i\)-derivative of its log is \((b_i(z)-\sigma(y_i))/\tau\).
Softmax differentiation therefore gives \(\nabla p(y)=\operatorname{Cov}_{q(y)}[b(z)]/\tau\).
Differentiating~\eqref{eq:general-potential} gives \(\nabla\Phi=R\), and differentiating \(R(y)=2[p(y)+\lambda\mathbf1-(1+\lambda)\sigma(y)]\) gives the stated Hessian.
The covariance term has operator norm at most \(N/4\) (uniform bound over all distributions on \(\{0,1\}^N\)), and \(\sigma'(t) \le 1/4\) uniformly.
The Lipschitz property follows from the mean-value theorem.
\end{proof}

\begin{lemma}[Ascent inequality]
\label{lem:ascent-inequality}
Let \(K \succ 0\) and \(y \in \mathbb R^N\) with \(R(y) \ne 0\).
Then
\[
\Phi(F_K(y)) - \Phi(y)
\ge
\eta R(y)^\top K R(y) - \frac{\eta^2 L_\Phi}{2} \|K R(y)\|^2.
\]
Consequently,
\[
\Phi(F_K(y)) - \Phi(y)
\ge
\eta\left(1 - \frac{\eta L_\Phi \lambda_{\max}(K)}{2}\right) R(y)^\top K R(y).
\]
If the step condition
\[
\eta \lambda_{\max}(K) \left(\frac{N}{\tau} + 1 + \lambda\right) < 4
\]
holds, the coefficient in the preceding display is positive.
Consequently, \(\Phi(F_K(y))>\Phi(y)\) whenever \(R(y)\ne0\).
\end{lemma}
\begin{proof}
Set \(\Delta=\eta K R(y)\).
The lower smoothness inequality from Lemma~\ref{lem:global-smoothness} gives
\[
\Phi(F_K(y)) - \Phi(y)
\ge
\eta R(y)^\top K R(y) - \frac{\eta^2 L_\Phi}{2} \|K R(y)\|^2.
\]
Because \(K\succ0\), spectral calculus gives \(K^2\preceq\lambda_{\max}(K)K\), and hence
\[
\|KR(y)\|^2
=R(y)^\top K^2R(y)
\le\lambda_{\max}(K)R(y)^\top KR(y).
\]
Substitution gives the second inequality.
Since \(L_\Phi=\tfrac12(N/\tau+1+\lambda)\), the step condition is equivalent to \(\eta L_\Phi\lambda_{\max}(K)<2\), making the coefficient \(1-\eta L_\Phi\lambda_{\max}(K)/2\) positive.
Positive definiteness also gives \(R(y)^\top KR(y)>0\) whenever \(R(y)\ne0\), proving strict ascent at every nonstationary state.
\end{proof}

\begin{lemma}[SPD spectral neighborhood]
\label{lem:spd-neighborhood}
Let \(K_0 \succ 0\) with eigenvalues \(\lambda_{\min}(K_0) > 0\) and \(\lambda_{\max}(K_0)\).
For any \(0 < \delta < \tfrac12 \lambda_{\min}(K_0)\), the set
\[
\mathcal V_\delta
=
\{K \succ 0 : \|K - K_0\|_F < \delta\}
\]
is an open neighborhood of \(K_0\) in the cone of SPD matrices, and every \(K \in \mathcal V_\delta\) satisfies
\[
\lambda_{\min}(K) \ge \tfrac12 \lambda_{\min}(K_0),
\qquad
\lambda_{\max}(K) \le 2 \lambda_{\max}(K_0).
\]
\end{lemma}
\begin{proof}
Since \(\|K - K_0\|_{\mathrm{op}} \le \|K - K_0\|_F < \delta\), the Weyl inequality gives
\[
\lambda_{\min}(K) \ge \lambda_{\min}(K_0) - \delta > \lambda_{\min}(K_0) - \tfrac12 \lambda_{\min}(K_0) = \tfrac12 \lambda_{\min}(K_0),
\]
and similarly
\[
\lambda_{\max}(K) \le \lambda_{\max}(K_0) + \delta < \lambda_{\max}(K_0) + \tfrac12 \lambda_{\min}(K_0) \le \tfrac32 \lambda_{\max}(K_0) < 2\lambda_{\max}(K_0).
\]
\end{proof}

\begin{lemma}[Finite-horizon continuity]
\label{lem:finite-horizon-continuity}
Fix \(T \in \mathbb N\), a compact set \(\mathcal Y_0 \subset \mathbb R^N\), and a compact set \(\mathcal K \subset \{K \succ 0\}\) in the SPD cone.
Then the map
\[
(y_0, K) \mapsto F_K^T(y_0)
\]
is uniformly continuous on \(\mathcal Y_0 \times \mathcal K\).
\end{lemma}
\begin{proof}
The residual \(R\) is continuous, so \((K,y)\mapsto F_K(y)=y+\eta KR(y)\) is jointly continuous.
Induction on \(T\) shows that every finite iterate \((K,y_0)\mapsto F_K^T(y_0)\) is jointly continuous.
Its domain \(\mathcal Y_0\times\mathcal K\) is compact, so the Heine--Cantor theorem makes this map uniformly continuous there.
\end{proof}

\section{Local Endpoint Stability}
\label{app:local-stability}

We prove the local classifications in Theorem~\ref{cor:native-a3bl} and Proposition~\ref{cor:local-iff} by expanding the parameterized potential~\eqref{eq:general-potential} near a candidate corner and controlling the discrete updates there.
After clarifying the role of boundary faces, we derive these bounds, extend them to posterior errors, and examine the local predictions numerically at finite radii along dense and near-face directions.

A corner is called \emph{potential-viable} when it is a strict local maximum of this potential on the closed compactified orthant.
With direct semantic supervision \(\lambda>0\), the anchor is the semantic code \(z=\one_N\).

\subsection{Closed faces and finite-logit reachability}
\label{app:local-scope}

In the native model, let \(\mathcal C=\{(+,+),(-,+)\}\) and take \(K_c=\left(\begin{smallmatrix}1&c\\c&1\end{smallmatrix}\right)\), \(0<c<1\), and \(3\eta(1+c)<4\).
Now \(R_1=0\) and \(R_2=2\sigmoid(-y_2)\).
Thus \(y_2\) increases without a finite limit, while \(y_1-cy_2\) is invariant: every finite initialization reaches \((+,+)\).
Both compactified coordinates decrease near that corner, yet the face \(a_2=0,a_1>0\) is stationary.
Theorem~\ref{cor:native-a3bl} therefore includes saturated-coordinate faces in its attraction criterion.
Replacing \(c\) by \(-c\) preserves the spectrum and sends every finite initialization to \((-,+)\).

\subsection{The anchored factorization}

Fix \(z\in\mathcal C\) and use the compactified coordinates \(a_i=e^{-z_iy_i}\in(0,\infty)\) of \eqref{eq:compactified-coords}.
For each coordinate,
\begin{equation*}
\sigmoid(z_iy_i)=\frac{1}{1+a_i},
\qquad
\frac{\sigmoid(-z_iy_i)}{\sigmoid(z_iy_i)}=a_i,
\end{equation*}
where the ratio follows from \(\sigmoid(-t)/\sigmoid(t)=e^{-t}\).
For \(w\in\mathcal C\), coordinates on which \(w\) agrees with \(z\) contribute \(\sigmoid(z_iy_i)\), whereas each defect coordinate contributes \(\sigmoid(-z_iy_i)\).
Hence
\begin{equation}
\prod_i\sigmoid(w_iy_i)
=\left[\prod_i\sigmoid(z_iy_i)\right]\prod_{i\in D(w,z)}a_i
=\frac{1}{\prod_i(1+a_i)}\prod_{i\in D(w,z)}a_i.
\label{eq:weight-factor}
\end{equation}
Writing \(W_w=\pi_w^{1/\tau}\prod_i\sigmoid(w_iy_i)^{1/\tau}\) for the unnormalized weight, \eqref{eq:weight-factor} gives the ratio
\begin{equation}
\frac{W_w}{W_z}
=\left(\frac{\pi_w}{\pi_z}\right)^{1/\tau}
\prod_{i\in D(w,z)}a_i^{1/\tau},
\label{eq:relative-weight}
\end{equation}
which is \eqref{eq:monomial-odds}: a monomial in \(a\) of total degree \(d_{\rm H}(w,z)/\tau\), with \emph{positive} exponents.
Factoring \(W_z\) out of the log-sum in \eqref{eq:general-potential} therefore yields, with no approximation and for every \(a\in(0,\infty)^N\),
\begin{equation}
\Phi_{\mathcal C,0}(a)
=2\log\pi_z
-2\sum_i\log(1+a_i)
+2\tau\log\Bigl[1+S_z(a)\Bigr],
\qquad
S_z(a)=\sum_{w\ne z}\frac{W_w}{W_z}.
\label{eq:exact-factorization}
\end{equation}
The anchor factor contributes \(-2\sum_i\log(1+a_i)\), while \(S_z\) collects the rival weights relative to the anchor.
Writing \(a=ru\) with \(r=\sum_ia_i\) and \(u\in\Delta^{N-1}\),
\begin{equation}
S_z(ru)=\sum_{w\ne z}\left(\frac{\pi_w}{\pi_z}\right)^{1/\tau}
r^{d_{\rm H}(w,z)/\tau}\prod_{i\in D(w,z)}u_i^{1/\tau}.
\label{eq:S-in-polar}
\end{equation}
The monomials in \eqref{eq:S-in-polar} have positive exponents, so they remain continuous when some \(u_i=0\).
Together with \eqref{eq:exact-factorization}, this gives a continuous potential on the closed compactified orthant near the origin, with corner value \(\Phi_{\mathcal C,0}(0)=2\log\pi_z\).
The coefficient bounds in the expansion below hold uniformly over the entire closed simplex, including its boundary faces.

\begin{remark}[Where direct semantic supervision may be added]
\label{rem:support-anchor}
The direct semantic supervision term of \eqref{eq:general-potential} is \(2\lambda\sum_i\log\sigmoid(y_i)\), written in the \emph{semantic} margins \(y_i\), not in the anchored margins \(z_iy_i\).
At the semantic anchor \(z=\one_N\) the two agree, \(\log\sigmoid(y_i)=-\log(1+a_i)\), and supervision rescales the first sum in \eqref{eq:exact-factorization} to \(-2(1+\lambda)\sum_i\log(1+a_i)\).
At any other anchor there is a coordinate with \(z_i=-1\), along which \(y_i\to-\infty\) as the corner is approached, so \(\log\sigmoid(y_i)\to-\infty\) and \(\Phi_{\mathcal C,\lambda}\) has no finite value at that corner.
We therefore restrict \(\lambda>0\) to \(z=\one_N\) and set \(\lambda=0\) otherwise.
\end{remark}

\subsection{Local maximality is necessary}

\begin{lemma}[Viability is necessary]
\label{prop:viability-necessary}
Under the ascent-step condition \eqref{eq:general-ascent-condition}, if a corner \(a=0\) is an asymptotically stable endpoint of \eqref{eq:general-map}, then it is a strict local maximum of \(\Phi_{\mathcal C,\lambda}\).
\end{lemma}

\begin{proof}[Proof of Lemma~\ref{prop:viability-necessary}]
The monomial ratios in~\eqref{eq:relative-weight} make the posterior and residual continuous on a neighborhood of the origin in the closed compactified orthant.
The update extends there as
\begin{equation}
T_{z,K}(a)_i
=a_i\exp\!\left(-\eta z_i[KR(a)]_i\right),
\label{eq:compactified-map-extension}
\end{equation}
with \(T_{z,K}(0)=0\).
The factorization \eqref{eq:exact-factorization} gives the corresponding continuous extension of \(\Phi\).

Suppose that the origin is asymptotically stable for this compactified map.
Then some neighborhood \(\mathcal N\) of the origin is attracted to it.
We claim that every \(a\in\mathcal N\setminus\{0\}\) satisfies \(\Phi(a)<\Phi(0)\).
If \(R(a)=0\), then \eqref{eq:compactified-map-extension} fixes \(a\), contradicting attraction.
If \(R(a)\ne0\), the quantitative ascent inequality from Lemma~\ref{lem:ascent-inequality}, extended to the closed orthant by continuity, gives
\begin{equation*}
\Phi\!\left(T_{z,K}(a)\right)-\Phi(a)
\ge
\eta\left(1-\frac{\eta L_\Phi\lambda_{\max}(K)}2\right)
R(a)^\top K R(a)>0.
\end{equation*}
If \(\Phi(a)\ge\Phi(0)\), this strict increase and the subsequent nondecrease of \(\Phi\) contradict \(T_{z,K}^t(a)\to0\) and continuity of the potential.
Hence \(\Phi(a)<\Phi(0)\) throughout a punctured neighborhood of the origin in the closed orthant, establishing strict local maximality even along its boundary faces.
\end{proof}

\subsection{Expansion and the three regimes}
\label{app:uniform-expansion-proof}

\begin{lemma}[Uniform boundary expansion]
\label{lem:uniform-expansion}
Let priors be positive, write \(\alpha=d_{\min}(z)/\tau\), and take \(\lambda=0\) unless \(z=\one_N\) is the semantic code, in which case any \(\lambda\ge0\) is allowed.
Then, uniformly over \(u\in\Delta^{N-1}\) including its faces,
\begin{equation*}
\Phi_{\mathcal C,\lambda}(ru)-\Phi_{\mathcal C,\lambda}(0)
=-2(1+\lambda)\,r
+2\tau\!\!\sum_{w\ne z}\left(\frac{\pi_w}{\pi_z}\right)^{1/\tau}\!\!
r^{d_{\rm H}(w,z)/\tau}\prod_{i\in D(w,z)}u_i^{1/\tau}
+O\!\left(r^{\min\{2,\,2\alpha\}}\right),
\end{equation*}
where the constant depends on \(N\), \(\tau\), \(\lambda\), \(|\mathcal C|\) and the prior ratios \((\pi_w/\pi_z)^{1/\tau}\), but not on \(u\) or \(r\).
\end{lemma}

\begin{proof}[Proof of Lemma~\ref{lem:uniform-expansion}]
Put \(\alpha=d_{\min}(z)/\tau\) and let \(C_\pi=\sum_{w\ne z}(\pi_w/\pi_z)^{1/\tau}\).
Since \(u_i\le1\) and every exponent in \eqref{eq:S-in-polar} is at least \(\alpha\), for \(r\le1\)
\begin{equation}
0\le S_z(ru)\le C_\pi\,r^{\alpha}
\qquad\text{uniformly in }u\in\Delta^{N-1}.
\label{eq:S-bound}
\end{equation}
Apply \(\log(1+x)=x+O(x^2)\) to \eqref{eq:exact-factorization} and \(\log(1+a_i)=a_i+O(a_i^2)\) to the first sum, using \(\sum_ia_i=r\) and \(\sum_ia_i^2\le r^2\):
\begin{equation}
\Phi_{\mathcal C,\lambda}(ru)-\Phi_{\mathcal C,\lambda}(0)
=-2(1+\lambda)\,r+2\tau\,S_z(ru)
+O\!\left(r^{\min\{2,\,2\alpha\}}\right),
\label{eq:expansion-correct}
\end{equation}
where \(\lambda=0\) unless \(z=\one_N\) (Remark~\ref{rem:support-anchor}), and the \(O\) constant depends on \(N\), \(\tau\), \(\lambda\), \(|\mathcal C|\) and \(C_\pi\) (that is, on the prior ratios), but not on \(u\) or \(r\).
The two remainder sources are the quadratic term of the first logarithm, which is \(O(r^2)\), and that of the outer logarithm, which by \eqref{eq:S-bound} is \(O(S_z^2)=O(r^{2\alpha})\).
For \(\alpha<1\), the latter is the larger bound, giving the remainder \(O(r^{\min\{2,2\alpha\}})\).
\end{proof}

\subsection{Distance, temperature, and the critical coefficient}
\label{app:viability-proof}

\begin{theorem}[Candidate-geometry viability criterion]
\label{thm:viability}
Let \(\lambda=0\) and the priors be positive.
If \(d_{\min}(z)>\tau\), then \(z\) is potential-viable.
If \(d_{\min}(z)<\tau\), it is not.
At \(d_{\min}(z)=\tau\), the functional \(H_z\) of~\eqref{eq:H-functional} gives, uniformly over \(u\in\Delta^{N-1}\),
\[
\Phi(ru)-\Phi(0)=2r[H_z(u)-1]+o(r).
\]
Thus \(\chi_z<1\) implies viability and \(\chi_z>1\) excludes it.
At \(\chi_z=1\), this test is inconclusive.
\end{theorem}

\begin{proof}
Take \(\lambda=0\) throughout, so \eqref{eq:expansion-correct} reads \(\Phi(ru)-\Phi(0)=-2r+2\tau S_z(ru)+O(r^{\min\{2,2\alpha\}})\).

\emph{(i) \(\alpha>1\).}
By \eqref{eq:S-bound}, \(2\tau S_z(ru)\le2\tau C_\pi r^{\alpha}=o(r)\) uniformly in \(u\), and the remainder is \(O(r^{\min\{2,2\alpha\}})=o(r)\) since \(\min\{2,2\alpha\}>1\).
Hence there is \(r_0>0\), depending only on \(N,\tau,|\mathcal C|,C_\pi\), with
\begin{equation*}
\Phi(ru)-\Phi(0)\le-r<0
\qquad\text{for all }0<r\le r_0\text{ and all }u\in\Delta^{N-1},
\end{equation*}
so \(a=0\) is a strict local maximum: \(z\) is potential-viable.
The correction is \(O(r^{\min\{2,\alpha\}})\), combining rival terms of order \(r^\alpha\) with the quadratic logistic term.
For \(\alpha>2\), the logistic term sets this order.

\emph{(ii) \(\alpha<1\).}
Let \(w\) be a nearest rival and choose \(u\) uniform on its defect set, \(u_i=1/d_{\min}\) for \(i\in D(w,z)\) and \(0\) elsewhere.
This is a legitimate point of the closed simplex, and \eqref{eq:S-in-polar} keeps only terms whose defect set lies inside \(D(w,z)\), so \(S_z(ru)\ge(\pi_w/\pi_z)^{1/\tau}d_{\min}^{-\alpha}r^{\alpha}=:c\,r^{\alpha}\) with \(c>0\).
Then
\begin{equation}
\Phi(ru)-\Phi(0)\ge2\tau c\,r^{\alpha}-2r-O(r^{2\alpha})>0
\label{eq:not-max}
\end{equation}
for all small \(r>0\), because \(\alpha<\min\{1,2\alpha\}\).
This excludes local maximality and potential viability of \(z\).

\emph{(iii) \(\alpha=1\).}
Now the nearest-rival terms of \eqref{eq:S-in-polar} are exactly linear in \(r\) and the rest are \(O(r^{d_2/\tau})\) with \(d_2>\tau\) the next-nearest distance, so \(2\tau S_z(ru)=2r\,H_z(u)+o(r)\) with \(H_z\) as in \eqref{eq:H-functional}, uniformly in \(u\) because \(H_z\) is a finite sum of monomials bounded on the simplex.
The remainder in \eqref{eq:expansion-correct} is \(O(r^2)\).
Hence
\begin{equation}
\Phi(ru)-\Phi(0)=2r\bigl[H_z(u)-1\bigr]+o(r)
\qquad\text{uniformly in }u,
\label{eq:critical-expansion}
\end{equation}
The maximum \(\chi_z=\max_{\Delta^{N-1}}H_z\) is attained because \(H_z\) is continuous on the compact simplex.
The case \(\chi_z<1\) gives \(\Phi(ru)-\Phi(0)\le2r(\chi_z-1)+o(r)<0\) uniformly, so \(z\) is potential-viable, while \(\chi_z>1\) gives a direction along which the expression is positive.
At \(\chi_z=1\) the leading term vanishes on the maximizing set, requiring higher-order terms or the exact expression.
\end{proof}

\subsection{The critical functional is a concave maximization}
\label{app:critical-functional}

\begin{proposition}[\(\chi_z\) is computable]
\label{prop:chi-concave}
In the critical case \(d_{\min}(z)=\tau\), each summand of \(H_z\) is a weighted geometric mean of the coordinates of \(u\) whose exponents \(1/\tau\) over \(i\in D(w,z)\) sum to \(d_{\min}(z)/\tau=1\).
Weighted geometric means with exponents summing to one are concave on the nonnegative orthant, so \(H_z\) is concave and
\begin{equation*}
\chi_z=\max_{u\in\Delta^{N-1}}H_z(u)
\end{equation*}
is a concave maximization over the simplex, solvable to global optimality.
\end{proposition}

For integer critical degree \(k=\tau=d_{\min}(z)\ge2\), \(\chi_z\) coincides with the nonnegative adjacency spectral radius of the weighted \(k\)-uniform defect hypergraph \citep{cooper2012spectra}.
The two arrangements in Example~\ref{ex:overlap} give explicit values:
Write \(c=(\pi_w/\pi_z)^{1/\tau}\) for the common prior factor and take \(\tau=2\), \(d_{\min}=2\), so each summand is \(2c\sqrt{u_iu_j}\).
With either rival alone, \(\chi_z=c\), since \(2\sqrt{u_iu_j}\le u_i+u_j\le1\), with equality at \(u_i=u_j=1/2\).

\emph{Disjoint defect sets} \(\{1,2\}\) and \(\{3,4\}\).
By AM--GM, \(\sqrt{u_1u_2}+\sqrt{u_3u_4}\le\frac{u_1+u_2}{2}+\frac{u_3+u_4}{2}=\frac12\), with equality at \(u_1=u_2\), \(u_3=u_4\).
Hence \(\chi_z=2c\cdot\frac12=c\).

\emph{Overlapping defect sets} \(\{1,2\}\) and \(\{1,3\}\).
Here \(H_z=2c\sqrt{u_1}\bigl(\sqrt{u_2}+\sqrt{u_3}\bigr)\).
Setting \(u_4=0\) and \(u_2=u_3=(1-u_1)/2\) by symmetry gives \(H_z=2c\sqrt{2u_1(1-u_1)}\), maximized at \(u_1=\tfrac12\), so \(\chi_z=\sqrt2\,c\).

With \(c=0.8\), that is \(\pi_w/\pi_z=c^\tau=0.64\), the two values are \(0.800\) and \(\sqrt2\cdot0.8\approx1.131\): overlap alone moves \(\chi_z\) across the threshold, and no numerical maximization is needed to see it.

\subsection{Discrete trapping and the compactified rate}
\label{app:viability-sufficiency-proof}

\begin{theorem}[Local endpoint stability]
\label{thm:viability-sufficiency}
Fix a candidate code \(z\in\mathcal C\).
Suppose either
\begin{equation*}
d_{\min}(z)>\tau,
\qquad\text{or}\qquad
d_{\min}(z)=\tau
\quad\text{and}\quad
\chi_z<1+\lambda.
\end{equation*}
If \(\lambda>0\), assume \(z=\one_N\) is the semantic code.
Fix \(K\succ0\) and \(\eta>0\) satisfying \eqref{eq:general-ascent-condition}.

Then there are positive constants \(c_1,c_2,c_K,r_\star\), with \(c_K\) depending additionally on \(K\) and \(\eta\), such that for every \(0<\varepsilon\le r_\star\), the compactified initialization condition
\begin{equation*}
r_0<\frac{c_1}{c_2}\varepsilon,
\end{equation*}
implies \(r_t<\varepsilon\) for all \(t\ge0\) and
\begin{equation*}
c_1 r_t\le V_z(a_t)\le c_2 r_t,
V_{z,t+1}\le V_{z,t}-c_K r_t^2,
\end{equation*}
where \(r_t=\sum_i a_{i,t}\), \(a_i=e^{-z_iy_i}\), and \(V_z(a)=-[\Phi(a)-\Phi(0)]\) is the potential deficit.

Consequently, \(z\) is an asymptotically stable endpoint with \(r_t=O(1/t)\).
\end{theorem}

\begin{proof}
First pass to signed coordinates:
\begin{equation}
x_i=z_iy_i,\qquad a_i=e^{-x_i},\qquad
D_z=\operatorname{diag}(z),\qquad K_z=D_zKD_z\succ0,
\label{eq:signed-setup}
\end{equation}
so that \(x=D_zy\), \(dx=D_zdy\), and the map~\eqref{eq:general-map} becomes
\begin{equation}
x^+=x+\eta K_zg_z(a),\qquad g_z(a):=D_zR(y)=\nabla_x\Phi.
\label{eq:signed-map}
\end{equation}
The potential deficit is \(V_z(a)=-[\Phi(a)-\Phi(0)]\), and \(z\) is approached as \(a\to0\).
Write \(r=\sum_ia_i\) and \(u=a/r\in\Delta^{N-1}\).

\paragraph{Exact gradient identity and uniform bounds.}
For each rival \(w\ne z\), define the monomial
\begin{equation*}
m_w(a)
=
\left(\frac{\pi_w}{\pi_z}\right)^{1/\tau}
\prod_{i\in D(w,z)}a_i^{1/\tau},
\qquad
S_z(a)=\sum_{w\ne z}m_w(a).
\end{equation*}
By direct computation from the posterior~\eqref{eq:general-posterior} and the residual definition \(R_i=2[p_i+\lambda-(1+\lambda)\sigmoid(y_i)]\), the signed gradient is
\begin{equation}
g_{z,i}(a)
=
\frac{2(1+\lambda)a_i}{1+a_i}
-
\frac{2}{1+S_z(a)}
\sum_{\substack{w\ne z\\i\in D(w,z)}}m_w(a).
\label{eq:exact-gradient}
\end{equation}
This identity holds for all \(a\), including simplex faces \(a_i=0\), because \(m_w\) has positive exponents.

\emph{Upper bound.}
Each \(m_w(a)\le C_wr^{\deg(w)}\) with \(\deg(w)=d_{\rm H}(w,z)/\tau\), and \(1/(1+S_z)\le1\).
Since \(a_i/(1+a_i)<a_i\), we get
\begin{equation}
|g_{z,i}(a)|\le2(1+\lambda)a_i+2\!\!\sum_{w:i\in D(w,z)}\!\!m_w(a)\le C_4r,
\label{eq:gradient-upper}
\end{equation}
uniformly in \(u\), where \(C_4\) depends on \(N\), \(\tau\), \(|\mathcal C|\), \(\lambda\), and the prior ratios.

\emph{Lower bound, strictly viable case \(d_{\min}>\tau\).}
Set \(\alpha=d_{\min}(z)/\tau>1\).
Every rival degree is at least \(\alpha\), so \(S_z(a)=O(r^\alpha)\) and each monomial sum in~\eqref{eq:exact-gradient} is \(O(r^\alpha)\).
The leading term of \(g_{z,i}\) is \(2(1+\lambda)a_i\), with correction \(O(r^{\min\{2,\alpha\}})=O(r^{1+\delta})\) where \(\delta=\min\{1,\alpha-1\}>0\).
Thus
\begin{equation*}
g_z=2(1+\lambda)a+O(r^{1+\delta}),
\end{equation*}
giving \(\|g_z\|_2\ge(1+\lambda)r/\sqrt N\) for sufficiently small \(r\).

\emph{Lower bound, critically viable case \(d_{\min}=\tau\), \(\chi_z<1+\lambda\).}
The nearest rivals have \(\deg(w)=1\).
Summing~\eqref{eq:exact-gradient} over \(i\) and using \(\sum_{i:i\in D(w,z)}1=d_{\rm H}(w,z)\):
\begin{equation*}
\mathbf{1}^\top g_z
=
2(1+\lambda)\sum_i\frac{a_i}{1+a_i}
-
\frac{2}{1+S_z(a)}
\sum_{w\ne z}d_{\rm H}(w,z)\,m_w(a).
\end{equation*}
For critical nearest rivals \(w\) with \(d_{\rm H}(w,z)=\tau\), define the monomial coefficient
\begin{equation*}
h_w(u)
=
\left(\frac{\pi_w}{\pi_z}\right)^{1/\tau}
\prod_{i\in D(w,z)}u_i^{1/\tau},
\end{equation*}
so that \(m_w(ru)=rh_w(u)\) exactly.
The critical functional is then \(H_z(u)=\tau\sum_{w:\,d_{\rm H}(w,z)=\tau}h_w(u)\), as in Theorem~\ref{thm:viability}.
For farther rivals with \(d_{\rm H}(w,z)>\tau\) the contributions are \(o(r)\).
Since \(\sum_ia_i/(1+a_i)=r-O(r^2)\) and \(1/(1+S_z)=1-O(r)\), we obtain
\begin{equation*}
\mathbf{1}^\top g_z
=
2r[(1+\lambda)-H_z(u)]+o(r),
\end{equation*}
uniformly in \(u\).
The viability gap \(\varepsilon=(1+\lambda)-\chi_z>0\) makes \(H_z(u)\le\chi_z<1+\lambda\) uniformly, so \(\mathbf{1}^\top g_z\ge\varepsilon r\) for small \(r\).
By Cauchy--Schwarz, \(\|g_z\|_2\ge\varepsilon r/\sqrt N\).

Set the gradient lower-bound constant \(c_3\) as follows:
\begin{equation*}
c_3=
\begin{cases}
\dfrac{1+\lambda}{\sqrt N},
&
d_{\min}(z)>\tau,\\[1.5ex]
\dfrac{(1+\lambda)-\chi_z}{\sqrt N},
&
d_{\min}(z)=\tau,\ \chi_z<1+\lambda.
\end{cases}
\end{equation*}
From~\eqref{eq:gradient-upper}, the componentwise bound implies \(\|g_z\|_2\le\sqrt N\,C_4r\).
Renaming \(\widetilde C_4=\sqrt N\,C_4\), we have
\begin{equation}
c_3r\le\|g_z\|_2\le\widetilde C_4r
\label{eq:gradient-bounds}
\end{equation}
uniformly in \(u\in\Delta^{N-1}\) for sufficiently small \(r\).

\paragraph{Potential gap bounds.}
By Lemma~\ref{lem:uniform-expansion}, uniformly in \(u\in\Delta^{N-1}\):
\begin{equation*}
\Phi(ru)-\Phi(0)
=
-2(1+\lambda)r+2\tau S_z(ru)+O\!\left(r^{\min\{2,2\alpha\}}\right),
\end{equation*}
where \(\alpha=d_{\min}/\tau\).
Since \(V_z(a)=-[\Phi(a)-\Phi(0)]\):

\emph{Strictly viable (\(\alpha>1\)):} \(S_z=O(r^\alpha)\), so \(V_z=2(1+\lambda)r+O(r^{\min\{2,\alpha\}})\).
For small \(r\), set \(c_1=1+\lambda\) and \(c_2=3(1+\lambda)\).

\emph{Critically viable (\(\alpha=1\)):} \(V_z=2r[(1+\lambda)-H_z(u)]+o(r)\).
The gap \(\delta=(1+\lambda)-\chi_z>0\) ensures \((1+\lambda)-H_z(u)\ge\delta\) uniformly.
For small \(r\), set \(c_1=\delta\) and \(c_2=2(1+\lambda)+\delta\).

\paragraph{Quantitative ascent and trapping.}
The ascent condition~\eqref{eq:general-ascent-condition} gives the smoothness constant \(L_\Phi\le N/(2\tau)+(1+\lambda)/2\).
Set \(\theta=\eta L_\Phi\lambda_{\max}(K)/2<1\).
By the signed-coordinate map~\eqref{eq:signed-map} and the lower smoothness inequality,
\begin{equation*}
\Phi(x^+)-\Phi(x)
\ge
\eta g_z^\top K_zg_z-\frac{L_\Phi\eta^2}{2}\|K_zg_z\|^2
\ge
\eta(1-\theta)\lambda_{\min}(K)\|g_z\|^2.
\end{equation*}
Combined with \(\|g_z\|\ge c_3r\), this gives
\begin{equation}
V_{t+1}\le V_t-\eta(1-\theta)\lambda_{\min}(K)c_3^2r_t^2.
\label{eq:descent-inequality}
\end{equation}
Set \(c_K=\eta(1-\theta)\lambda_{\min}(K)c_3^2>0\).

\emph{Trapping.}
From~\eqref{eq:gradient-upper} and the one-step update,
\begin{equation*}
\|\Delta x\|_\infty=\eta\|K_zg_z\|_\infty
\le\eta\lambda_{\max}(K)\widetilde C_4r_t=:B_Kr_t,
\end{equation*}
and since \(a_i^+=a_ie^{-\Delta x_i}\),
\begin{equation}
e^{-B_Kr_t}r_t\le r_{t+1}\le e^{B_Kr_t}r_t.
\label{eq:radial-evolution}
\end{equation}
Choose \(r_\star>0\) small enough that the gap bounds \(c_1r\le V_z\le c_2r\), the gradient bounds~\eqref{eq:gradient-bounds}, and \(e^{2B_Kr_\star}<2\) all hold for \(r\le2r_\star\).

For any \(0<\varepsilon\le r_\star\), suppose the initial condition satisfies
\begin{equation*}
r_0<\frac{c_1}{c_2}\varepsilon,
\end{equation*}
so that \(V_0\le c_2r_0<c_1\varepsilon\).
If there is a first time \(t^\star\) with \(r_{t^\star}\ge\varepsilon\), then \(r_{t^\star-1}<\varepsilon\le r_\star\), so \(r_{t^\star}<2\varepsilon\le2r_\star\) by~\eqref{eq:radial-evolution}, and the local bounds apply at \(t^\star\):
\begin{equation*}
V_{t^\star}\ge c_1r_{t^\star}\ge c_1\varepsilon.
\end{equation*}
But the potential deficit \(V_t\) is nonincreasing by~\eqref{eq:descent-inequality}, so \(V_{t^\star}\le V_0<c_1\varepsilon\), a contradiction.
Hence \(r_t<\varepsilon\) for all \(t\ge0\).
This proves Lyapunov stability for every \(\varepsilon\le r_\star\).
The Riccati argument below also gives \(r_t\to0\) throughout the trapping neighborhood, proving local asymptotic stability.

\paragraph{Convergence rate.}
From~\eqref{eq:descent-inequality} and the upper gap bound,
\begin{equation}
V_{t+1}\le V_t-c_Kr_t^2\le V_t-\frac{c_K}{c_2^2}V_t^2.
\label{eq:V-riccati}
\end{equation}
Let \(\gamma_K=c_K/c_2^2>0\).
For small \(V_t\) (which holds once \(t\ge t_0\)),
\begin{equation}
\frac{1}{V_{t+1}}\ge\frac{1}{V_t}+\gamma_K.
\label{eq:reciprocal-riccati}
\end{equation}
Summing from \(t_0\) to \(T-1\) gives \(1/V_T\ge\gamma_K(T-t_0)+1/V_{t_0}\), so \(V_T=O(1/T)\).
The lower gap bound \(c_1r_T\le V_T\) then yields \(r_T\le V_T/c_1=O(1/T)\).
\end{proof}

\subsection{Proofs of the main local classifications}
\label{app:local-classifications}

\paragraph{Proposition~\ref{cor:local-iff}.}
\begin{proof}
With \(\lambda=0\), Theorem~\ref{thm:viability} classifies strict local maxima in every nondegenerate regime.
Lemma~\ref{prop:viability-necessary} gives necessity for asymptotic stability and Theorem~\ref{thm:viability-sufficiency} gives sufficiency.
\end{proof}

\paragraph{Theorem~\ref{cor:native-a3bl}.}
\begin{proof}
Distance at least two gives stability and the \(O(1/t)\) rate by Theorem~\ref{thm:viability-sufficiency}.
Otherwise a one-bit rival differs only at coordinate \(i\).
On the edge \(a_i=s>0\), \(a_{k\ne i}=0\), the exact factorization~\eqref{eq:exact-factorization} has \(S_z(a)=s\), so \(\Phi(a)-\Phi(0)=-2\log(1+s)+2\log(1+s)=0\).
These points are also residual zeros.
They accumulate at the corner, excluding local attraction and strict local maximality.
\end{proof}

\subsection{Direct semantic supervision}
\label{app:supporting}
\label{app:support}

Direct semantic supervision changes the semantic corner's linear saturation term but no rival exponent.
Its threshold concerns local stability at the semantic anchor, not global recovery.

\begin{corollary}[Direct semantic supervision threshold]
\label{cor:support-coefficient}
Let \(+\) be the semantic code and let direct semantic supervision have relative weight \(\lambda\ge0\).
Then, uniformly in \(u\),
\begin{equation*}
\Phi_{+,\lambda}(ru)-\Phi_{+,\lambda}(0)
=2r\bigl[H_+(u)-(1+\lambda)\bigr]+o(r)
\qquad\text{when }d_{\min}(+)=\tau,
\end{equation*}
so the viability threshold becomes \(\chi_+<1+\lambda\).
If instead \(d_{\min}(+)<\tau\), the offending monomial has degree strictly below one and no finite \(\lambda\) makes \(+\) potential-viable.
\end{corollary}

\begin{proof}[Proof of Corollary~\ref{cor:support-coefficient}]
By Remark~\ref{rem:support-anchor} the anchor is \(z=\one_N\) and the supervision term contributes \(-2\lambda\sum_i\log(1+a_i)=-2\lambda r+O(r^2)\), so the only change in \eqref{eq:expansion-correct} is the linear coefficient, from \(-2\) to \(-2(1+\lambda)\).
Every rival exponent \(d_{\rm H}(w,+)/\tau\) is untouched.
At \(d_{\min}(+)=\tau\) the critical comparison in \eqref{eq:critical-expansion} becomes \(H_+(u)\) against \(1+\lambda\), giving the threshold \(\chi_+<1+\lambda\).
At \(d_{\min}(+)<\tau\) the bound \eqref{eq:not-max} holds with \(2r\) replaced by \(2(1+\lambda)r\), and since \(\alpha<1\) the positive \(r^{\alpha}\) term still dominates for small \(r\) no matter how large \(\lambda\) is.
Thus no finite \(\lambda\) makes \(+\) potential-viable.
\end{proof}

Combining the threshold with Lemma~\ref{prop:viability-necessary} and Theorem~\ref{thm:viability-sufficiency} gives the semantic local classification: outside the equality case, local asymptotic stability holds exactly when \(d_{\min}(+)>\tau\), or when \(d_{\min}(+)=\tau\) and \(\chi_+<1+\lambda\).
At \(\chi_+=1+\lambda\), the first-order expansion does not determine stability.

\subsection{Controlled posterior-refresh errors}
\label{app:inexact}
\label{app:inexact-local-stability}

We retain the parameterized posterior and potential of Appendix~\ref{app:general-model} and a fixed positive-definite Gram matrix.
An approximate posterior refresh returns code distribution \(\widetilde q_t\) and unary marginals \(\widetilde p_t\), inducing
\begin{equation}
\widetilde R_t=R(y_t)+e_t,\qquad
e_t=2[\widetilde p_t-p(y_t)],\qquad
y_{t+1}=y_t+\eta K\widetilde R_t.
\label{eq:inexact-map}
\end{equation}

\begin{lemma}[Inexact ascent and posterior error]
\label{prop:inexact-ascent}
\label{lem:inexact-ascent}
Write \(L_\Phi=\tfrac12(N/\tau+1+\lambda)\) and \(\lVert v\rVert_K^2=v^\top Kv\).
For the update~\eqref{eq:inexact-map}, suppose \(\lVert e_t\rVert_K\le\delta\lVert R(y_t)\rVert_K\).
With \(\theta=\eta L_\Phi\lambda_{\max}(K)/2\),
\begin{equation}
\Phi(y_{t+1})-\Phi(y_t)
\ge
\eta[1-\delta-\theta(1+\delta)^2]
\lVert R(y_t)\rVert_K^2.
\label{eq:inexact-ascent}
\end{equation}
For \(\mathrm{TV}(\widetilde q_t,q(y_t))\le\varepsilon_t\), \(\lVert e_t\rVert_K\le2\sqrt{N\lambda_{\max}(K)}\varepsilon_t\).
\end{lemma}

\begin{proof}
The lower smoothness inequality, Cauchy--Schwarz in the \(K\)-norm, and \(K^2\preceq\lambda_{\max}(K)K\) give
\begin{align*}
\Phi(y_{t+1})-\Phi(y_t)
&\ge \eta R^\top K(R+e)
-\frac{L_\Phi\eta^2}{2}\lVert K(R+e)\rVert_2^2\\
&\ge \eta(1-\delta)\lVert R\rVert_K^2
-\frac{L_\Phi\eta^2\lambda_{\max}(K)}2
 (1+\delta)^2\lVert R\rVert_K^2,
\end{align*}
which is~\eqref{eq:inexact-ascent}.

If \(\mathrm{TV}(\widetilde q,q)\le\varepsilon\), each unary label is an event under the code distribution, so \(\lvert\widetilde p_i-p_i\rvert\le\varepsilon\).
Hence
\[
\lVert e\rVert_K
\le\sqrt{\lambda_{\max}(K)}\lVert e\rVert_2
\le2\sqrt{N\lambda_{\max}(K)}\,\varepsilon.
\]
\end{proof}

The same relative-error condition also preserves local endpoint attraction.

\begin{corollary}[Endpoint stability under controlled inexact abduction]
\label{cor:inexact-local-stability}
Under the assumptions of Theorem~\ref{thm:viability-sufficiency}, replace the exact map by~\eqref{eq:inexact-map}.
Fix \(\delta\ge0\) such that \(\beta_\delta =1-\delta- \eta L_\Phi\lambda_{\max}(K)(1+\delta)^2/2>0\), and suppose \(\lVert e_t\rVert_K\le\delta\lVert R(y_t)\rVert_K\) at every step.
Then the same corner is locally asymptotically stable uniformly over all such time-varying E-step errors: a sufficiently small compactified neighborhood is trapping and \(r_t=O(1/t)\).
\end{corollary}

\begin{proof}[Proof of Corollary~\ref{cor:inexact-local-stability}]
In the signed coordinates of~\eqref{eq:signed-setup}, put \(\epsilon_t=D_ze_t\).
Since \(K_z=D_zKD_z\), orthogonal sign conjugation preserves the relevant norms:
\[
\lVert\epsilon_t\rVert_{K_z}
\le\delta\lVert g_z(a_t)\rVert_{K_z}.
\]
Lemma~\ref{lem:inexact-ascent} and the gradient lower bound \eqref{eq:gradient-bounds} therefore give
\begin{equation*}
V_{t+1}
\le V_t-
\eta\beta_\delta\lambda_{\min}(K)c_3^2r_t^2.
\end{equation*}

It remains to check that an admissible error cannot jump over the local region.
Norm equivalence and~\eqref{eq:gradient-bounds} yield
\[
\lVert\epsilon_t\rVert_2
\le\delta\sqrt{\frac{\lambda_{\max}(K)}{\lambda_{\min}(K)}}
\lVert g_z(a_t)\rVert_2
=O(r_t).
\]
The inexact signed update \(x^+=x+\eta K_z(g_z+\epsilon_t)\) satisfies \(\lVert\Delta x\rVert_\infty\le B_{K,\delta}r_t\) for a finite constant \(B_{K,\delta}\).
Using this bound in \eqref{eq:radial-evolution} gives the same first-exit contradiction, with \(r_\star\) reduced if needed.

Finally, substitute \(c_{K,\delta}=\eta\beta_\delta\lambda_{\min}(K)c_3^2>0\) for \(c_K\) in \eqref{eq:V-riccati}--\eqref{eq:reciprocal-riccati}.
This proves trapping, uniform attraction, and \(r_t=O(1/t)\).
Relative to the exact coefficient \(c_K=\eta(1-\theta)\lambda_{\min}(K)c_3^2\), the coefficient in the uniform deficit-decrease bound is multiplied by \(\beta_\delta/(1-\theta)\).
\end{proof}

\subsection{Random finite candidate sets}
\label{app:random-candidates}

The local criterion guarantees attraction sufficiently close to a viable endpoint.
We examine contraction and escape at fixed finite radii across randomly sampled candidate sets and initial directions, without assuming that these initial states lie inside a certified trapping neighborhood.

For each \(N\in\{6,8,10,12\}\), each candidate count in \(\{4,8,16\}\), and fifteen repeats, we sample distinct binary codes containing the all-positive anchor, giving 180 candidate sets.
Strict tests use \(\tau=d_{\min}\mp0.5\) and \(\lambda=0\).
Critical tests set \(\tau=d_{\min}\), cycling target values of \(\chi_z\) through \((0.6,0.8,0.95,1.05,1.2,1.4)\) (30 sets per target) by maximizing the base-prior functional and rescaling the anchor prior mass.
All three regimes for a set use \(K=I\) and two random SPD Grams with eigenvalues in \([0.35,1]\), initial radii \(r_0\in\{0.1,0.05,0.01\}\), and constant step size \(\eta=1.6/(N/\tau+1)\).

A run is classified as contracting when its sampled maximum radius stays below \(1.5r_0\), its final anchor posterior exceeds 0.9, and its final radius is below \(0.995\) times its midpoint radius (sampled every ten steps).
It escapes when its radius reaches \(4r_0\) or a rival obtains a log-posterior lead above 10, stopping upon detection; remaining runs after 8,000 steps are unresolved.
All cases evaluate a witness, a dense, and a near-face simplex direction (floored componentwise at \(10^{-10}\) and renormalized before forming logits), where the witness is supported on a nearest defect for strict tests and is the maximizing direction for critical tests.
For viable cases we pool all three directions, whereas for non-viable cases we evaluate instability along the witness direction, yielding 3,240 trajectories per radius (9,720 overall; Table~\ref{tab:local-breakdown}).

\begin{table}[!htbp]
\centering
\caption{Finite-radius outcomes by regime and initial radius \(r_0\), in percent.
Resolved runs meet either the contraction or escape criterion.
The last three columns report contraction for viable cases and witness-direction escape for non-viable cases, conditional on resolution.
The critical and combined rows pool these two outcomes.}
\label{tab:local-breakdown}
\small
\begin{tabular}{lrrrrrr}
\toprule
& \multicolumn{3}{c}{Resolved} & \multicolumn{3}{c}{Contraction / escape} \\
\cmidrule(lr){2-4}\cmidrule(lr){5-7}
Regime (runs per radius) & \(0.01\) & \(0.05\) & \(0.10\)
                         & \(0.01\) & \(0.05\) & \(0.10\) \\
\midrule
Strict viable (1{,}620)  & 100 & 100 & 99.9 & 99.8 & 99.6 & 98.7 \\
Strict non-viable (540)  & 100 & 97.2 & 92.2 & 100 & 100 & 100 \\
Critical (1{,}080)       & 95.0 & 99.4 & 99.0 & 99.3 & 97.5 & 96.9 \\
\midrule
All regimes (3{,}240)    & 98.3 & 99.3 & 98.3 & 99.7 & 99.0 & 98.3 \\
\bottomrule
\end{tabular}
\end{table}

\begin{figure}[!htbp]
\centering
\includegraphics[width=0.67\textwidth]{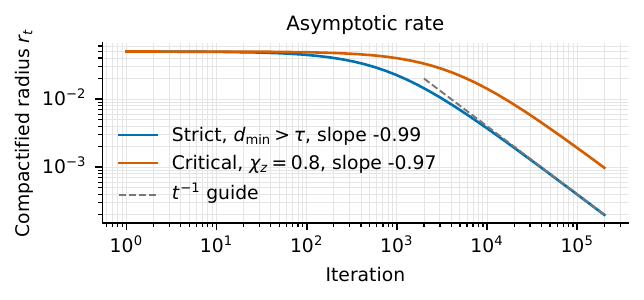}
\caption{Compactified-radius traces for one strict and one critical stable example, with a \(t^{-1}\) guide.
Log-log tail slopes are fitted over the final half of each 200,000-update trace.}
\label{fig:local-rate}
\end{figure}

Overall, 98.7\% of the 9,720 trajectories resolve within budget, and 99.0\% of resolved runs contract toward a viable corner or escape along a non-viable corner's witness direction.
The remaining 97 runs escape from viable corners: 10 at \(r_0=0.01\), 33 at \(r_0=0.05\), and 54 at \(r_0=0.10\) (Table~\ref{tab:local-breakdown}), illustrating why local stability differs from attraction at a fixed finite distance.
Unresolved runs appear only at larger radii in the strict non-viable regime and only at \(\chi_z\in\{0.95,1.05\}\) nearest the boundary in the critical regime.
In Figure~\ref{fig:local-rate}, the strict and critical 200,000-step traces have tail slopes \(-0.995\) and \(-0.971\).

\FloatBarrier

\section{Isospectral Routing}
\label{app:routing}

The routing construction places one slow direction within the candidate-active subspace.
One orientation preserves the symmetric route to the semantic code, whereas the others keep a target block negative while the remaining blocks move toward positive labels.
Local trapping and finite-horizon continuity extend these routes to open neighborhoods of the initial state and Gram matrix.

\subsection{Candidate family, block reduction, and isospectral Grams}
\label{app:m-setup}
\label{app:block-dynamics}
\label{app:isospectral-construction}

Recall from Section~\ref{sec:robust-isospectral} that for \(m\ge2\) and \(N=2m\), the coordinates partition into blocks \(\{2j-1,2j\}\) (\(j=1,\ldots,m\)) and \(\mathcal C_m=\{z_0,z_1,\ldots,z_m\}\), where \(z_0=\mathbf1_{2m}\) and \(z_j\) flips only block~\(j\).
Under native parameters (\(\tau=1\), \(\lambda=0\), uniform priors), \(d_{\mathrm H}(z_0,z_j)=2\) and \(d_{\mathrm H}(z_j,z_k)=4\) for \(j\ne k\), so \(d_{\min}(z)=2>\tau\) and every \(z\in\mathcal C_m\) is locally asymptotically stable by Theorem~\ref{thm:viability-sufficiency}.
Let \(B=[c_1,\ldots,c_m]\) with \(c_j=(e_{2j-1}+e_{2j})/\sqrt2\).

\begin{lemma}[Exact \(m\)-dimensional reduction]
\label{prop:m-block-reduction}
Put \(\widehat K=B^\top KB\).
If \(K\succ0\) preserves \(\operatorname{span}(B)\), the block-constant subspace \(y=\sqrt2 Bu=(u_1,u_1,\ldots,u_m,u_m)\) is invariant, with exact dynamics \(u^+=u+\eta\widehat K r(u)\), where \(r_j(u)\) is the residual of either coordinate in block~\(j\) and \(R(\sqrt2 Bu)=\sqrt2 Br(u)\).
\end{lemma}

\begin{proof}
Within-block exchange symmetry makes \(R(y)\) block-constant whenever \(y\) is, so \(R(\sqrt2 Bu)=\sqrt2 Br(u)\), and a block-preserving \(K\) keeps \(KR(y)\) in \(\operatorname{span}(B)\).
Left-multiplying \(y^+=y+\eta KR(y)\) by \(B^\top/\sqrt2\) gives \(u^+=u+\eta B^\top KB\,r(u)\).
\end{proof}

\begin{lemma}[General block residual]
\label{lem:m-block-residual}
On \(y=(u_1,u_1,\ldots,u_m,u_m)\) with \(a_j=e^{-u_j}\), either coordinate in block~\(j\) has residual
\begin{equation}
\frac{R_j}{2}
=
\frac{a_j\left(1+\sum_{k\ne j}a_k^2-a_j\right)}
{(1+a_j)\left(1+\sum_{k=1}^m a_k^2\right)}.
\label{eq:m-residual}
\end{equation}
\end{lemma}

\begin{proof}
The posterior weights on the block-constant subspace are \(q_{z_0}=(1+\sum_{k=1}^m a_k^2)^{-1}\) and \(q_{z_j}=a_j^2 q_{z_0}\).
Every code except \(z_j\) assigns \(+1\) to block~\(j\), so its marginal is \(p_i=(1+\sum_{k\ne j}a_k^2)/(1+\sum_{k=1}^m a_k^2)\).
Substituting \(\sigmoid(u_j)=(1+a_j)^{-1}\) into \(R_i/2=p_i-\sigmoid(u_j)\) yields~\eqref{eq:m-residual}.
\end{proof}

\begin{lemma}[Full and active-subspace isospectrality]
\label{prop:m-isospectral}
For the matrices \(H_j\) and \(K_j=BH_jB^\top+I_{2m}-BB^\top\) (\(j=0,\ldots,m\)) in~\eqref{eq:main-isospectral-grams}, every \(K_j\) acts as the identity on \(\operatorname{span}(B)^\perp\) and satisfies \(\operatorname{spec}(K_j)=\{\alpha,1^{[2m-1]}\}\) and \(\operatorname{spec}(B^\top K_jB)=\operatorname{spec}(H_j)=\{\alpha,1^{[m-1]}\}\).
Moreover, \(H_0\mathbf1_m=\mathbf1_m\), while \(K_j c_j=\alpha c_j\) and \(K_j c_\ell=c_\ell\) for \(\ell\ne j\) when \(j\ge1\).
\end{lemma}

\begin{proof}
Because \(v\) and \(e_1,\ldots,e_m\) are unit vectors in \(\mathbb R^m\), each \(H_j\) has eigenvalue \(\alpha\) along its rank-one direction and eigenvalue \(1\) on the \((m-1)\)-dimensional orthogonal complement.
Since \(B\) has orthonormal columns, \(B^\top K_jB=H_j\), while \(I_{2m}-BB^\top\) contributes \(m\) additional unit eigenvalues on \(\operatorname{span}(B)^\perp\).
Finally, \(v^\top\mathbf1_m=0\) gives \(H_0\mathbf1_m=\mathbf1_m\), and \(H_j e_\ell\) gives the stated action on \(c_\ell\) for \(j\ge1\).
\end{proof}

\subsection{Semantic routing}
\label{app:isotropic-routing}

\begin{lemma}[Symmetric semantic selection]
\label{thm:m-isotropic}
For a reduced metric \(H\) satisfying \(H\mathbf1_m=k\mathbf1_m\) with \(k>0\), every symmetric adverse initialization \(u_{1,0}=\cdots=u_{m,0}=s_0<0\) converges to \(z_0\).
\end{lemma}

\begin{proof}
The residual on the symmetric line is a multiple of \(\mathbf1_m\), and \(H\mathbf1_m=k\mathbf1_m\) keeps this line invariant.
Thus \(u_{1,t}=\cdots=u_{m,t}=:s_t\) for every~\(t\).
With \(a=e^{-s}\), Lemma~\ref{lem:m-block-residual} reduces to
\begin{equation*}
\frac{R(s)}{2}
=\frac{a\left(1+(m-1)a^2-a\right)}{(1+a)(1+ma^2)}.
\end{equation*}
The quadratic \(1+(m-1)a^2-a\) has discriminant \(1-4(m-1)<0\), so \(R(s)>0\) for every finite~\(s\).
The sequence \(s_t\) is strictly increasing.
If bounded, it would converge to a finite \(L\), contradicting \(s_{t+1}-s_t\to\eta kR(L)>0\).
Hence \(s_t\to+\infty\), giving convergence to \(z_0\).
\end{proof}

\subsection{Rival routing and the absorbing cone}
\label{app:anisotropic-routing}

Fix a target \(z_j\).
Starting from the symmetric initialization and assigning one weight to block~\(j\) and a common weight to all other blocks preserves the two-variable subspace
\begin{equation*}
u_j=u,
\qquad
u_k=v \quad(k\ne j).
\end{equation*}
Put \(a=e^{-u}\), \(b=e^{-v}\).
Lemma~\ref{lem:m-block-residual} gives
\begin{align*}
\frac{R_u}{2}
&=\frac{a\left(1+(m-1)b^2-a\right)}
{(1+a)\left(1+a^2+(m-1)b^2\right)},\\
\frac{R_v}{2}
&=\frac{b\left(1+a^2+(m-2)b^2-b\right)}
{(1+b)\left(1+a^2+(m-1)b^2\right)}.
\end{align*}

\begin{lemma}[\(m\)-code absorbing cone]
\label{lem:m-cone-absorption}
On this two-variable subspace, the region
\begin{equation*}
\mathcal R_j^{(m)}
=\left\{a\ge1+(m-1)b^2\right\}
\end{equation*}
is forward invariant for every fixed pair of positive effective weights \(h_u,h_v\).
Every trajectory entering it satisfies \(a_t\to\infty\) and \(b_t\to0\), and hence converges to \(z_j\).
\end{lemma}

\begin{proof}
Inside the cone, \(R_u\le0\).
Moreover, \(a\ge1+b^2\), so the numerator factor of \(R_v\) obeys
\begin{align*}
1+a^2+(m-2)b^2-b
&\ge2+mb^2+b^4-b\\
&>0,
\end{align*}
where \(mb^2-b\ge-1/(4m)\) and \(m\ge2\).
Thus \(R_v>0\).
For positive effective steps \(h_u,h_v\),
\begin{equation}
a^+=a e^{-h_uR_u}\ge a,
\qquad
b^+=b e^{-h_vR_v}<b.
\label{eq:cone-exponential-updates}
\end{equation}
It follows that \(a^+\ge a\ge1+(m-1)b^2\ge1+(m-1)(b^+)^2\), proving invariance.

The sequences \(a_t\) and \(b_t\) are respectively nondecreasing and strictly decreasing.
Let \(w_t=a_t-1-(m-1)b_t^2\ge0\).
Equation~\eqref{eq:cone-exponential-updates} makes \(w_t\) nondecreasing and \(w_t>0\) after at most one step.
If \(a_t\to a_\infty<\infty\) and \(b_t\to0\), this positive slack makes \(R_u\) eventually negative and bounded away from zero, contradicting finite convergence of \(a_t\).
If \(a_t\to\infty\) and \(b_t\to b_\infty>0\), then
\[
R_v(a_t,b_t)\longrightarrow\frac{2b_\infty}{1+b_\infty}>0,
\]
contradicting finite convergence of \(b_t\).
If both limits are finite, both residuals must vanish.
Their numerator equations would be
\[
a=1+(m-1)b^2,
\qquad
b=1+a^2+(m-2)b^2.
\]
After substitution, the second right-hand side minus \(b\) is
\[
2+(3m-4)b^2+(m-1)^2b^4-b
\ge 2+2b^2-b>0,
\]
so no such equilibrium exists.
The only remaining limits are \(a_t\to\infty\) and \(b_t\to0\).
\end{proof}

\begin{lemma}[Anisotropic \(m\)-code selection]
\label{thm:m-anisotropic}
Fix \(s_0<0\) and a target-block step \(h_1>0\) satisfying
\begin{equation*}
h_1\le\frac{|s_0|}{4}.
\end{equation*}
Give every non-target block step \(h_2=\rho h_1\).
There is a finite \(\rho_m^\star(s_0,h_1)\) such that every \(\rho>\rho_m^\star(s_0,h_1)\) routes the symmetric initialization to \(z_j\).
\end{lemma}

\begin{proof}
Set \(M=e^{-s_0}>1\) and
\begin{equation*}
\mathcal D_m
=\left[s_0,\frac{s_0}{2}\right]\times[s_0,V_m],
\qquad
V_m=-\frac12\log\frac{\sqrt M-1}{m-1}.
\end{equation*}
Here \(V_m>s_0\), so the rectangle is nonempty.
If the trajectory enters the absorbing cone, Lemma~\ref{lem:m-cone-absorption} gives the desired convergence.
The following estimates apply until that entry.
Before the trajectory enters the cone, \(a<1+(m-1)b^2\), so \(R_u>0\).
Thus \(u_t\) cannot leave the rectangle through its lower edge.
Also \(|R_u|\le2\), so the slow coordinate remains in \([s_0,s_0/2]\) for at least
\begin{equation*}
n_\star=\left\lfloor\frac{|s_0|}{4h_1}\right\rfloor\ge1
\end{equation*}
steps.

While \(v\in[s_0,V_m]\), we have \(a\in[\sqrt M,M]\), \(b\le M\), and \(b\ge b_{\min}:=e^{-V_m}=\sqrt{(\sqrt M-1)/(m-1)}\).
The numerator factor of \(R_v\) is at least \(1+M-b\ge1\), and its denominator is at most \((1+M)(1+mM^2)\).
Hence throughout the rectangle
\begin{equation}
R_v\ge
\underline c_{2,m}
:=\frac{2b_{\min}}{(1+M)(1+mM^2)}>0.
\label{eq:c2m-lower-bound}
\end{equation}
Let \(c_{2,m}:=\min_{\mathcal D_m}R_v\), so \(c_{2,m}\ge\underline c_{2,m}>0\).
Define
\begin{equation*}
\rho_m^\star(s_0,h_1)
:=\frac{V_m-s_0}{n_\star h_1c_{2,m}}.
\end{equation*}
If \(\rho>\rho_m^\star\), the fast coordinate reaches or crosses \(V_m\) within \(n_\star\) steps, since otherwise its cumulative increase would exceed \(V_m-s_0\).
At that step \(v\ge V_m\), so \(b^2\le(\sqrt M-1)/(m-1)\), while \(a\ge\sqrt M\).
Therefore \(a\ge1+(m-1)b^2\).
The trajectory enters \(\mathcal R_j^{(m)}\), so Lemma~\ref{lem:m-cone-absorption} gives the result.
\end{proof}

\subsection{Integer-aligned parameter construction}
\label{app:integer-parameters}

\begin{lemma}[Compatible parameter choice]
\label{lem:m-parameter-choice}
For every \(m\ge2\) and \(s_0<0\), there are \(\rho>1\), \(\alpha=1/\rho\), and \(\eta>0\) such that \(h_1=\eta\alpha\le|s_0|/4\), \(\rho>\rho_m^\star(s_0,h_1)\), and \(\eta(2m+1)<4\).
\end{lemma}

\begin{proof}
Use \(M,V_m,\underline c_{2,m}\) from the proof of Lemma~\ref{thm:m-anisotropic} and define
\begin{equation*}
\bar\rho_m
=\frac{4(V_m-s_0)}{|s_0|\underline c_{2,m}},
\qquad
\rho=2\max\{1,\bar\rho_m\},
\qquad
\alpha=\rho^{-1}.
\end{equation*}
Choose
\begin{equation*}
q=\left\lceil\frac{\rho|s_0|(2m+1)}{8}\right\rceil+1
\end{equation*}
and set
\begin{equation*}
h_1=\frac{|s_0|}{4q},
\qquad
\eta=\rho h_1=\frac{\rho|s_0|}{4q}.
\end{equation*}
Then \(\eta\alpha=h_1\le|s_0|/4\), and
\[
\eta(2m+1)<2<4.
\]
Moreover, \(n_\star=\lfloor|s_0|/(4h_1)\rfloor=q\), so
\[
\rho_m^\star(s_0,h_1)
\le\frac{V_m-s_0}{q h_1\underline c_{2,m}}
=\bar\rho_m
<\rho.
\]
This choice satisfies the slow-step, routing, and ascent requirements simultaneously.
\end{proof}

\subsection{Robust continuation}
\label{app:robust-continuation}

\begin{lemma}[Robust continuation of endpoint selection]
\label{lem:robust-continuation}
Use the parameterized update~\eqref{eq:general-map} and let \(F_K(y)=y+\eta KR(y)\).
If \(\lambda>0\), take \(z=\one_N\).
Suppose that a code \(z\) satisfies either \(d_{\min}(z)>\tau\), or \(d_{\min}(z)=\tau\) and \(\chi_z<1+\lambda\).
Suppose also that \(F_{K_\star}^t(y_\star)\to z\) for some \(K_\star\succ0\), and that the ascent condition holds strictly at \(K_\star\).
Then there are open neighborhoods \(\mathcal U\) of \(y_\star\) and \(\mathcal V\subset\mathbb S_{++}^N\) of \(K_\star\) such that
\begin{equation*}
(K,y_0)\in\mathcal V\times\mathcal U
\quad\Longrightarrow\quad
F_K^t(y_0)\to z.
\end{equation*}
\end{lemma}

\begin{proof}
Let \(r_z(y)=\sum_i e^{-z_i y_i}\).
Choose a compact SPD neighborhood \(\overline{\mathcal V}\) of \(K_\star\) small enough that, for some \(0<\kappa_-\le\kappa_+<\infty\),
\begin{equation*}
\kappa_-\le\lambda_{\min}(K)
\le\lambda_{\max}(K)\le\kappa_+
\end{equation*}
and, for some \(\delta>0\),
\begin{equation*}
\frac{\eta L_\Phi\kappa_+}{2}\le1-\delta
\end{equation*}
for every \(K\in\overline{\mathcal V}\).
Such a neighborhood exists by the strict baseline inequality and Lemma~\ref{lem:spd-neighborhood}.

The proof of Theorem~\ref{thm:viability-sufficiency} uses candidate-dependent gap and gradient constants \(c_1,c_2,c_3>0\).
On \(\overline{\mathcal V}\), its descent coefficient is uniformly bounded below by
\[
\eta\delta\kappa_- c_3^2>0,
\]
and its one-step radial coefficient is uniformly bounded above by \(\eta\kappa_+\widetilde C_4\).
Therefore one may choose a common \(\bar r>0\) on which the expansion, gradient, and radial-step bounds in that proof all hold.
Every \(K\in\overline{\mathcal V}\) then has a local basin containing
\begin{equation}
\left\{y:r_z(y)<\frac{c_1}{c_2}\bar r\right\}.
\label{eq:common-trap}
\end{equation}

Baseline convergence supplies a finite \(T_0\) with
\begin{equation*}
r_z\!\left(F_{K_\star}^{T_0}(y_\star)\right)
<\frac{c_1}{c_2}\bar r.
\end{equation*}
By Lemma~\ref{lem:finite-horizon-continuity}, the left-hand side remains below the same strict threshold for all \((K,y_0)\) in sufficiently small open neighborhoods \(\mathcal V\times\mathcal U\) of \((K_\star,y_\star)\), with \(\mathcal V\subset\overline{\mathcal V}\).
At time \(T_0\), every such trajectory is in the common basin neighborhood~\eqref{eq:common-trap}, and the local theorem gives endpoint convergence to \(z\).
\end{proof}

The Gram neighborhood lies in the full positive-definite matrix cone, so its perturbations may introduce off-diagonal blocks that couple the block-constant and transverse subspaces.

\subsection{Proof of Theorem~\ref{thm:arbitrary-many-isospectral}}
\label{app:routing-main-proof}

\begin{proof}
The family in Section~\ref{app:m-setup} contains \(m+1\) strictly viable codes.
Choose \(\rho,\alpha,\eta\) by Lemma~\ref{lem:m-parameter-choice} and construct \(K_0,\ldots,K_m\) as in Section~\ref{app:isospectral-construction}.
These Grams are positive definite, pairwise isospectral, and have \(\lambda_{\max}(K_j)=1\).
The parameter choice gives \(\eta(2m+1)<4\), the strict ascent condition for \(N=2m,\tau=1,\lambda=0\), so a single step size is valid for all \(m+1\) routes.

Since \(H_0\mathbf1_m=\mathbf1_m\), Lemma~\ref{thm:m-isotropic} routes \(y_\star=(s_0,\ldots,s_0)\) to \(z_0\).
For \(j\ge1\), \(K_j\) gives block~\(j\) step \(h_1=\eta\alpha\) and other blocks step \(h_2=\eta=\rho h_1\).
Lemma~\ref{thm:m-anisotropic} routes the same \(y_\star\) to \(z_j\).

Apply Lemma~\ref{lem:robust-continuation} to each baseline pair \((K_j,y_\star)\).
It yields open neighborhoods \(\mathcal U_j\) and \(\mathcal V_{m,j}\) with routing to \(z_j\).
Their finite intersection
\[
\mathcal U_m=\bigcap_{j=0}^m\mathcal U_j
\]
is an open neighborhood of \(y_\star\) common to every route, proving Theorem~\ref{thm:arbitrary-many-isospectral}.
\end{proof}

The spectra in the theorem agree at the centers \(K_j\).
The ambient neighborhoods \(\mathcal V_{m,j}\) include perturbations that can change the spectra and break block-subspace invariance.
To preserve both spectra during perturbation, restrict them to the family
\begin{equation*}
\mathcal M_\alpha=
\{I_{2m}-(1-\alpha)Bww^\top B^\top:\lVert w\rVert_2=1\}.
\end{equation*}
Each \(\mathcal V_{m,j}\cap\mathcal M_\alpha\) is a relatively open neighborhood of \(K_j\) with the same routing conclusion and both spectra fixed.
These restricted perturbations preserve the active subspace.

\subsection{Parameters for the \texorpdfstring{\(m=2\)}{m=2} phase portrait}
\label{app:three-code-illustration}

Figure~\ref{fig:routing-capture} instantiates~\eqref{eq:main-isospectral-grams} at \(m=2\) (\(z_0=(+,+,+,+)\), \(z_1=(-,-,+,+)\), \(z_2=(+,+,-,-)\)) with slow vector \(v=(1,-1)^\top/\sqrt2\), \(\alpha=1/6\), \(\eta=0.02\), and \(y_0=-0.8\mathbf1_4\).
On the invariant plane \(y=(u,u,v,v)\), the active blocks are
\[
H_0=\frac12\begin{pmatrix}1+\alpha&1-\alpha\\1-\alpha&1+\alpha\end{pmatrix},
\qquad H_1=\operatorname{diag}(\alpha,1),\qquad
H_2=\operatorname{diag}(1,\alpha).
\]
Here \(H_0\) acts as the identity along \(u=v\) and drives \((u_t,v_t)\) into the positive quadrant \(\{u>0,v>0\}\), where both residuals in~\eqref{eq:m-residual} stay positive and send \((u_t,v_t)\to(+\infty,+\infty)\), whereas \(H_1\) and \(H_2\) enter the absorbing cones \(\mathcal R_1^{(2)}\) and \(\mathcal R_2^{(2)}\) of Lemma~\ref{lem:m-cone-absorption}.

\subsection{Isospectral selection with a rank-deficient shared head}
\label{app:rank-deficient-selection}

The candidate-active directions in Section~\ref{sec:robust-isospectral} also give a shared representation in which predictions cannot vary independently.
Set \(X=B\in\mathbb R^{2m\times m}\), so that the two predictions in each pair are tied.
This restriction preserves every competing explanation, since all candidates assign equal labels within each pair.

\begin{corollary}[Isospectral selection with tied predictions]
\label{cor:rank-deficient-selection}
For every \(m\ge2\) and \(s_0<0\), choose \(\alpha_m,\eta_m,H_j\) as in Theorem~\ref{thm:arbitrary-many-isospectral}.
The same fixed representation \(X=B\), of rank \(m<2m\), admits positive-definite parameter metrics \(M_j=H_j\) whose induced sample Grams \(\widetilde K_j=BM_jB^\top\) satisfy
\[
\operatorname{spec}(M_j)=\{\alpha_m,1^{[m-1]}\},\qquad
\operatorname{spec}(\widetilde K_j)=\{0^{[m]},\alpha_m,1^{[m-1]}\}.
\]
Their candidate-active spectra also coincide with \(\operatorname{spec}(M_j)\).
There are a common open neighborhood \(W\subset\mathbb R^m\) of \(\sqrt2s_0\mathbf1_m\) and open parameter-metric neighborhoods \(O_j\subset\mathbb S_{++}^m\) of \(M_j\) such that every \(w_0\in W\), \(M\in O_j\), and update
\[
w_{t+1}=w_t+\eta_m M B^\top R(Bw_t)
\]
satisfy \(z_{j,i}(Bw_t)_i\to+\infty\) for every coordinate.
Restricting \(O_j\) to \(\{I_m-(1-\alpha_m)uu^\top:\|u\|_2=1\}\) preserves the parameter, full sample, and active spectra exactly.
\end{corollary}

\begin{proof}
Let \(P=BB^\top\) and \(S=\operatorname{im}B\).
For \(y\in S\), the two logits within each pair coincide.
Their posterior marginal targets also coincide because every candidate assigns the same label within that pair.
Thus \(R(y)\in S\), including when different pairs have different logits.
For any \(M\succ0\), its positive-definite lift \(K^+(M)=BMB^\top+I-P\) therefore satisfies
\[
B\bigl[w+\eta_m MB^\top R(Bw)\bigr]
=Bw+\eta_m K^+(M)R(Bw).
\]
The two trajectories agree at every step and remain in \(S\).
At \(M_j=H_j\), the lift is exactly the matrix \(K_j\) in Theorem~\ref{thm:arbitrary-many-isospectral}.
Take the inverse images of its neighborhoods under \(w\mapsto Bw\) and \(M\mapsto K^+(M)\).
They are open, contain the stated centers, and inherit the common step size and routing conclusion.
The neighborhoods can be shrunk to retain the strict ascent bound
\(\eta_m(2m+1)\max\{1,\lambda_{\max}(M)\}<4\).
On \(S\), the singular Gram acts as \(M\), while on \(S^\perp\) it vanishes, giving the spectra.
Finally, \(z_j\in S\) implies that \(w=cB^\top z_j\), \(c>0\), realizes its signs.
\end{proof}

The common initialization set is open in parameter space and relatively open in the attainable prediction space \(S\).
Thus singularity of the sample Gram does not prevent robust selection when the shared representation retains the directions distinguishing the candidates.

\section{Certified Capture and Threshold Sharpness}
\label{app:capture}
\label{app:route-certificate}

We prove Theorem~\ref{thm:routing-certificate} for arbitrary \(\mathcal C\subseteq\{\pm1\}^N\) with \(|\mathcal C|\ge2\) and pairwise Hamming distance at least two.
Throughout this appendix, priors are uniform, \(\tau=1\), and \(\lambda=0\).
The legal evidence in~\eqref{eq:posterior} gives
\[
\Phi(y)=2\log\mathcal E(y),
\qquad
V(y):=-2\log\mathcal E(y).
\]
Above half evidence, the signs of the attained state form a legal code.
Segment ascent prevents subsequent updates from crossing its sign boundaries, and bounds on the residual give the convergence rate.
After the proof, Appendix~\ref{app:half-evidence-sharpness} proves that the threshold \(1/2\) is optimal, Appendices~\ref{app:gap-capture}--\ref{app:fixed-linear-capture} treat posterior concentration without capture, controlled posterior errors, and shared heads of arbitrary rank, and Appendix~\ref{app:multiclass-capture} records the corresponding categorical majority-class barrier.

\begin{lemma}[Conditional slices and legality above one half]
\label{lem:half-evidence-legality}
Write \(s_i=\sigmoid(y_i)\).
For each coordinate \(i\), let \(A_i\) and \(B_i\) be the probabilities of legal completions of the other coordinates when coordinate \(i\) is fixed to \(+1\) and \(-1\), respectively.
Then
\begin{equation}
A_i+B_i\le1,\qquad
\mathcal E=s_iA_i+(1-s_i)B_i\le\max\{s_i,1-s_i\}.
\label{eq:conditional-evidence-barrier}
\end{equation}
In particular, \(\mathcal E(y)>1/2\) implies that every coordinate has a strict sign and that \(\operatorname{sign}(y)\in\mathcal C\).
\end{lemma}

\begin{proof}
The two completion events are disjoint.
A common completion would produce two legal codes differing only in coordinate~\(i\), violating the distance condition.
This proves the first inequality, including for degenerate product distributions, and the second follows by conditioning on coordinate~\(i\).
If \(y_i=0\), the bound gives \(\mathcal E\le1/2\), so evidence above one half excludes every sign tie.

Orient each coordinate toward its majority bit \(z_i=\operatorname{sign}(y_i)\) and write \(p_i=\sigmoid(z_i y_i)>1/2\).
For this orientation let \(A_i\) denote the probability of a legal completion with bit \(z_i\), and \(B_i\) that with bit \(-z_i\).
If \(A_i\le B_i\), then
\[
\mathcal E=\frac{A_i+B_i}{2}
 +(p_i-\tfrac12)(A_i-B_i)\le\tfrac12,
\]
a contradiction.
Thus setting \(p_i\) to one increases the evidence.
Repeat this operation coordinate by coordinate.
The evidence remains above one half, and disjointness continues to hold at each intermediate product distribution.
The final distribution is concentrated at \(z\), so its evidence equals the indicator of \(z\in\mathcal C\).
This indicator must be one because rounding has kept the evidence above one half.
\end{proof}

\begin{lemma}[Evidence along an entire update segment]
\label{lem:capture-segment}
Let \(y_{t+1}=y_t+\eta K_tR(y_t)\) with \(0\preceq K_t\preceq\kappa_+I\), fixed \(\eta>0\), and \(\beta_0=1-\eta\kappa_+(N+1)/4>0\).
For every \(u\in[0,1]\),
\begin{equation*}
\Phi\bigl(y_t+u\eta K_tR(y_t)\bigr)-\Phi(y_t)
\ge u\eta\beta_0\lVert R(y_t)\rVert_{K_t}^2.
\end{equation*}
Consequently, if \(c=\mathcal E(y_T)>1/2\), every subsequent update segment has evidence at least \(c\) and stays in the hard orthant of \(y_T\).
\end{lemma}

\begin{proof}
Lemma~\ref{lem:global-smoothness} gives \(L=(N+1)/2\).
At one step, write \(R=R(y_t)\) and \(K=K_t\).
The spectral upper bound gives \(\lVert KR\rVert_2^2\le\kappa_+\lVert R\rVert_K^2\), so lower smoothness yields
\[
\Phi(y_t+u\eta KR)-\Phi(y_t)
\ge u\eta\left(1-\frac{u\eta L\kappa_+}{2}\right)
\lVert R\rVert_K^2
\ge u\eta\beta_0\lVert R\rVert_K^2.
\]
Thus the evidence on the segment is no smaller than at its start.
By~\eqref{eq:conditional-evidence-barrier}, no segment with evidence above one half intersects \(y_i=0\).
Induction proves the claim.
The inequality uses only the current spectral bound and therefore also holds for metrics chosen from the preceding history.
\end{proof}

These conditional slices bound the potential deficit and residual throughout the captured region, allowing convergence estimates from capture without a separate local trapping argument.

\begin{lemma}[Global gap and gradient bounds after capture]
\label{lem:capture-global-bounds}
Let \(z=\operatorname{sign}(y)\in\mathcal C\) and \(\mathcal E(y)\ge c>1/2\).
Set \(a_i=e^{-z_i y_i}\), \(r=\sum_i a_i\), and \(d=2c^2(2c-1)\).
Then
\begin{equation*}
0<a_i\le\frac{1-c}{c},\qquad
\frac{2c}{N}r\le V(y)\le2r,\qquad
\lVert R(y)\rVert_2\ge\frac d{\sqrt N}r.
\end{equation*}
Moreover, \(z_iR_i(y)\ge d a_i>0\) for every coordinate.
\end{lemma}

\begin{proof}
Put \(p_i=\sigmoid(z_i y_i)\).
Equation~\eqref{eq:conditional-evidence-barrier} gives \(p_i\ge\mathcal E\ge c\), proving the coordinate bound.
With the conditional slices oriented toward \(z_i\), the proof of Lemma~\ref{lem:half-evidence-legality} gives \(A_i>B_i\).
Moreover,
\[
\mathcal E
=\frac{A_i+B_i}{2}+(p_i-\tfrac12)(A_i-B_i)
\le\frac12+\frac12(A_i-B_i),
\]
so \(A_i-B_i\ge2c-1\).
Direct differentiation, together with \(\mathcal E\le1\), gives
\[
z_iR_i(y)
=\frac{2p_i(1-p_i)(A_i-B_i)}{\mathcal E}
\ge2c^2(2c-1)a_i=d a_i.
\]
Here \(p_i(1-p_i)=p_i^2a_i\ge c^2a_i\).
Taking norms gives the gradient bound since \(\lVert a\rVert_2\ge r/\sqrt N\).

Since \(z\) is legal, its product probability contributes to \(\mathcal E\), and hence
\[
V\le2\sum_i\log(1+a_i)\le2r.
\]
Conversely, \(\mathcal E\le p_i\) implies \(V\ge2\log(1+a_i)\ge2a_i/(1+a_i)\ge2c a_i\) for every~\(i\).
At least one coordinate satisfies \(a_i\ge r/N\), so the bound for that coordinate yields \(V\ge(2c/N)r\).
\end{proof}

\begin{proof}[Proof of Theorem~\ref{thm:routing-certificate}]
\phantomsection\label{proof:half-evidence-capture}
Lemma~\ref{lem:half-evidence-legality} identifies a unique legal hard code \(z\) at time \(T\).
Lemma~\ref{lem:capture-segment} keeps this code fixed and gives
\[
V_t-V_{t+1}\ge\eta\beta_0\kappa_-\lVert R(y_t)\rVert_2^2.
\]
Apply Lemma~\ref{lem:capture-global-bounds} with \(c=\mathcal E(y_T)>1/2\), and set
\[
\gamma=\frac{\eta\beta_0\kappa_-d^2}{4N}
=\frac{\eta\beta_0\kappa_-c^4(2c-1)^2}{N}.
\]
The gap and gradient estimates imply
\[
V_{t+1}\le V_t-\frac{\eta\beta_0\kappa_-d^2}{N}r_t^2
\le V_t-\gamma V_t^2.
\]
The distance condition makes \(\mathcal C\) a proper subset of the cube.
Finite logits assign positive probability to every vertex, so \(0<\mathcal E(y_t)<1\) and \(V_t>0\).
Therefore \(1-\gamma V_t>0\), and the last recurrence yields \(V_{t+1}^{-1}\ge V_t^{-1}+\gamma\).
Summing gives the explicit bound
\begin{equation}
r_t\le\frac{N}{2c}\,
\frac{1}{V_T^{-1}+\gamma(t-T)},\qquad t\ge T.
\label{eq:capture-global-rate}
\end{equation}
Its constants depend on the initial evidence and the uniform metric bounds, not on the chosen continuation.
Thus \(r_t\to0\), and every signed margin diverges to \(+\infty\).

Conversely, approaching \(z\in\mathcal C\) concentrates the independent prediction distribution at that legal code.
Hence \(\mathcal E(y_t)\to1\), so the trajectory must cross the threshold~\eqref{eq:half-evidence-capture} at a finite iteration.
\end{proof}

\subsection{Optimality of the universal capture threshold}
\label{app:half-evidence-sharpness}

The threshold in Theorem~\ref{thm:routing-certificate} is the smallest universal threshold based only on legal evidence.
Even a legal assignment with evidence arbitrarily close to one half from below can change under an admissible native update.

\begin{proposition}[Sharpness of the half-evidence threshold]
\label{prop:half-evidence-sharpness}
For every \(c<1/2\), there is a finite state with a legal hard assignment and \(c<\mathcal E<1/2\) whose next native exact update changes that assignment to another legal assignment while legal evidence increases.
The candidate set, positive-definite Gram and admissible step size can be fixed independently of \(c\).
\end{proposition}

\begin{proof}
Take \(N=3\), \(\mathcal C=\{(+,+,+),(-,-,+)\}\), and
\[
u=(1/2,1/2,1/\sqrt2)^\top,\qquad
K=4I-\frac{15}{4}uu^\top,\qquad \eta=\frac15.
\]
The candidates differ in two labels, \(K\) has spectrum \(\{1/4,4,4\}\), and \(\eta\lambda_{\max}(K)(N+1)=16/5<4\).
For \(0<\varepsilon<1/4\), set \(a=\varepsilon^2\), \(b=\log((1-\varepsilon)/\varepsilon)\) and \(y=(a,a,b)^\top\).
Its hard assignment is \((+,+,+)\), with evidence
\[
\mathcal E(y)=(1-\varepsilon)\bigl(\sigma(a)^2+[1-\sigma(a)]^2\bigr).
\]
Since \(0<\sigma(a)-1/2\le a/4\),
\[
\frac{1-\varepsilon}{2}\le\mathcal E(y)
\le(1-\varepsilon)\left(\frac12+\frac{\varepsilon^4}{8}\right)<\frac12.
\]
Choosing \(\varepsilon<\min\{1/4,1-2c\}\) therefore gives \(\mathcal E(y)>c\).
The posterior marginals are \((\sigma(2a),\sigma(2a),1)\), so the native residual is \(R(y)=(r,r,2\varepsilon)^\top\), where \(r=2[\sigma(2a)-\sigma(a)]\) satisfies \(0<r\le a/2\).
The first two updated logits equal
\[
a^+=a+\frac{17}{40}r-\frac{3\varepsilon}{4\sqrt2}
\le\frac{97}{80}\varepsilon^2-\frac{3\varepsilon}{4\sqrt2}<0,
\]
while the third obeys
\[
b^+=b+\frac{17}{20}\varepsilon-\frac{3r}{4\sqrt2}
\ge b+\frac{17}{20}\varepsilon-\frac{3\varepsilon^2}{8\sqrt2}>0.
\]
Thus the next hard assignment is the other legal code \((-,-,+)\).
The step satisfies the same strict ascent condition as Theorem~\ref{thm:routing-certificate}, so this change occurs while legal evidence increases.
\end{proof}

The pair's signed logits are of order \(\varepsilon^2\), whereas the third concept supplies a residual of order \(\varepsilon\).
Coupling this residual into the nearly tied pair can therefore reverse its labels even as evidence approaches one half.
This result concerns a uniform guarantee over the theorem's model class; a particular trajectory can become irreversible earlier, which we examine in Appendix~\ref{app:capture-timing}.

\subsection{Posterior concentration without capture}
\label{app:gap-capture}

\begin{example}[Posterior concentration without capture]
\label{ex:gap-not-capture}
Take \(\mathcal C=\{z^+,z^-\}\), where \(z^+=\one_6\), \(z^-=(-1,-1,1,1,1,1)\), with native parameters and \(y_0=(1.1,1.1,0.01,0.01,0.01,0.01)\).
The hard code is \(z^+\), and
\begin{align*}
\log(q_+/q_-)&=2.2>2,\qquad q_+=0.9002495\ldots,\\
\mathcal E(y_0)&=[\sigma(1.1)^2+\sigma(-1.1)^2]\sigma(0.01)^4
=0.0398662\ldots.
\end{align*}
The four common coordinates cancel from the posterior odds but reduce the legal evidence.
Define
\[
v_0=(0,0,1,-1,0,0)^\top/\sqrt2,\qquad
v_1=(\tfrac12,\tfrac12,\tfrac1{2\sqrt2},\tfrac1{2\sqrt2},
\tfrac1{2\sqrt2},\tfrac1{2\sqrt2})^\top,
\qquad K_j=I-\tfrac45v_jv_j^\top.
\]
Both Grams have spectrum \(\{1/5,1,1,1,1,1\}\).
With \(\eta=2/7\), the ascent quantity is \(\eta\lambda_{\max}(K_j)(N+1)=2<4\), yet \(K_0\) selects \(z^+\) and \(K_1\) selects \(z^-\) from this state.
The Grams share their full six-dimensional spectrum but differ in their candidate-active compressions.
\end{example}

\begin{proof}
The subspace \(y=(s,s,r,r,r,r)\) is invariant.
Its residual coordinates are \(R_a(s)=2[\sigma(2s)-\sigma(s)]\) and \(R_b(r)=2\sigma(-r)\).
On this subspace \(K_0\) acts as identity.
For \(s>0\), both residuals are positive, so a finite coordinate limit would leave its increment bounded away from zero.
Hence \(s,r\to+\infty\).
Under \(K_1\) the exact update is
\begin{align*}
s^+&=s+\eta(3R_a/5-2\sqrt2 R_b/5),\\
r^+&=r+\eta(3R_b/5-\sqrt2 R_a/5).
\end{align*}
Seventy-digit interval propagation of the full six-dimensional map gives
\begin{align*}
s_{17}&\in[-0.04414193676449,-0.04414193676448],\\
r_{17}&\in[1.54375619445710,1.54375619445711].
\end{align*}
The region \(s<0<r\) is invariant because \(R_a<0<R_b\): \(s\) strictly decreases and \(r\) strictly increases.
A finite negative limit of \(s\) would keep its negative drift away from zero.
A finite limit of \(r\) would keep its positive drift away from zero.
Thus \(s\to-\infty\) and \(r\to+\infty\), proving selection of \(z^-\).
\end{proof}

Along \(K_0\), the half-evidence test first passes numerically at step 15 (\(\mathcal E(y_{14})=0.4870554\ldots\), \(\mathcal E(y_{15})=0.5074577\ldots\)).
Switching to \(K_1\) after capture preserves \(z^+\), as Theorem~\ref{thm:routing-certificate} guarantees.

\subsection{Capture under changing metrics and controlled errors}
\label{app:inexact-capture}

Theorem~\ref{thm:routing-certificate} extends to controlled posterior-refresh errors under the same uniform positive spectral bounds on the sample Grams.
An error budget relative to the current exact residual preserves ascent along each update segment and prevents crossing the half-evidence barrier.

\begin{theorem}[Half-evidence capture with controlled errors]
\label{thm:inexact-capture}
Let \(|\mathcal C|\ge2\) and \(d_{\rm H}(z,w)\ge2\) for all distinct \(z,w\in\mathcal C\).
Use uniform priors, \(\tau=1\), and \(\lambda=0\).
After any finite prefix, allow the continuation
\begin{equation*}
y_{t+1}=y_t+\eta K_t[R(y_t)+e_t],\qquad
0<\kappa_-I\preceq K_t\preceq\kappa_+I,
\end{equation*}
with fixed \(\eta>0\), \(0<\kappa_-\le\kappa_+<\infty\), and \(\delta\ge0\) satisfying
\begin{equation*}
\lVert e_t\rVert_{K_t}\le\delta\lVert R(y_t)\rVert_{K_t},\qquad
\beta:=1-\delta-\frac{\eta\kappa_+(N+1)}4(1+\delta)^2>0.
\end{equation*}
If the finite state \(y_T\) satisfies
\begin{equation}
\mathcal E(y_T)>\tfrac12,
\label{eq:inexact-half-evidence-capture}
\end{equation}
then \(z=\operatorname{sign}(y_T)\) is a well-defined member of \(\mathcal C\).
For all \(t\ge T\), \(\operatorname{sign}(y_t)=z\), and \(r_t:=\sum_i e^{-z_i y_{i,t}}=O((t-T+1)^{-1})\), so every signed margin diverges to \(+\infty\).
The Gram matrices \(K_t\) and errors \(e_t\) may depend arbitrarily on the training history.
Conversely, every trajectory approaching a code endpoint eventually satisfies \eqref{eq:inexact-half-evidence-capture}.
\end{theorem}

\begin{proof}
Here \(\Phi=2\log\mathcal E\), so its deficit is \(V=-2\log\mathcal E\).
Lemmas~\ref{lem:half-evidence-legality} and~\ref{lem:capture-global-bounds} remain unchanged.
We extend the ascent estimate to include the error along the entire update segment.

Lemma~\ref{lem:global-smoothness} gives \(L=(N+1)/2\).
At one step abbreviate \(R=R(y_t)\), \(K=K_t\), and \(e=e_t\).
Metric-norm Cauchy--Schwarz gives \(\langle R,Ke\rangle\ge-\delta\lVert R\rVert_K^2\), while
\[
\lVert K(R+e)\rVert_2^2
\le\lambda_{\max}(K)\lVert R+e\rVert_K^2
\le\kappa_+(1+\delta)^2\lVert R\rVert_K^2.
\]
For \(u\in[0,1]\), lower smoothness yields
\begin{align*}
\Phi(y_t+u\eta K(R+e))-\Phi(y_t)
&\ge u\eta\left[1-\delta
 -\frac{u\eta L\kappa_+}{2}(1+\delta)^2\right]\lVert R\rVert_K^2\\
&\ge u\eta\beta\lVert R\rVert_K^2.
\end{align*}
Thus the evidence along each update segment stays at or above its starting value, so by~\eqref{eq:conditional-evidence-barrier} no segment with evidence above one half crosses \(y_i=0\), regardless of how \(K_t\) and \(e_t\) depend on the history.
Applying Lemma~\ref{lem:capture-global-bounds} with \(c=\mathcal E(y_T)>1/2\) yields the exact recurrence of~\eqref{eq:capture-global-rate} with \(\beta_0\) replaced by \(\beta\), giving \(\gamma=\eta\beta\kappa_-c^4(2c-1)^2/N>0\), uniform \(O((t-T+1)^{-1})\) convergence, and eventual passage whenever \(y_t\) approaches a legal corner.
\end{proof}

\subsection{Capture with limited control over individual predictions}
\label{app:fixed-linear-capture}

Corollary~\ref{cor:fixed-linear-capture} covers exact updates with a fixed linear representation of arbitrary rank.

\begin{proof}[Proof of Corollary~\ref{cor:fixed-linear-capture}]
Set \(K_t=XM_tX^\top\) and \(\kappa_+=m_+\|X\|_{\mathrm{op}}^2\).
These matrices satisfy \(0\preceq K_t\preceq\kappa_+I\), even when \(X\) is rank deficient.
Lemma~\ref{lem:capture-segment} therefore preserves the legal code \(z=\operatorname{sign}(Xw_T)\) and the evidence lower bound \(c=\mathcal E(Xw_T)>1/2\).
The proof uses \(K_t^2\preceq\kappa_+K_t\), which holds for positive-semidefinite matrices even when some sample-space eigenvalues vanish.

Let \(\mu=\min_i|(Xw_T)_i|>0\) and \(d_c=2c^2(2c-1)\).
The captured head is nonzero.
By the coordinatewise bound in Lemma~\ref{lem:capture-global-bounds}, each exact residual has the same sign as the corresponding coordinate of \(Xw_T\).
Hence
\begin{align*}
\|w_T\|_2\|X^\top R_t\|_2
&\ge (Xw_T)^\top R_t
=\sum_i|(Xw_T)_i|\,|R_{i,t}|\\
&\ge\mu\|R_t\|_1\ge\mu d_c r_t,\\
R_t^\top K_tR_t
&\ge m_-\|X^\top R_t\|_2^2
\ge\frac{m_-\mu^2d_c^2}{\|w_T\|_2^2}r_t^2.
\end{align*}
Thus \(X^\top R_t\) cannot vanish in the captured region.
The separating head \(w_T\) supplies a quantitative lower bound for the captured labels, which need not agree with the true semantic labels.

Put \(V_t=-2\log\mathcal E(Xw_t)\) and \(\beta_0=1-\eta\kappa_+(N+1)/4\).
Segment ascent and \(V_t\le2r_t\) give
\[
V_{t+1}\le V_t-\gamma_X V_t^2,\qquad
\gamma_X=\frac{\eta\beta_0m_-\mu^2d_c^2}{4\|w_T\|_2^2}>0.
\]
Every finite iterate remains finite because \(|R_i|\le2\), \(X\) is fixed, and \(M_t\preceq m_+I\).
Thus \(V_t>0\), and the same reciprocal recurrence as in~\eqref{eq:capture-global-rate} yields
\begin{equation}
r_t\le\frac{N}{2c}\,
\frac{1}{V_T^{-1}+\gamma_X(t-T)},\qquad t\ge T.
\label{eq:fixed-linear-rate}
\end{equation}
For this captured head, the constants are uniform over all admissible continuations.
A trainable bias is a column of \(X\).
The representation is exactly \(y=Xw\), with no fixed sample-specific offset.
\end{proof}

\paragraph{Disjoint-bag step bound.}
If the coordinates partition into bags \(I_j\) of size at most \(b\), with \(\mathcal C=\prod_j\mathcal C_j\), then \(\Phi(y)=2\sum_j\log\mathcal E_j(y_{I_j})\).
The Hessian is block diagonal, so applying Lemma~\ref{lem:global-smoothness} to each bag gives
\[
\|\nabla^2\Phi(y)\|_{\mathrm{op}}
=\max_j\|\nabla^2(2\log\mathcal E_j)(y_{I_j})\|_{\mathrm{op}}
\le\frac{b+1}{2}.
\]
Consequently, the corollary and~\eqref{eq:fixed-linear-rate} hold with \(\eta m_+\|X\|_{\mathrm{op}}^2(b+1)<4\) and \(\beta_0=1-\eta m_+\|X\|_{\mathrm{op}}^2(b+1)/4\).
The threshold remains the global product \(\prod_j\mathcal E_j>1/2\), and the gap constants still use the full \(N\).
Only the constraints must factor by bag, whereas \(X\) and \(M_t\) may couple different bags.

\subsection{The categorical half-evidence barrier}
\label{app:multiclass-capture}

The static half-evidence boundary of Lemma~\ref{lem:half-evidence-legality} extends directly to categorical concepts.
Let concept \(i\) take values in \(\{1,\ldots,K_i\}\) under independent categorical predictions \(p_{ik}\), and let \(\mathcal C\subseteq\prod_{i=1}^N\{1,\ldots,K_i\}\) be a finite nonempty set of legal explanations.

\begin{lemma}[Categorical half-evidence boundary]
\label{lem:categorical-barrier}
Suppose distinct candidates in \(\mathcal C\) differ in at least two concept coordinates, and put \(E(p)=\sum_{z\in\mathcal C}\prod_i p_{i,z_i}\).
For every product distribution \(p\) and coordinate \(i\), \(E(p)\le\max_k p_{ik}\).
If \(E(p)>1/2\), each coordinate has a unique majority class and the tuple of majority classes belongs to \(\mathcal C\); at every tie for the largest class probability on any coordinate, \(E(p)\le1/2\).
\end{lemma}

\begin{proof}
Let \(A_{ik}\) be the probability of legal completions of the other coordinates when coordinate \(i\) takes class \(k\).
These completion events are disjoint across \(k\), since a shared completion would produce two legal candidates differing only at coordinate \(i\).
Thus \(\sum_k A_{ik}\le1\) and \(E(p)=\sum_k p_{ik}A_{ik}\le\max_k p_{ik}\).
At any tie for the largest class probability on coordinate \(i\), that maximum is at most \(1/2\), giving \(E(p)\le1/2\).
When \(E(p)>1/2\), coordinate \(i\) has a unique majority class \(k_*\) with \(p=p_{i,k_*}>1/2\) and completion probability \(A=A_{i,k_*}\).
If \(A\le1/2\), then \(E(p)\le pA+(1-p)(1-A)\le1/2\), a contradiction.
Hence \(A>1/2\), while every other completion probability is at most \(1-A<A\), so replacing \(p_i\) by a point mass at \(k_*\) does not decrease \(E\).
Repeating this rounding across all coordinates retains evidence above \(1/2\) and concentrates the distribution on the majority tuple, which must therefore belong to \(\mathcal C\).
\end{proof}

\section{Suboptimal Attraction and Recovery under Identifiable Supervision}
\label{app:identifiable-recovery}

Appendix~\ref{app:identifiable-minima} proves Proposition~\ref{prop:identifiable-minimum} and extends its finite wrong stationary set to regular pairwise count-query graphs on fixed image features (Proposition~\ref{prop:regular-query-minimum}).
Appendix~\ref{app:global-recovery} characterizes global recovery across query weights and derives a label-free criterion for restoring the original weights after temporary reweighting (Proposition~\ref{prop:global-recovery} and Corollary~\ref{cor:safe-restoration}), while Appendix~\ref{app:wrong-basin-certificate} constructs the explicit contracting ball used in Section~\ref{sec:identifiable-counts}.

\subsection{Suboptimal attracting states under identifiable queries}
\label{app:identifiable-minima}

We first prove Proposition~\ref{prop:identifiable-minimum} in a two-parameter prototype and then extend the mechanism to regular count-query graphs on fixed image features.
In both settings, abduction conditions separately on each pair's count while repeated inputs share predictions through the predictor; jointly normalizing over one label variable per distinct input would instead leave the true assignment as the sole candidate.

\begin{proof}[Proof of Proposition~\ref{prop:identifiable-minimum}]
Let a shared predictor assign positive-label probabilities \(p_0=\sigma(w_0)\) and \(p_1=\sigma(w_1)\) to true classes \(0\) and \(1\).
Use binary sum supervision with total weight \(\rho\) on the two mixed class pairs and weight \((1-\rho)/2\) on each homogeneous pair.
The mean negative log evidence is
\[
\mathcal L_\rho(p_0,p_1)
=-(1-\rho)\log(1-p_0)-(1-\rho)\log p_1
-\rho\log\bigl[p_0(1-p_1)+(1-p_0)p_1\bigr].
\]
For \(0<\rho<1\), its unique zero-loss probability endpoint is \((0,1)\), since the homogeneous queries force both correct labels.
At \((p_0,p_1)=(s,1-s)\), put \(D=s^2+(1-s)^2\).
The probability gradient is \((g,-g)\), where
\[
g=\frac{1-\rho}{1-s}-\frac{\rho(2s-1)}D
=\frac{2s^2-(2+\rho)s+1}{(1-s)D}.
\]
For \(2\sqrt2-2<\rho<1\), define
\[
\Delta=\sqrt{\rho^2+4\rho-4},\qquad
s_+=\frac{2+\rho+\Delta}{4}\in(1/2,1).
\]
At \((s_+,1-s_+)\), the full two-dimensional probability Hessian has eigenvalues
\[
\lambda_- =\frac{\Delta}{(1-s_+)D}>0,\qquad
\lambda_+ =\frac{1-\rho}{(1-s_+)^2}+\frac{2\rho}{D}>0.
\]
The logit Hessian at this stationary point is \([s_+(1-s_+)]^2\) times the probability Hessian, so it is also positive definite.
Thus ordinary gradient descent on \(\mathcal L_\rho\) has a locally asymptotically stable wrong fixed point for every sufficiently small positive step.
Its loss is positive, whereas the correct endpoint has loss approaching zero.

To obtain a finite native instance, take \(\rho=0.9\) and 20 disjoint pairs: one \((0,0)\), one \((1,1)\), nine \((0,1)\), and nine \((1,0)\).
There are \(N=40\) occurrence coordinates, with semantically oriented feature row \(-e_0^\top\) for true class 0 and \(e_1^\top\) for true class 1.
The resulting \(X\) has rank two and \(X^\top X=20I_2\).
Each mixed pair has two legal assignments differing in both labels, so the product candidate set has \(2^{18}\) codes and minimum Hamming distance two.
The homogeneous pairs force every shared class prediction, giving \(\mathcal C_X=\{\one_{40}\}\).
The global potential and parameter update are
\[
\Phi(Xw)=-40\mathcal L_{0.9}(w),\qquad
w^+=w-40\eta\nabla_w\mathcal L_{0.9}(w).
\]
At the wrong fixed point, \((p_0,p_1)=(0.8850781\ldots,0.1149219\ldots)\), and
\[
\mathcal E=(1-s_+)^4D^{18}=2.90834\ldots\times10^{-6}<1/2.
\]
Its hard predictions violate both homogeneous queries.
At \(\eta=0.001\), the update eigenvalues are approximately \(0.997105\) and \(0.995931\), and \(\eta\|X\|_{\mathrm{op}}^2(N+1)=0.82<4\).
\end{proof}

\paragraph{Regular count-query graphs.}
Consider a balanced set of \(n\) distinct inputs with true labels \(c_i\in\{0,1\}\) and the following pairings.
Each input participates in one same-class pair (label sum \(0\) or \(2\), forming \(n/2\) disjoint pairs that directly identify all \(n\) labels) and \(k\ge1\) mixed-class pairs (label sum \(1\), forming a \(k\)-regular bipartite graph \(G=([n],E_G)\) with \(m=nk/2\) edges between opposite classes).
Let \(F\in\mathbb R^{n\times d}\) collect the fixed feature vectors of the \(n\) distinct inputs.
When \(\operatorname{rank}(F)=n\), any \(n\) input logits are realizable.
By contrast, the occurrence-level design across all \(n/2+m\) pair appearances has one row per input occurrence: repeated inputs give repeated rows, so the sample Gram is singular.
For a shared linear head \(w\in\mathbb R^d\), write \(v_i=\sigma((2c_i-1)(Fw)_i)\) for the predicted probability of the true label \(c_i\) on input \(i\).
Assigning weight \(\rho\in[0,1)\) to the mean mixed-pair negative log evidence and weight \(1-\rho\) to the mean same-class-pair negative log evidence defines the objective (written interchangeably in terms of \(w\) or \(v=(v_1,\ldots,v_n)\in(0,1)^n\))
\begin{equation}
\mathcal L_\rho(w)
=-\frac{2(1-\rho)}n\sum_{i=1}^n\log v_i
-\frac\rho m\sum_{\{i,j\}\in E_G}
\log\bigl[v_iv_j+(1-v_i)(1-v_j)\bigr],
\label{eq:identifiable-graph-loss}
\end{equation}
whose same-class term makes \(v=\one_n\) the unique zero-loss probability endpoint whenever \(\rho<1\).

\begin{proposition}[Suboptimal attraction on regular query graphs]
\label{prop:regular-query-minimum}
For the regular count-query design above, suppose \(2\sqrt2-2<\rho<1\) and set \(v_*=(2-\rho-\sqrt{\rho^2+4\rho-4})/4\).
Then \(v_*\one_n\) is a strictly suboptimal local minimum of \(\mathcal L_\rho\) in all \(n\) prediction coordinates, and every hard label is incorrect.
For any full-row-rank feature matrix \(F\in\mathbb R^{n\times d}\), a finite head \(w_*\) realizes this center, and \(w_*+\ker F\) is a locally attracting stationary set under ordinary parameter gradient descent with sufficiently small positive step size: nearby heads converge to its member sharing their initial null-space component.
\end{proposition}

\begin{proof}
Write \(\Delta=\sqrt{\rho^2+4\rho-4}\) and \(D=v_*^2+(1-v_*)^2\).
The center satisfies \(v_*=1-s_+\in(0,1/2)\), with \(s_+\) as in the prototype.
At \(v=v_*\one_n\), every coordinate gradient equals
\[
\frac2n\left[-\frac{1-\rho}{v_*}-\rho\frac{2v_*-1}D\right]=0.
\]
Let \(A_G\) be the adjacency matrix and \(L_G=kI_n-A_G\) the graph Laplacian.
An adjacent off-diagonal Hessian entry is \(-\rho/(mD^2)\), and each diagonal entry is
\[
\frac{2(1-\rho)}{nv_*^2}+\frac{k\rho(2v_*-1)^2}{mD^2}.
\]
Using \(k/m=2/n\) and the stationary equation gives the full Hessian
\begin{equation}
H_v:=\nabla_v^2\mathcal L_\rho(v_*\one_n)
=\frac{2\Delta}{nv_*D}I_n+\frac{\rho}{mD^2}L_G\succ0.
\label{eq:identifiable-graph-hessian}
\end{equation}
Because \(L_G\) is positive semidefinite and annihilates the constant vector, \(\lambda_{\min}(H_v)=2\Delta/(nv_*D)\), including disconnected graphs.
For semantic logits \(u_i=\log(v_i/(1-v_i))\), stationarity gives
\[
H_u=[v_*(1-v_*)]^2H_v\succ0.
\]
This proves strict local minimality in all \(n\) prediction directions for the homogeneous reversal branch.

\paragraph{Real image features and redundant parameters.}
Let \(F\in\mathbb R^{n\times d}\) be any full-row-rank image-feature matrix, including a bias if used, with raw logits \(\ell=Fw\).
With \(S=\operatorname{diag}(2c_i-1)\), the semantic logits are \(u=S\ell\).
The wrong center has \(\ell_*=S\one_n\log(v_* /(1-v_*))\), which a finite head \(w_*\) realizes by full row rank.
Its raw-logit Hessian is \(H_\ell=SH_uS\succ0\), and its parameter Hessian is \(F^\top H_\ell F\).
For a thin singular-value decomposition \(F=U\Sigma V^\top\), its restriction to the full effective parameter space \(\operatorname{row}(F)\) is
\[
H_{\mathrm{eff}}=\Sigma U^\top H_\ell U\Sigma\succ0.
\]
The remaining \(d-n\) directions form \(\ker F\) and leave every training prediction unchanged.
Ordinary parameter gradients lie in \(\operatorname{row}(F)\), so each trajectory preserves its initial null-space component.
For \(0<\alpha<2/\lambda_{\max}(H_{\mathrm{eff}})\), the effective update Jacobian \(I-\alpha H_{\mathrm{eff}}\) has spectral radius below one.
Hence the affine stationary set \(w_*+\ker F\) is locally attracting: nearby heads converge to the member with their own initial null-space component, and their predictions converge to the wrong center.
\end{proof}

\paragraph{Native occurrence representation and step size.}
For any real \(\rho\in[0,1)\), a gradient descent step \(-\alpha\nabla_w\mathcal L_\rho(w)\) on~\eqref{eq:identifiable-graph-loss} equals the \(\rho\)-weighted sum of the native per-query detached soft-target cross-entropy updates across the same-class and mixed queries.
When \(\rho\in(0,1)\) is rational, repeating queries an integer number of times so that their normalized frequencies match the weights \(1-\rho\) and \(\rho\) in~\eqref{eq:identifiable-graph-loss} realizes this weighted update directly as an unweighted native instance of Section~\ref{sec:dynamics} with \(|\mathcal C|\ge2\).
Specifically, if there are \(B\) repeated queries, let \(X\) contain their semantically oriented occurrence feature rows and let \(\mathcal C\) be the product of their legal candidate sets.
Every image has total normalized exposure \(2/n\), giving
\[
\Phi(Xw)=-2B\mathcal L_\rho(w),\qquad
X^\top X=\frac{2B}{n}F^\top F,\qquad
\eta=\frac{\alpha}{2B}.
\]
Thus ordinary descent with step \(\alpha\) on the mean loss is exactly the native parameter update with step \(\eta\).
Distinct complete candidates differ in at least two occurrence labels.
The homogeneous matching uniquely fixes every shared image label, and full row rank realizes the correct signs, so \(\mathcal C_X=\{\one_{2B}\}\).
The occurrence Gram is singular even though \(F\) has full row rank on distinct images.
For the dense experimental step \(\alpha=n/(1.5\|F\|_{\mathrm{op}}^2)\), the two-image bag bound above gives
\[
\eta\|X\|_{\mathrm{op}}^2(2+1)
=\frac{3\alpha\|F\|_{\mathrm{op}}^2}{n}=2<4.
\]
The wrong center violates every homogeneous count and lies below the half-evidence barrier.

\subsection{Regular pairwise counts and safe restoration}
\label{app:global-recovery}

\begin{proposition}[Recovery and wrong attraction on regular pairwise counts]
\label{prop:global-recovery}
In the regular pairwise count-query model~\eqref{eq:identifiable-graph-loss}, assume \(\operatorname{rank}(F)=n\), set \(\rho_c=2\sqrt2-2\) and \(\beta=\|F\|_{\mathrm{op}}^2/(2n)\), and run exact full-batch gradient descent \(w_{t+1}=w_t-\alpha\nabla\mathcal L_\rho(w_t)\) with fixed step size satisfying \(0<\alpha\beta<2\).
\begin{enumerate}[label=(\roman*)]
\item If \(0\le\rho<\rho_c\), then \(\min_i v_{i,t}\to1\) from every finite initialization \(w_0\in\mathbb R^d\), recovering all true concept labels.
\item If \(\rho=\rho_c\), a finite stationary set with all concept labels wrong exists (realizing \(v_i=(2-\rho_c)/4<1/2\)), so global recovery from arbitrary finite initializations fails.
\item If \(\rho_c<\rho<1\), the affine stationary set realizing \(v_*\one_n\), where \(v_*=(2-\rho-\sqrt{\rho^2+4\rho-4})/4<1/2\), is strictly suboptimal and locally asymptotically stable, with every hard concept label wrong.
\end{enumerate}
\end{proposition}

Under equal weight per query, \(\rho=k/(k+1)\), so up to four mixed queries per input (\(k\le4\)) guarantee recovery from every finite initialization, whereas five or more (\(k\ge5\)) create a locally attracting all-wrong state.

\begin{proof}[Proof of Proposition~\ref{prop:global-recovery}]
Write \(S=\operatorname{diag}(2c_i-1)\) and \(u=SFw\), so \(v_i=\sigma(u_i)\).
The loss of a mixed edge in~\eqref{eq:identifiable-graph-loss} equals \(\operatorname{softplus}(u_i)+\operatorname{softplus}(u_j)-\operatorname{softplus}(u_i+u_j)\).
Regularity gives
\[
\frac n2\partial_{u_i}\mathcal L_\rho
=v_i-(1-\rho)-\frac\rho k\sum_{j\sim i}\sigma(u_i+u_j).
\]
Choose \(i\) minimizing correctness probability, \(v_i=a:=\min_jv_j\).
Every neighbor satisfies \(u_j\ge u_i\), so
\[
\frac n2\partial_{u_i}\mathcal L_\rho
\le\frac{(a-1)Q_\rho(a)}{a^2+(1-a)^2},\qquad
Q_\rho(a)=2a^2+(\rho-2)a+1-\rho.
\]
For \(0\le\rho<\rho_c\), this quadratic has the positive lower bound \(\delta_\rho=(4-4\rho-\rho^2)/8>0\).
Since \(a^2+(1-a)^2\le1\),
\begin{equation}
\|\nabla_u\mathcal L_\rho\|_2
\ge\frac{2\delta_\rho}{n}(1-\min_i v_i).
\label{eq:recovery-gradient-bound}
\end{equation}
Thus the gradient cannot vanish while even one concept remains bounded away from correctness probability one.

With \(h(t)=\sigma(t)(1-\sigma(t))\), the semantic-logit Hessian satisfies
\begin{equation}
H_u(u):=\nabla_u^2\mathcal L_\rho(u)
=\frac2n\operatorname{diag}(h(u_i))
-\frac\rho m\sum_{\{i,j\}\in E_G}h(u_i+u_j)(e_i+e_j)(e_i+e_j)^\top
\preceq\frac1{2n}I.
\label{eq:identifiable-logit-hessian}
\end{equation}
Consequently \(\nabla_w^2\mathcal L_\rho\preceq\beta I\), and Taylor's integral formula yields
\begin{equation}
\mathcal L_\rho(w_{t+1})
\le\mathcal L_\rho(w_t)-\alpha(1-\alpha\beta/2)\|\nabla_w\mathcal L_\rho(w_t)\|_2^2.
\label{eq:recovery-descent}
\end{equation}
Because \(\mathcal L_\rho\ge0\) and \(0<\alpha\beta<2\), the parameter gradient tends to zero.
Full row rank gives \(\|\nabla_w\mathcal L_\rho\|_2=\|F^\top S\nabla_u\mathcal L_\rho\|_2\ge\sigma_{\min}(F)\|\nabla_u\mathcal L_\rho\|_2\), so~\eqref{eq:recovery-gradient-bound} implies \(\min_i v_{i,t}\to1\) from every finite initialization without assuming bounded iterates or coordinatewise monotonicity, proving~(i).

At \(\rho=\rho_c\), \(Q_{\rho_c}\) has the double root \(a_c=(2-\rho_c)/4\in(0,1/2)\).
Every coordinate gradient vanishes at \(v=a_c\one_n\), and full row rank realizes this all-wrong stationary set by a finite head \(w_0\), from which \(w_t=w_0\) for all \(t\), proving~(ii).

For \(\rho_c<\rho<1\), Proposition~\ref{prop:regular-query-minimum} shows that the wrong stationary manifold realizing \(v_*\one_n\) has positive-definite Hessian \(H_{\mathrm{eff}}\) on \(\operatorname{row}(F)\), and the global upper bound above gives \(\lambda_{\max}(H_{\mathrm{eff}})\le\beta\).
Hence \(0<\alpha\beta<2\) makes \(I-\alpha H_{\mathrm{eff}}\) strictly contractive under the stated step bound while preserving the component in \(\ker F\), proving~(iii).
\end{proof}

When \(\rho_o>\rho_c\), temporarily lowering the mixed-query weight below \(\rho_c\) guarantees progress toward the true concepts, raising the practical question of when the original objective \(\mathcal L_{\rho_o}\) can be safely restored without concept labels.

\begin{corollary}[Safe restoration after a finite intervention]
\label{cor:safe-restoration}
In the model of Proposition~\ref{prop:global-recovery}, let \(0\le\rho_o<1\) be the original weight.
From any finite head, train temporarily with \(0\le\rho_{\mathrm{low}}<\rho_c\) until
\begin{equation}
\text{all observed hard counts are correct},\qquad
\mathcal L_{\rho_o}(w)<B:=\frac{2\log2}{n}.
\label{eq:safe-restoration-rule}
\end{equation}
Then restore \(\rho_o\) and continue exact full-batch GD, using fixed positive steps satisfying \(\alpha\beta<2\) in both phases.
The stopping time is finite (including when checked along any unbounded sequence of discrete observation times), every concept remains correct after restoration, and \(\min_i v_{i,t}\to1\).
\end{corollary}

\begin{proof}
Proposition~\ref{prop:global-recovery}(i) gives \(v_i\to1\) during the \(\rho_{\mathrm{low}}\) phase, so eventually every hard count is correct and \(\mathcal L_{\rho_o}(w)\to0\), meeting both conditions in~\eqref{eq:safe-restoration-rule} in finitely many updates; the covering homogeneous matching ensures that hard-count correctness implies concept correctness.

At any label-boundary state \(u_i=0\), each of the \(k\) incident mixed queries has evidence \(1/2\) (contributing \(2\rho_o\log2/n\) in total) and the covering homogeneous query has evidence at most \(1/2\) (contributing at least \(2(1-\rho_o)\log2/n\)), while all other query losses are nonnegative, so \(\mathcal L_{\rho_o}\ge B\).
For a restored update with \(g=\nabla_w\mathcal L_{\rho_o}(w)\) and any \(s\in[0,1]\), the global Hessian bound \(\nabla_w^2\mathcal L_{\rho_o}\preceq\beta I\) gives
\[
\mathcal L_{\rho_o}(w-s\alpha g)
\le\mathcal L_{\rho_o}(w)-s\alpha(1-s\alpha\beta/2)\|g\|_2^2
\le\mathcal L_{\rho_o}(w)<B,
\]
so no restored update segment can cross a label boundary, preserving all correct concepts by induction.
In the correct orthant, \(a=\min_i v_i>1/2\) gives \(Q_{\rho_o}(a)\ge Q_{\rho_o}(1/2)=(1-\rho_o)/2>0\) and hence \(\|\nabla_u\mathcal L_{\rho_o}\|_2\ge(1-\rho_o)(1-a)/n\), which together with~\eqref{eq:recovery-descent} and full row rank yields \(\min_i v_{i,t}\to1\).
\end{proof}

\subsection{A contracting neighborhood of the finite wrong state}
\label{app:wrong-basin-certificate}

The regular-query model in Appendix~\ref{app:identifiable-minima} also admits an explicit contracting ball around its finite wrong state, allowing direct verification that trajectories from random initialization enter its attraction basin.
Let \(F\in\mathbb R^{n\times d}\) have full row rank, set \(s_i=2c_i-1\), and write the semantic logits as \(u_i=s_i(Fw)_i\).
For \(2\sqrt2-2<\rho<1\), take \(v_*\) from Proposition~\ref{prop:regular-query-minimum} and set \(a=-\log(v_*/(1-v_*))\) and \(\ell_*=-a s\).
For a trajectory starting at \(w_0\), the corresponding finite wrong head preserving its \(\ker F\) component is \(w_*=w_0+F^+(\ell_*-Fw_0)\).

Choose \(0<b<a\) such that \(1-2\rho\kappa_b>0\), where
\[
t=e^{-b},\qquad \kappa_b=\frac{t(1+t)^2}{(1+t^2)^2},\qquad
R=\frac{a-b}{\max_i\|F_i\|_2}.
\]
For a thin SVD \(F=U\Sigma V^\top\), consider the ball within the affine subspace preserved by gradient updates,
\[
\mathcal B_{w_0}=\{w\in w_0+\operatorname{im}F^\top:\|V^\top(w-w_*)\|_2\le R\}.
\]
Every semantic logit in \(\mathcal B_{w_0}\) lies in \([-(2a-b),-b]\).
Writing \(r_i=e^{u_i}\le t\) and \(r_j=e^{u_j}\le t\) in~\eqref{eq:identifiable-logit-hessian}, monotonicity of \((r_i,r_j)\mapsto r_j(1+r_i)^2/(1+r_ir_j)^2\) on \((0,t]^2\) gives \(h(u_i+u_j)/h(u_i)\le\kappa_b\), so Gershgorin's theorem with \(k/m=2/n\) yields
\[
H_u\succeq\mu_\ell I,\qquad
\mu_\ell=\frac2n h(2a-b)(1-2\rho\kappa_b)>0,
\]
while \(H_u\preceq I/(2n)\) globally.
Consequently, the effective parameter Hessian on \(\mathcal B_{w_0}\) satisfies
\[
\mu I\preceq H_{\mathrm{eff}}\preceq L I,\qquad
\mu=\sigma_{\min}(F)^2\mu_\ell,\qquad L=\frac{\sigma_{\max}(F)^2}{2n}.
\]
For ordinary GD \(w^+=w-\alpha\nabla\mathcal L_\rho(w)\) with \(0<\alpha L<1\), integrating along the segment from \(w_*\) to \(w\) gives
\[
\|V^\top(w^+-w_*)\|_2\le(1-\alpha\mu)\|V^\top(w-w_*)\|_2.
\]
Thus \(\mathcal B_{w_0}\) is forward invariant and contracts geometrically to \(w_*\), with membership checkable directly from logits via \(\|V^\top(w-w_*)\|_2=\|\Sigma^{-1}U^\top(Fw-\ell_*)\|_2\).
At \(\rho=0.98\), the choice \(b=\log4\) satisfies \(1-2\rho\kappa_b>0\).

\section{Experimental Protocols and Supplemental Results}
\label{app:experiments}

This appendix follows the order of Section~\ref{sec:experiments}.
Appendices~\ref{app:cifar-llp-model} and~\ref{app:cifar-llp-results} detail the 24-image CIFAR-10 count-supervision protocol, selected explanations, and paired test-accuracy differences (Section~\ref{sec:selection-experiments}).
Appendices~\ref{app:cifar-llp-timing}--\ref{app:cifar-restricted} treat post-capture metric interventions, step-size scaling and certification-delay bounds, and the rank-deficient shared head (Section~\ref{sec:capture-experiments}).
Appendices~\ref{app:identifiable-experiments} and~\ref{app:natural-initialization} cover suboptimal attraction under identifiable counts, natural reachability from standard random initialization, sparse direct-query coverage, and safe restoration after temporary reweighting (Section~\ref{sec:identifiable-counts}).
Appendix~\ref{app:end-to-end-optimizers} documents the 1,024-image end-to-end SGD and Adam comparison and frozen-encoder counterpart (Section~\ref{sec:end-to-end}), and Appendix~\ref{app:sudoku} tests end-to-end four-class Sudoku rules beyond count supervision.
Finite-radius checks of local stability accompany Appendix~\ref{app:random-candidates}.
Every binary experiment uses exact posterior marginals as detached targets and refreshes them at each update.

\subsection{CIFAR-10 count supervision and the exact binary model}
\label{app:cifar-llp-model}

We use the binary bird/cat setting of learning from label proportions \citep{barucic2022fast}, specializing its complete-configuration posterior and detached marginal-target update to a fixed representation so that training coincides with Section~\ref{sec:dynamics}.
The encoder is frozen, the posterior is refreshed at every full-batch head step, and no probability floor or auxiliary supervised loss is added.

\subsubsection{Disjoint counts and global evidence}

Each sampled training set contains 24 official CIFAR-10 training images partitioned into six bags of four with cat-positive counts \(k=(1,2,3,1,2,3)\).
Let \(A\in\{0,1\}^{6\times24}\) be the bag-membership matrix, \(A_{bi}=1\) iff image \(i\) belongs to bag \(b\).
With \(h_i\in\{0,1\}\) and complete code \(z=2h-\one\), the legal candidates are
\begin{equation*}
\mathcal C=\{2h-\one:Ah=k\}=\prod_{b=1}^6\mathcal C_b,
\qquad |\mathcal C|=\left[\binom41\binom42\binom43\right]^2=9216.
\end{equation*}
Every change between two legal assignments exchanges a positive and a negative bit within some bag, so \(d_{\min}(\mathcal C)=2\).
The one-cat and three-cat bags exclude a global label complement, and within-bag exchanges span \(S_{\mathcal C}=\ker A\) of dimension \(24-6=18\).

Training and metric construction operate on raw cat-positive logits without per-image labels.
To relate raw logits \(\ell\) to the semantic coordinates of Section~\ref{sec:dynamics} for analysis, let \(c^\star\in\{\pm1\}^{24}\) be the true label code and \(D=\operatorname{diag}(c^\star)\).
Setting \(y=D\ell\), \(\mathcal C_{\rm sem}=D\mathcal C_{\rm raw}\), and \(K_{\rm sem}=DK_{\rm raw}D\) leaves candidate likelihoods, global evidence, Hamming distances, positive definiteness, and active spectra invariant while mapping \(R_{\rm sem}(D\ell)=DR_{\rm raw}(\ell)\).

For raw cat-positive logits \(y\), let \(\mathcal E_b\) sum the product Bernoulli probabilities of the \(\binom4{k_b}\) legal assignments in bag \(b\).
Bag disjointness gives
\begin{equation*}
\mathcal E_{\rm all}(y)=\prod_{b=1}^6\mathcal E_b(y_b),\qquad
\Phi(y)=2\sum_{b=1}^6\log\mathcal E_b(y_b),\qquad
R_i(y)=2\{p_i(y)-\sigma(y_i)\},
\end{equation*}
where \(p_i\) is the exact bag-count posterior marginal for image \(i\), coinciding with the global complete-code marginal.
Capture requires \(\sum_b\log\mathcal E_b>-\log2\), certifying the full 24-image code even when off-diagonal Gram entries couple different bags.
An independent Bernoulli-count dynamic program verifies all enumerated marginals and evidence values.

\subsubsection{A shared head with a full-rank sample metric}

Features are the 512 pre-classifier outputs of torchvision ResNet-18 (\texttt{IMAGENET1K\_V1}) in evaluation mode under the official deterministic transform (shorter edge resized to 256, \(224\times224\) center crop, ImageNet normalization, no augmentation, extracted in float32 with head updates and posteriors in float64).
For extracted row \(f_i\), the design row is \(x_i=(f_i/\|f_i\|_2,1)/\sqrt2\in\mathbb R^{513}\).
All sampled training sets retain full row rank 24 with \(\sigma_{24}/\sigma_1\) between \(0.0475\) and \(0.0507\).

Every entry of \(w_0\), including the bias, is drawn independently from \(\operatorname{Uniform}(-1/\sqrt{513},1/\sqrt{513})\) and shared across metric conditions.
For any prescribed \(K\succ0\), the positive-definite parameter metric
\begin{equation*}
M=X^+K(X^+)^\top+(I-X^+X),\qquad XMX^\top=K,
\end{equation*}
acts as the identity on \(\ker X\) and induces
\begin{equation}
w_{t+1}=w_t+\eta MX^\top R_t,\qquad
y_{t+1}=y_t+\eta KR_t,\qquad
w_t=w_0+X^+(y_t-y_0),
\label{eq:cifar-head-update}
\end{equation}
with fixed effective step \(\eta=0.032\) (corresponding to parameter step \(2N\eta\) under mean binary cross-entropy without the factor two).
Direct integration of the parameter updates reproduces sample-coordinate trajectories within \(3.78\times10^{-13}\) maximum logit difference.

\subsubsection{Four Gram orientations with common full and active spectra}

Let \(P\in\mathbb R^{24\times18}\) contain orthonormal within-bag Helmert contrasts and \(T\in\mathbb R^{24\times6}\) contain normalized bag indicators (nonzero entries \(1/2\)), so \(\operatorname{im}P=\ker A\) and \([P\ T]\) is orthogonal.
With \(\Lambda=\operatorname{diag}(\lambda_1,\ldots,\lambda_{18})\) and
\begin{equation*}
\lambda_j=4\,16^{-(j-1)/17},\qquad
K_Q=PQ\Lambda Q^\top P^\top+TT^\top,
\end{equation*}
every orthogonal \(Q\in\mathbb R^{18\times18}\) yields identical full spectrum \(\operatorname{spec}(\Lambda)\cup\{1\}^6\) and active spectrum \(\operatorname{spec}(\Lambda)\) (verified to floating-point roundoff).
The \emph{feature} condition sets \(Q\) to the descending eigenbasis of \(P^\top XX^\top P\).
The \emph{feature-reverse} condition reverses the eigenvalue ordering on that basis, while \emph{random 1} and \emph{random 2} use two independent Gaussian-QR bases drawn once per training set without labels or per-head tuning.
Two non-isospectral references are also recorded: \(K_{\rm iso}=\operatorname{tr}(K_Q)I/24\) and \(K_{\rm raw}=4XX^\top/\lambda_{\max}(XX^\top)\) (scaled ordinary parameter GD).

\subsection{Selected explanations, semantic accuracy, and paired test differences}
\label{app:cifar-llp-results}

Six disjoint 24-image training sets are sampled from CIFAR-10 with four random heads per set.
Figure~\ref{fig:cifar-llp} illustrates feature and feature-reverse on the first evaluation set and head.
A state is \emph{well fitted} when its hard bag counts are legal, mean bag evidence is at least \(0.95\), and every image's modal probability is at least \(0.90\); a \emph{strictly resolved} final endpoint also retains the same complete 24-bit hard code across steps 7,900--8,000.

\begin{table}[!htbp]
\centering
\caption{Per-condition resolution, capture, and training concept accuracy across the six 24-image evaluation sets (24 trajectories per row, cat-positive convention without label swap).
Accuracies report median [minimum, maximum] percentages, and the bottom two reference rows do not share the matched spectrum.}
\label{tab:cifar-semantic}
\small
\begin{tabular}{lrrl}
\toprule
Metric & Resolved & Captured & Train concept accuracy\\
\midrule
Feature & 24/24 & 24/24 & 50.00 [33.33, 100]\\
Feature-reverse & 24/24 & 24/24 & 58.33 [50.00, 83.33]\\
Random 1 & 24/24 & 24/24 & 58.33 [41.67, 83.33]\\
Random 2 & 24/24 & 24/24 & 58.33 [41.67, 83.33]\\
Isotropic (reference) & 24/24 & 24/24 & 58.33 [41.67, 100]\\
Raw scaled (reference) & 0/24 & 0/24 & 85.42 [79.17, 100]\\
\bottomrule
\end{tabular}
\end{table}

Across all six sets, all 96 matched-spectrum runs resolve and cross half evidence, and in all 144 within-group pairs the four matched-spectrum Grams select distinct noncomplement complete codes from every initial head, with pairwise Hamming distances ranging from 2 to 14 (median 10), final mean bag evidence in \([0.9922,0.9961]\), and minimum per-image modal probability in \([0.9929,0.9975]\).
Of the 96 captured matched-spectrum assignments, 93 mislabel at least one training image (Table~\ref{tab:cifar-semantic}).
The raw-scaled reference attains higher median training concept accuracy within the 8,000-step budget, yet none of its 24 runs meets the fitting or capture criteria, illustrating the distinction between recovering true labels and fitting counts to a captured explanation.

\paragraph{Paired test-accuracy differences.}
Every trained head is evaluated on all 2,000 official CIFAR-10 bird/cat test images (concept classification) and 500 fixed four-image test bags (count reasoning, requiring the predicted bag count to match exactly).
Across the 96 matched-spectrum runs, mean (\(\pm\) sample SD) test accuracy is \(57.23\%\pm13.24\%\) on concepts and \(31.28\%\pm8.78\%\) on counts.
Across all \(24\times6=144\) unordered pairs sharing training images and initial head, the descriptive median absolute difference in test accuracy is \(11.23\) percentage points on concept classification (per-set medians: \(10.08,8.60,11.50,15.75,14.95,9.23\)) and \(5.00\) percentage points on count reasoning.

\subsection{Capture under subsequent metric changes}
\label{app:cifar-llp-timing}

The four matched-spectrum Grams and isotropic reference satisfy \(0.25I\preceq K\preceq4I\) and \(\eta\kappa_+(N+1)=0.032\cdot4\cdot25=3.2<4\).
At each run's first global capture step \(T_{\rm cap}\) (\(\mathcal E_{\rm all}>1/2\)), we record the current sign code \(z\) and launch three 4,000-step continuations.
The first swaps feature with feature-reverse, swaps random 1 with random 2, and switches the isotropic reference to feature.
For the other two continuations, let \(i\) minimize \(z_i y_i\) and set
\begin{equation*}
u=\frac{z_i e_i+R/\|R\|_2}{\|z_i e_i+R/\|R\|_2\|_2},
\qquad K=4I-3.75uu^\top,
\end{equation*}
which minimizes instantaneous signed drift on the weakest captured coordinate within the one-slow-eigenvalue family on \([0.25,4]\) (using an orthogonal unit vector if the numerator vanishes), without access to future endpoints or ground-truth labels.
One continuation fixes \(u\) at \(T_{\rm cap}\) and the other recomputes \(u\) at every update.
Across the 288 matched-spectrum continuations and 72 isotropic-reference continuations, zero runs change a captured label or decrease global log evidence beyond \(10^{-10}\), as guaranteed by Theorem~\ref{thm:routing-certificate}.

\subsection{Timing of capture and step-size dependence}
\label{app:capture-timing}

For the six-bag model above, the sufficient step-size bound is \(4/[4(24+1)]=0.04\).
Repeating the six-set experiment at \(\eta\in\{0.008,0.020,0.032,0.038\}\) (20\%, 50\%, 80\%, and 95\% of the bound), using training horizon \(\lceil 256/\eta\rceil\) and continuation budget \(\lceil 128/\eta\rceil\) across the same six training sets, four initial heads, and four matched-spectrum Grams, yields 384 runs: all 384 cross \(\mathcal E_{\rm all}>1/2\), fit, and select the exact same final assignment as at \(\eta=0.032\).
Median capture steps are \(3{,}608\), \(1{,}446.5\), \(906\), and \(763.5\), corresponding to nearly invariant cumulative step lengths \(\eta T_{\rm cap}\in[28.86,29.01]\) and occurring at a median 65\% of the updates required to meet the high-confidence fitting criterion.
Applying the three continuations of Appendix~\ref{app:cifar-llp-timing} at each first crossing gives \(1{,}152\) further continuations with zero label changes.

\paragraph{Upper-bounding the certification delay.}
Along an individual trajectory, hard labels may stop changing before \(\mathcal E_{\rm all}>1/2\) triggers.
Let \(T_*\) be the earliest step on a trajectory from which every continuation obeying \(0.25I\preceq K_t\preceq4I\) preserves the current hard labels, so \(T_*\le T_{\rm cap}\) by Theorem~\ref{thm:routing-certificate}.
On all 96 baseline trajectories (\(\eta=0.032\)), we search from saved states after the original run's last label change for admissible continuations (constructed from current predictions and bag counts without concept labels) that still flip at least one label.
Such witnesses \(T_{\rm flip}<T_*\) exist on all 96 runs: the resulting upper bounds \(T_{\rm cap}-T_{\rm flip}\ge T_{\rm cap}-T_*\) on certification delay range from 161 to 322 updates (median 240), and the ratio \((T_{\rm cap}-T_{\rm flip})/T_{\rm cap}\) has a median of 26.6\%.
Apparent label stabilization along a single trajectory therefore does not imply irreversibility, while \(\mathcal E_{\rm all}>1/2\) triggers within this bounded fraction of the capture time.

\subsection{Capture with a rank-deficient shared head}
\label{app:cifar-restricted}

To test Corollary~\ref{cor:fixed-linear-capture} under a singular sample Gram, we train \(d=33\) parameters on \(N=96\) images arranged in 24 four-image bags with counts \((1,2,3)\) repeated eight times.
Frozen ResNet-18 features are row-normalized, centered on the 10,000-image unlabeled training pool, projected onto its top 32 covariance eigenvectors, whitened by pool standard deviations, augmented with a bias, and scaled by \(1/\sqrt{33}\) without using per-image labels or bag counts.
We evaluate four disjoint 96-image training sets (disjoint from the 24-image sets above) with four random heads \(\operatorname{Uniform}(-1/\sqrt{33},1/\sqrt{33})\) per set.

Each head is trained under three parameter metrics \(M\): \(I_{33}\), a diagonal with 33 geometrically spaced entries from \(0.5\) to \(2\), and its reverse, using the shared disjoint-bag step \(\eta=3.2/[2\|X\|_{\mathrm{op}}^2\cdot5]\) satisfying \(\eta m_+\|X\|_{\mathrm{op}}^2(4+1)=3.2<4\) (\(m_+=2\)).
Every design matrix has rank 33 and sample nullity 63 (parameter updates are verified within \(1.43\times10^{-14}\)).
At the first step where \(\prod_{b=1}^{24}\mathcal E_b>1/2\), three 4,000-step continuations apply \(I_{33}\), the opposite fixed diagonal, or an adaptive diagonal choosing entries in \(\{0.5,2\}\) at each step to minimize instantaneous drift on the weakest captured margin \(z_i(Xw)_i\).

All 48 evaluation runs fit and cross global half evidence, every captured assignment contains at least one concept error, and all \(48\times3=144\) post-capture continuations preserve every captured label (Table~\ref{tab:cifar-restricted}).

\begin{table}[!htbp]
\centering
\caption{Capture and training concept accuracy with a 33-parameter shared head on 96 images (four evaluation sets, 16 runs per row; fit, resolution, and global capture counts coincide).}
\label{tab:cifar-restricted}
\small
\begin{tabular}{lrrr}
\toprule
Parameter metric & Fit and capture & Median \(T_{\rm cap}\) & Median train concept acc.\\
\midrule
Identity & 16/16 & 3{,}163.5 & 83.33\%\\
Forward diagonal & 16/16 & 3{,}796.5 & 65.63\%\\
Reverse diagonal & 16/16 & 2{,}355.0 & 87.50\%\\
\bottomrule
\end{tabular}
\end{table}

\subsection{Suboptimal attraction under identifiable counts}
\label{app:identifiable-experiments}

We next detail the local-curvature and perturbation experiments accompanying Section~\ref{sec:identifiable-counts}, testing Proposition~\ref{prop:regular-query-minimum} and Proposition~\ref{prop:global-recovery} (Appendices~\ref{app:identifiable-minima}--\ref{app:global-recovery}) on three disjoint 256-image CIFAR-10 sets (128 birds and 128 cats per set) with full-batch GD step size \(\alpha=256/(1.5\|F\|_{\mathrm{op}}^2)\).
Raw 512-dimensional frozen ResNet-18 features plus a bias form \(F\in\mathbb R^{256\times513}\) of rank 256 (supervised controls reach 100\% training concept accuracy on every set).
Each set pairs images into 128 disjoint same-class queries covering all 256 images and 1,024 mixed queries forming a connected 8-regular bipartite graph, giving a count-constraint matrix of full rank 256.

\paragraph{Analytic wrong centers and perturbation contraction.}
On each of the three sets and \(\rho\in\{0.85,0.90\}\), we realize the analytic wrong probabilities \(1-v_*\in\{0.80,0.8850781\}\) via the minimum-norm head (losses \(0.81064448\) and \(0.63739642\), zero concept accuracy, violating every same-class query).
Explicit pair-product raw-logit Hessians match \([v_*(1-v_*)]^2H_v\) from~\eqref{eq:identifiable-graph-hessian} up to sign conjugation within \(5.9\times10^{-17}\), with positive effective eigenvalues reported in Table~\ref{tab:identifiable-centers} (the remaining 257 parameter directions span \(\ker F\)).

\begin{table}[!htbp]
\centering
\caption{Curvature and local update stability at the six analytic wrong centers on the three 256-image evaluation sets.
Here \(H_{\mathrm{eff}}\) is the parameter Hessian restricted to \(\operatorname{row}(F)\), and \(r_J\) is the spectral radius of \(I-\alpha H_{\mathrm{eff}}\) (maximum parameter-gradient component \(\le1.78\times10^{-13}\)).}
\label{tab:identifiable-centers}
\small
\begin{tabular}{ccrrr}
\toprule
Set & \(\rho\) & \(\lambda_{\min}(H_{\mathrm{eff}})\) & \(\lambda_{\max}(H_{\mathrm{eff}})\) & \(r_J\) \\
\midrule
1 & 0.85 & 0.00297946 & 165.5725 & 0.999996177 \\
1 & 0.90 & 0.00233763 & 105.4787 & 0.999997000 \\
2 & 0.85 & 0.00294513 & 169.1343 & 0.999996301 \\
2 & 0.90 & 0.00231732 & 107.7610 & 0.999997090 \\
3 & 0.85 & 0.00334100 & 164.3945 & 0.999995688 \\
3 & 0.90 & 0.00266123 & 104.7974 & 0.999996565 \\
\bottomrule
\end{tabular}
\end{table}

\begin{figure}[!htbp]
\centering
\includegraphics[width=\textwidth]{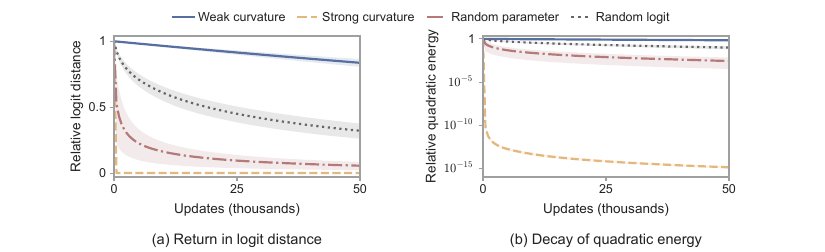}
\caption{Local contraction over 50,000 full-batch GD updates after perturbing the six analytic all-wrong stationary heads in Table~\ref{tab:identifiable-centers} at initial RMS logit distance \(0.01\) along four perturbation families.
(a) RMS logit distance and (b) local quadratic energy \(\delta\ell^\top H_\ell\delta\ell/2\), each normalized by its initial value, showing medians and full ranges across the three 256-image training sets and weights \(\rho\in\{0.85,0.90\}\).}
\label{fig:identifiable-energy}
\end{figure}

Perturbing each of the six centers along the weakest and strongest \(H_{\mathrm{eff}}\) eigenvectors, isotropic Gaussian parameter directions, and minimum-norm Gaussian logit displacements at initial RMS logit distance \(0.01\) produces 24 trajectories over 50,000 full-batch GD updates.
All 24 contract monotonically in logit distance, probability distance, effective parameter distance, and local quadratic energy while keeping zero concept accuracy and \(\ker F\) drift at most \(1.80\times10^{-13}\) (Figure~\ref{fig:identifiable-energy}), with contraction slowest along the weak-curvature direction and fastest along strong curvature.

\subsection{Natural reachability, sparse direct queries, and safe restoration}
\label{app:natural-initialization}

\paragraph{Weight sweep and sparse direct-query coverage from random initialization.}
For the random-initialization sweep, we sample three mutually disjoint 256-image sets under an independent random seed, with 32 shared label-free random heads \(\operatorname{Uniform}(-1/\sqrt{512},1/\sqrt{512})\) per set (\(96\) runs per condition), and train for 500,000 updates using step size \(\alpha=1/(1.5\max_i\|f_i\|_2^2)\) (where \(f_i\in\mathbb R^{513}\) is row \(i\) of \(F\)) under both full-batch GD and 32-query minibatch SGD (sampling 32 queries with replacement per step according to the query weights).
Table~\ref{tab:natural-failures} and Figure~\ref{fig:identifiable-minima}(a) vary \(\rho\in[0.84,0.99]\) on these shared sets and initial heads while keeping full same-class coverage: every run finishes either completely correct or completely wrong, with no label change over its final 20,000 updates.
In a companion experiment on the same three sets and heads fixing equal weight on every retained query while keeping same-class queries on \(256\), \(32\), or \(4\) of the \(256\) inputs (which raises the total mixed-query weight to \(8/9\), \(64/65\), and \(512/513\) while graph propagation still uniquely determines all 256 labels), \(0,37\), and \(49\) of \(96\) GD runs and \(0,36\), and \(48\) of \(96\) SGD runs finish with every concept label wrong.

\paragraph{Contracting-ball verification and safe restoration.}
At \(\rho=0.98\), the analytic wrong state assigns true-label probability \(0.0204\) to every input (mean loss \(0.1957\)).
Continuing all \(25/96\) failed GD trajectories at \(\rho=0.98\) from Table~\ref{tab:natural-failures} under unchanged full-batch GD brings every trajectory into the contracting ball \(\mathcal B_{w_0}\) of Appendix~\ref{app:wrong-basin-certificate} (with \(b=\log4\)) between \(11.2\times10^6\) and \(12.3\times10^6\) total updates (Figure~\ref{fig:identifiable-minima}(b)), certifying geometric convergence to the reversed-prediction stationary set of Proposition~\ref{prop:regular-query-minimum}.
Starting from those same 25 naturally reached wrong states at \(\rho_o=0.98\), we test Corollary~\ref{cor:safe-restoration} (\(\alpha\beta\le0.134<2\)) by temporarily setting \(\rho_{\mathrm{low}}=0.80<\rho_c\), checking the label-free criterion~\eqref{eq:safe-restoration-rule} every 1,000 updates, and restoring \(\rho_o=0.98\) upon first satisfaction.
All \(25/25\) runs meet~\eqref{eq:safe-restoration-rule} within \(380{,}000\text{--}454{,}000\) updates, retain 100\% training concept correctness at every recorded post-restoration state, and reach median unseen test concept and count accuracies of \(89.05\%\) and \(79.65\%\), whereas unchanged GD over 600,000 updates recovers \(0/25\) runs and remains at \(18.55\%\) and \(36.63\%\).

\begin{table}[!htbp]
\centering
\caption{Natural reachability of wrong predictions from standard random initialization under uniquely identifying counts (Figure~\ref{fig:identifiable-minima}(a)): runs out of 96 per optimizer and weight \(\rho\) whose concept predictions are all wrong after 500,000 updates, while every other run recovers all 256 labels.}
\label{tab:natural-failures}
\small
\begin{tabular}{lrrrrrrrrr}
\toprule
\(\rho\) & 0.84 & 0.86 & 0.88 & 0.90 & 0.92 & 0.94 & 0.96 & 0.98 & 0.99\\
\midrule
GD & 0 & 0 & 0 & 0 & 0 & 2 & 7 & 25 & 42\\
SGD & 0 & 0 & 0 & 0 & 0 & 2 & 6 & 26 & 42\\
\bottomrule
\end{tabular}
\end{table}

\subsection{Ordinary optimizers with jointly learned representations}
\label{app:end-to-end-optimizers}

Four disjoint 1,024-image bird/cat training sets are sampled from the official CIFAR-10 pool, each partitioned into 256 four-image bags with counts cycling through \((1,2,3)\) and eight random initial heads per set (\(32\) initial networks).
In the head parameterization \(v=w/8\) with logit \(8\,v^\top x\) on the normalized design row \(x\) of Appendix~\ref{app:cifar-llp-model} and \(v_0=w_0/8\), all ResNet-18 parameters and the 513-parameter head are trained end-to-end (with fixed pretrained BatchNorm running statistics, trainable affine parameters, and no augmentation or weight decay) for 100 epochs (\(3{,}200\) updates) on identical minibatches of eight bags, minimizing mean negative log exact bag-count evidence with cosine decay to 1\% of the initial step sizes.
Learning rates for (head, encoder) are chosen on a separate held-out 1,024-image tuning set (disjoint from the four evaluation training sets and the test set), where tuning runs optimize bag counts alone using no momentum for SGD and default moments for Adam.
On the \emph{original grid}, selecting solely by lowest mean final tuning count NLL gives \((0.05,0.003)\) for SGD and \((1.25\times10^{-3},10^{-5})\) for Adam.
In an \emph{expanded grid} search on the same held-out tuning set, we inspect the tuning set's true concept labels to identify rate directions that produce more concept prediction errors in the learned explanation (larger SGD head rate and larger Adam encoder rate), and along those directions select configurations that still reliably satisfy the tuning counts.
This yields the setting reported in Section~\ref{sec:end-to-end} and Figure~\ref{fig:joint-training}: \((0.15,0.001)\) for SGD, raising the head rate from \(0.05\) to \(0.15\) while lowering the encoder rate from \(0.003\) to \(0.001\) to maintain count satisfaction, and \((6.25\times10^{-3},5\times10^{-5})\) for Adam, which also achieves the lowest tuning count NLL on the expanded Adam grid and independently minimizes tuning concept cross-entropy for the 32 true-concept Adam references.
All evaluation runs on the four 1,024-image training sets optimize bag counts alone, using true concept labels only for the 32 paired references.

\begin{table}[!htbp]
\centering
\caption{End-to-end SGD vs.\ Adam under the original-grid and expanded-grid learning-rate settings on the four 1,024-image evaluation sets (32 initial networks).
The last two columns report median absolute differences in unseen accuracy (percentage points) across count-satisfying pairs.}
\label{tab:end-to-end-rates}
\small
\setlength{\tabcolsep}{4.5pt}
\begin{tabular}{@{}lrrrrr@{}}
\toprule
Learning-rate setting & Both fit counts & Different in pair & Median diff.\ labels & Concept gap & Count gap\\
\midrule
Original grid & 27/32 & 27/27 & 30 & 0.35 & 1.40\\
Expanded grid (Figure~\ref{fig:joint-training}) & 28/32 & 28/28 & 90 & 4.08 & 7.50\\
\bottomrule
\end{tabular}
\end{table}

\paragraph{Paired optimizer differences and comparison to true-concept references.}
Under both learning-rate settings, the two optimizers select different legal explanations in every common-initialization pair where both satisfy the training counts (Table~\ref{tab:end-to-end-rates}).
Under the expanded-grid setting (Figure~\ref{fig:joint-training}), Adam satisfies all 256 training counts in \(32/32\) runs and SGD in \(28/32\) (each of the other four SGD runs predicts a single class on all 2,000 test images, giving 50\% test concept accuracy).
Across the 28 count-satisfying pairs, SGD and Adam disagree on 48--200 of the 1,024 training labels (median 90, while normalized features move by a median 81\% in relative Frobenius norm), with median absolute test-accuracy differences of \(4.08\) (\(\approx4.1\), range \(0.05\text{--}10.75\)) percentage points on the 2,000 test concept images and \(7.50\) (range \(0.00\text{--}27.40\)) points on the 500 four-image test count questions.
All 32 true-concept Adam references learn 100\% of training concepts and average \(94.93\%\) test concept accuracy and \(81.92\%\) test count accuracy.
The 28 SGD and 32 Adam count-satisfying runs mislabel on average \(10.61\%\) and \(7.72\%\) of training images, average \(84.72\%\) and \(87.46\%\) on test concepts (\(10.2\) and \(7.5\) points below their paired references) and \(57.59\%\) and \(63.94\%\) on test counts (\(24.3\) and \(18.0\) points below), and across all 60 runs the test concept accuracy gap relative to the paired reference increases nearly linearly with the percentage of mislabeled training images (\(r=0.97\), least-squares slope \(1.01\), with \(r=0.95\) within SGD and \(r=0.99\) within Adam).

\paragraph{Frozen-encoder counterpart and high-evidence continuation.}
Repeating the SGD/Adam comparison on the frozen encoder of Appendix~\ref{app:cifar-llp-model} (singular 513-parameter head trained for 50,000 epochs on the first 96 images of each set and 5,000 epochs on all 1,024 images, with step sizes selected on the same held-out tuning set by lowest training count NLL without concept labels) fits all \(32/32\) pairs on both sizes, selects distinct explanations in every pair (median Hamming distances 4 and 15), and reaches \(\mathcal E_{\rm all}>1/2\) at all 64 endpoints on 1,024 inputs.
Among the 64 jointly trained models of Figure~\ref{fig:joint-training}, 49 finish with \(\mathcal E_{\rm all}>1/2\) (17 SGD, 32 Adam, all retaining concept errors).
Continuing each of these 49 models for 256 further minibatch updates with both encoder and head trainable and initial learning rates restarted changes zero training labels.

\subsection{Sudoku rules with four classes}
\label{app:sudoku}

To test whether end-to-end soft-ABL likewise satisfies non-count logical rules with incorrect concept semantics, we evaluate visual \(4\times4\) Sudoku \citep{augustine2022visual} over four CIFAR-10 classes (airplane, automobile, bird, cat).
Each board places 16 distinct \(112\times112\) images, supervised by the \(4\times4\) Sudoku rules plus the true class of the top-left cell on a prescribed fraction of training boards (always covering all four classes, with 288 legal completions unrevealed and 72 when revealed under the exact uniform-prior, unit-temperature posterior).
An ImageNet-pretrained ResNet-18 with a four-way head is trained end-to-end with Adam for 100 epochs on minibatches of 32 boards without augmentation.
The learning rate \(10^{-3}\) (from \(\{10^{-4},3\times10^{-4},10^{-3}\}\)) was selected on a separate held-out tuning board set solely by lowest final training logical NLL at 100\% clue boards, and four disjoint 250-board training sets (4,000 images) with two seeds give eight runs per row, evaluated on 200 unseen boards.

\begin{table}[!htbp]
\centering
\caption{End-to-end visual \(4\times4\) Sudoku on four CIFAR-10 classes (eight runs per row).
A run learns the concepts when training concept accuracy is at least 99\%; for remaining runs we report the percentage of training boards satisfying all rules and clues alongside the training concept accuracy range.}
\label{tab:sudoku}
\small
\begin{tabular}{lrrr}
\toprule
Boards revealing & Runs learning & Other runs: training boards & Other runs: training\\
one cell & the concepts & satisfying all constraints & concept accuracy\\
\midrule
100\% & 8 of 8 & -- & --\\
20\% & 2 of 8 & 100\% & 5\%--54\%\\
2\% & 1 of 8 & 100\% & 0.4\%--50\%\\
\bottomrule
\end{tabular}
\end{table}

When every board reveals one cell, all 8 runs recover the concepts (median test concept accuracy \(93.8\%\), matching \(93.9\%\) for true-label training; Table~\ref{tab:sudoku}).
Although revealed cells cover all four classes so only the true semantics satisfy all constraints with a consistent class mapping, when 20\% (or 2\%) of boards reveal a cell, 6 of 8 (or 7 of 8) runs satisfy every rule and revealed cell on 100\% of training boards while learning wrong concepts: each labels every revealed cell correctly and follows one class permutation on at least \(99.7\%\) of cells on unrevealed boards.

\end{document}